\documentclass[10pt,twocolumn,letterpaper]{article}
\usepackage[accsupp]{axessibility} 

\usepackage{wacv}
\usepackage{times}
\usepackage{epsfig}
\usepackage{graphicx}
\usepackage{amsmath}
\usepackage{amssymb}
\usepackage{comment}
\usepackage{multirow}
\usepackage{multicol}
\usepackage{listings}
\usepackage{fontawesome}
\usepackage{booktabs}
\usepackage{multirow}
\usepackage{caption}
\usepackage{subcaption}
\usepackage[utf8]{inputenc}
\usepackage[pagebackref=true,breaklinks=true,colorlinks,bookmarks=false]{hyperref}
\newcommand{\cls}[1]{#1}
\newcommand{\mytabsep}{\\[.5em]}
\usepackage[title]{appendix}

\usepackage{stfloats}

\def\wacvPaperID{892} 

\wacvfinalcopy 

\ifwacvfinal
\fi

\let\originalleft\left
\let\originalright\right
\renewcommand{\left}{\mathopen{}\mathclose\bgroup\originalleft}
\renewcommand{\right}{\aftergroup\egroup\originalright}

\usepackage{amsmath}
\usepackage{amssymb}  

\usepackage{mathtools}  
\usepackage{bm}

\makeatletter
\newcommand{\spx}[1]{%
	\if\relax\detokenize{#1}\relax
	\expandafter\@gobble
	\else
	\expandafter\@firstofone
	\fi
	{^{#1}}%
}
\makeatother

\newcommand{\genericdel}[4]{%
	\ifcase#3\relax
	\ifx#1.\else#1\fi#4\ifx#2.\else#2\fi\or
	\bigl#1#4\bigr#2\or
	\Bigl#1#4\Bigr#2\or
	\biggl#1#4\biggr#2\or
	\Biggl#1#4\Biggr#2\else
	\left#1#4\right#2\fi
}

\newcommand{\cbr}[2][-1]{\genericdel\{\}{#1}{#2}}
\let\set\cbr

\newcommand{\envert}[2][-1]{\genericdel||{#1}{#2}}

\usepackage{stmaryrd}  

\usepackage[OMLmathsfit]{isomath}  
\usepackage{upgreek}

\DeclareMathAlphabet{\mathbbmsl}{U}{bbm}{m}{sl}
\DeclareMathAlphabet{\mathbbmb}{U}{bbm}{b}{it}
\DeclareMathAlphabet{\mathbbmssit}{U}{bbmss}{m}{it}

\let\vec\relax
\let\set\relax
\newcommand{\vec}[1]{\bm{#1}}
\newcommand{\set}[1]{\mathbbmsl{#1}}

\newcommand{\nunder}[2][5]{\mathrlap{\mkern\the\numexpr#1/2mu\relax\underline{\phantom{\mathrm{#2}\mkern-#1mu}}}\mathrm{#2}}

\newcommand{\rvarstyle}[1]{\uppercase{#1}}
\newcommand{\rvar}[1]{\rvarstyle{#1}}

\DeclareMathOperator{\softmax}{softmax}

\let\P\relax
\DeclareMathOperator{\P}{P}

\DeclareMathOperator*{\E}{\mathrm{I\kern-.282em E}}
\DeclareMathOperator*{\D}{\mathrm{I\kern-.282em D}}

\newcommand{\enbbracket}[1]{{\mathinner{\left\llbracket{#1}\right\rrbracket}}}

\let\oldhat\hat
\renewcommand{\hat}[1]{\vphantom{#1}\smash[t]{\oldhat{#1}}}
\let\oldtilde\tilde
\renewcommand{\tilde}[1]{\vphantom{#1}\smash[t]{\oldtilde{#1}}}
\let\oldwidetilde\widetilde
\renewcommand{\widetilde}[1]{\vphantom{#1}\smash[t]{\oldwidetilde{#1}}}

\usepackage[nodisplayskipstretch]{setspace}

\newenvironment{talign*}
{\csname align*\endcsname}
{\endalign}

\usepackage{dashbox}%

\ifwacvfinal
\usepackage[breaklinks=true,bookmarks=false]{hyperref}
\else
\usepackage[pagebackref=true,breaklinks=true,colorlinks,bookmarks=false]{hyperref}
\fi

\ifwacvfinal
\fi

\begin{document}

\title{Multi-domain semantic segmentation 
  with overlapping labels\thanks{This work 
    has been supported by 
    Croatian Science Foundation -
    grant IP-2020-02-5851 ADEPT,
    by European Regional Development Fund -
    grant KK.01.2.1.02.0119 DATACROSS
    and by VSITE College for 
    Information Technologies
    who provided access to 6 GPU Tesla-V100 32GB.}}

\author{Petra Bevandić\thanks{Equal contribution.},
Marin Oršić\footnotemark[2]\textsuperscript{1}, 
Ivan Grubišić,
Josip Šarić,
Siniša Šegvić\\
University of Zagreb,
Faculty of Electrical Engineering and Computing\\
{\tt\small name.surname@fer.hr, marin.orsic@gmail.com}
}

\maketitle
\thispagestyle{empty}

\begin{abstract}
Deep supervised models 
have an unprecedented capacity 
to absorb large quantities of training data.
Hence, training on many datasets
becomes a method of choice 
towards graceful degradation 
in unusual scenes.
Unfortunately, different datasets 
often use incompatible labels.
For instance, the Cityscapes road class
subsumes all driving surfaces, 
while Vistas defines separate classes
for road markings, manholes etc.
We address this challenge 
by proposing a principled method 
for seamless learning
on datasets with overlapping classes
based on partial labels
and probabilistic loss.
Our method achieves competitive 
within-dataset and cross-dataset generalization,
as well as ability to learn visual concepts
which are not separately labeled
in any of the training datasets.
Experiments reveal competitive 
or state-of-the-art performance 
on two multi-domain dataset collections
and on the WildDash 2 benchmark. 
\end{abstract}

\section{Introduction}

Large realistic datasets
\cite{lin14eccv,cordts16cvpr,neuhold17iccv,zhou17cvpr}
have immensely contributed to development 
of dense prediction models. 
The reported accuracy grew rapidly 
over the last five years 
so that some of these datasets 
appear solved today. 
Unfortunately, the learned models  
often perform poorly in the wild 
\cite{zendel18eccv}. 
Consequently, a desire emerged
to measure the model performance by evaluating 
on multiple datasets from several domains
\cite{liang2018dynamic,zhao20eccv,lambert20cvpr}, 
akin to combined events in athletics.
The most straightforward approach
towards that goal is 
simultaneous training on many datasets.
Such training is likely to yield
outstanding resilience 
due to extremely large capacity 
of convolutional models 
\cite{zhang17iclr,sun17iccv}.

Training on multiple datasets 
is especially interesting
for dense prediction models
due to very expensive labels
\cite{zlateski18cvpr}.
However, this is not easily 
carried out in practice
since existing datasets
use incompatible taxonomies
\cite{liang2018dynamic,lambert20cvpr}.
These incompatibilities may be caused 
by discrepant granularity 
and overlapping classes.
Discrepant granularity \cite{liang2018dynamic}
arises where some class from dataset A
corresponds to several classes from dataset B.
For instance the class road in Cityscapes
is further divided into 8 classes in Vistas:
 road, bike\_lane, crosswalk\_plain, 
 marking\_zebra, marking\_other, 
 manhole, pothole, and service\_lane.
Overlapping classes occur when visual concepts
get inconsistently grouped across datasets.
For instance class truck in VIPER
\cite{richter17iccv}
includes trucks and pickups.
Class car in Vistas
\cite{neuhold17iccv}
includes cars, pickups and vans.
Class van in Ade20k
\cite{zhou17cvpr}
includes vans and pickups.
We say that VIPER truck overlaps 
Vistas car (and Ade20k van)
since the corresponding pixels have
non-empty intersection and 
non-empty set difference in both directions.
Fig.~\ref{fig:intro} shows that 
the three datasets assign pickups 
into three different classes.

\newcommand{\szfi}{.213\columnwidth}
\begin{figure}[t]
  \centering
  \begin{tabular}{@{$\,$}c@{$\;\,$}c@{$\;\,$}c@{$\,$}}
   \includegraphics[height=\szfi]{
    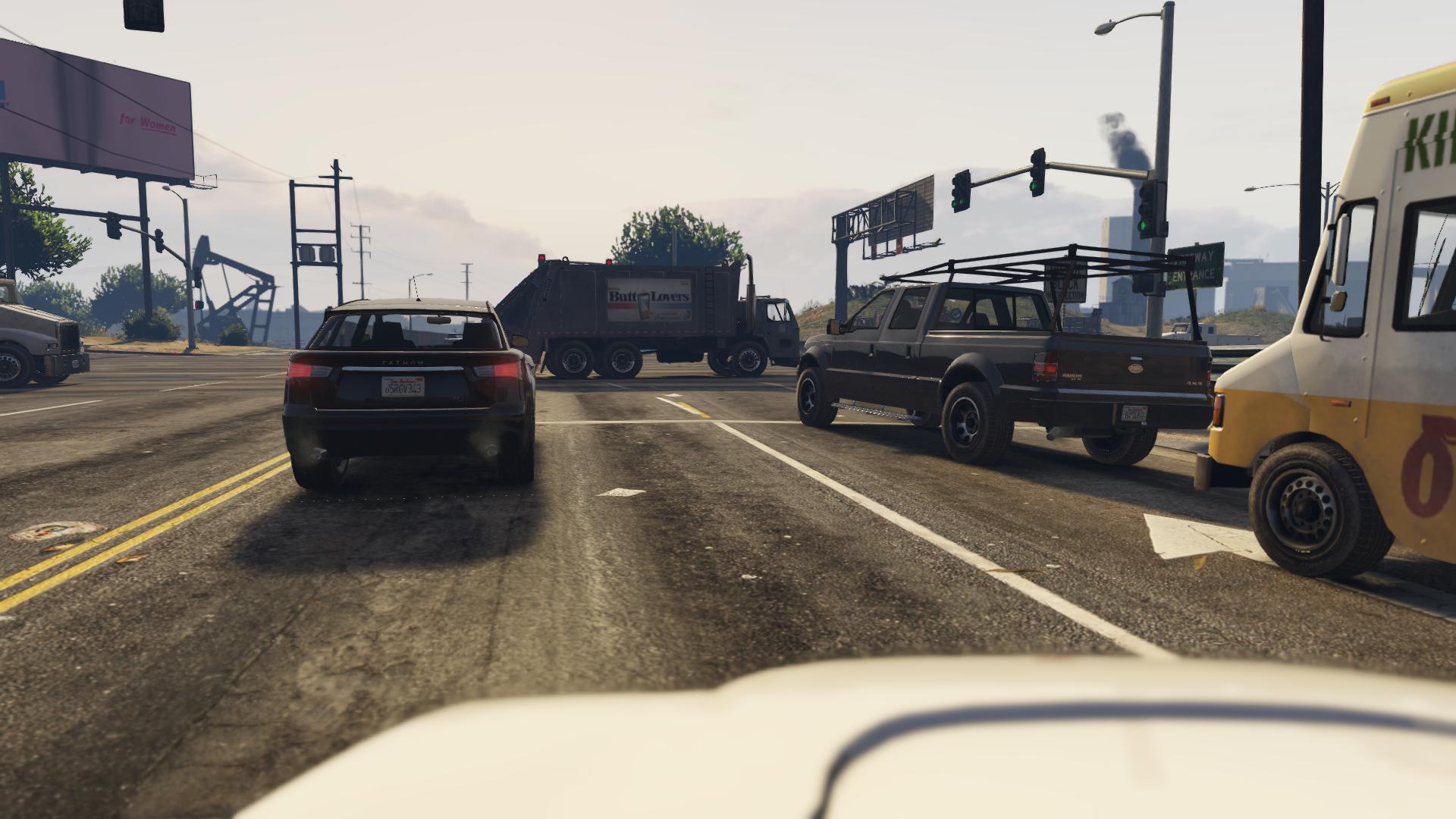}
    &
   \includegraphics[height=\szfi]{
    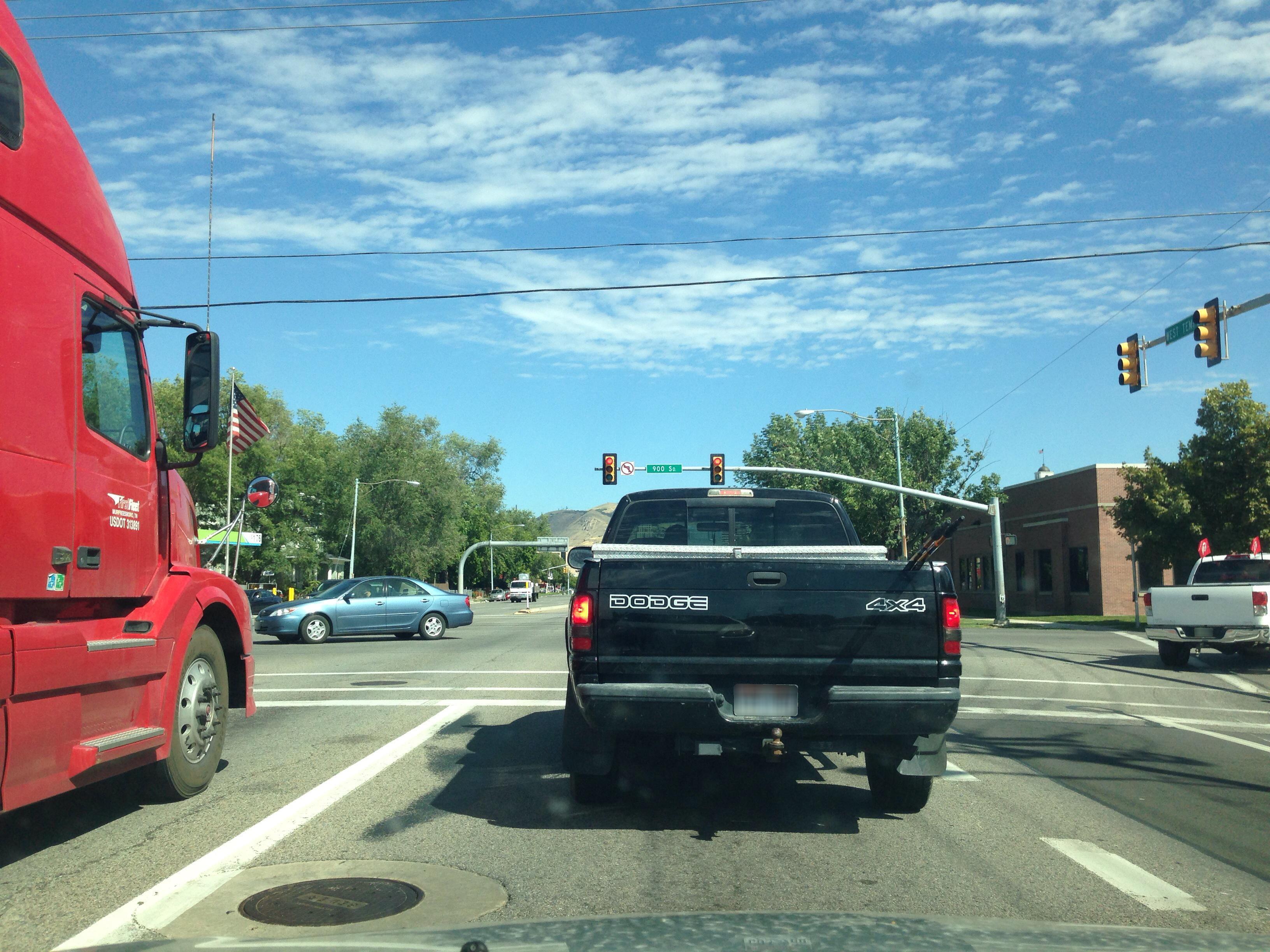}
   &
   \includegraphics[height=\szfi]{
     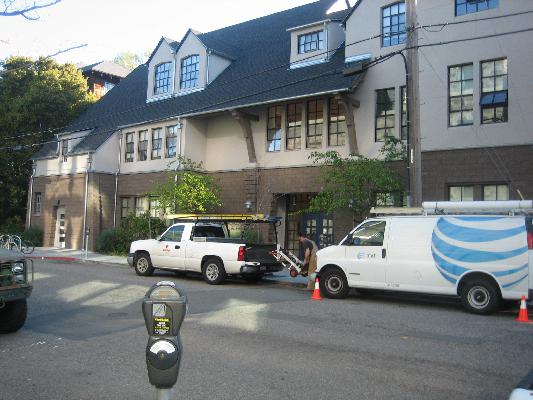}
  \\
   \includegraphics[height=\szfi]{
    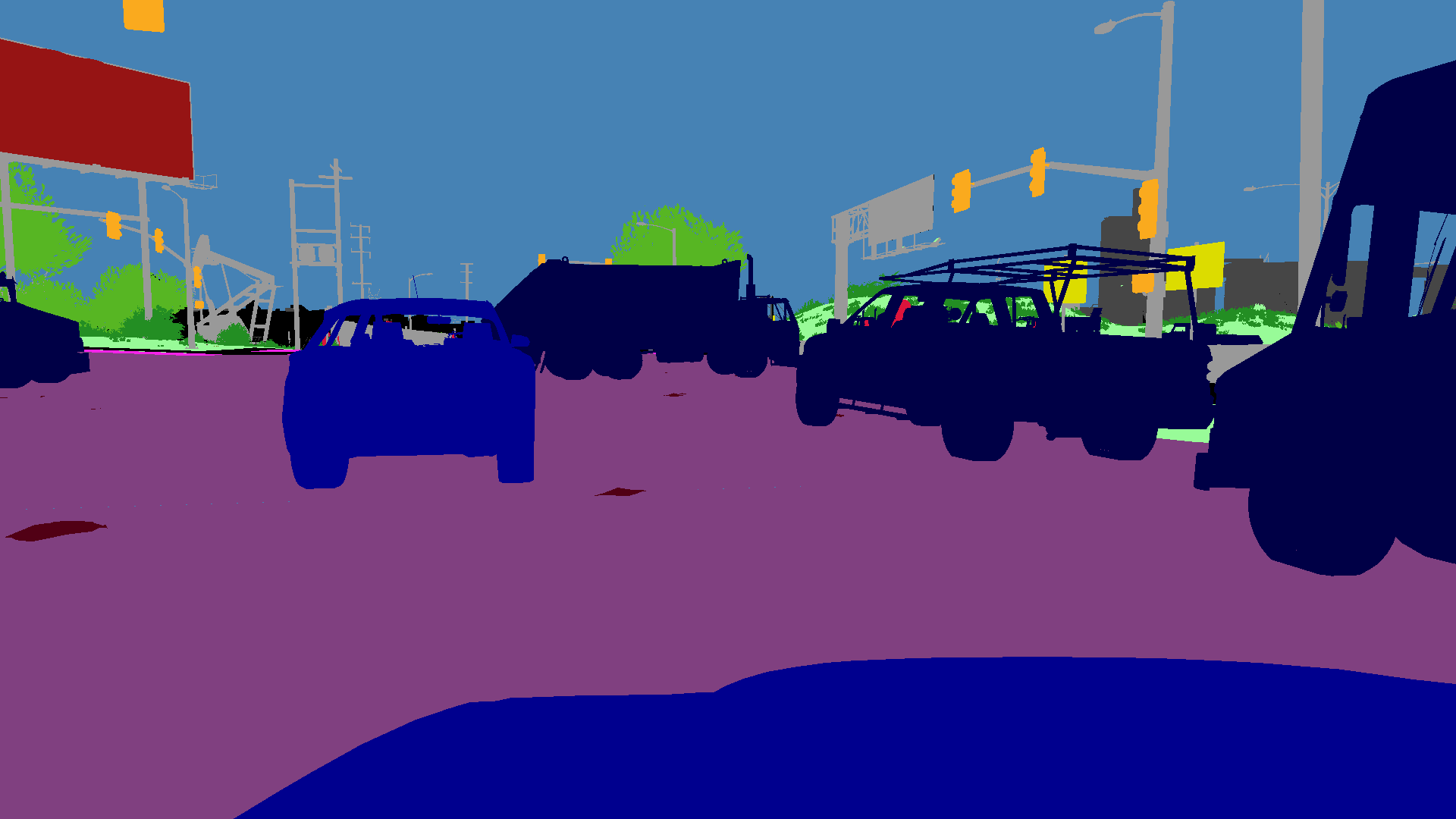}
    &
   \includegraphics[height=\szfi]{
    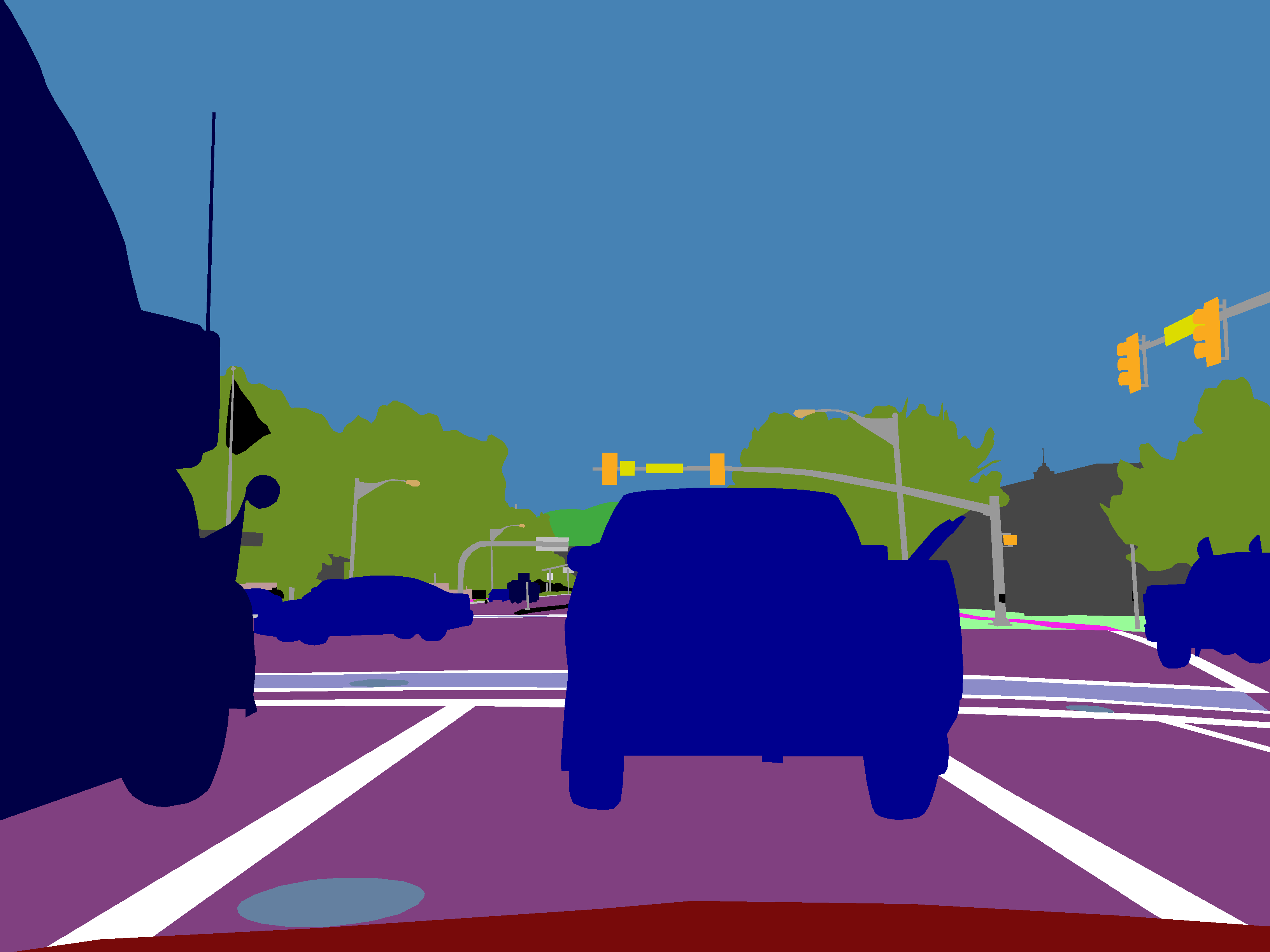}
    &
   \includegraphics[height=\szfi]{
     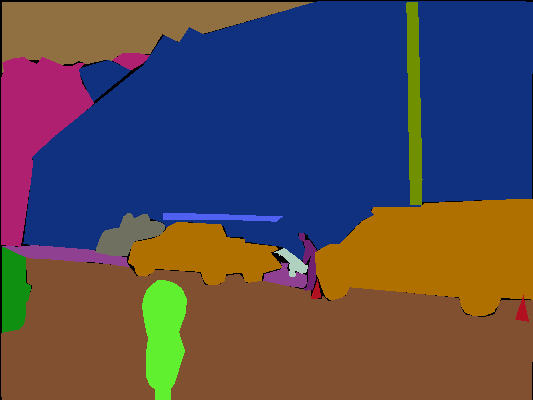}
  \mytabsep
  \end{tabular}
  \caption{
    We address dense-prediction models
    which can learn visual concepts
    from multiple datasets 
    with overlapping classes.
    Example: pickups are labeled as 
    class \lstinline{truck} 
    in VIPER 
    \cite{richter17iccv} 
    (left),
    class \lstinline{car} 
    in Vistas 
    \cite{neuhold17iccv} 
    (middle)
    and class \lstinline{van} 
    in Ade20k 
    \cite{zhou17cvpr} 
    (right).
    We resolve the class overlap  
    (\lstinline+Viper-truck+ vs 
     \lstinline+Vistas-car+ vs 
     \lstinline+Ade20k-van+)
     by learning on partial labels
     \cite{cour2011learning}.
  }
  \label{fig:intro}
\end{figure}

This paper addresses 
two important challenges 
in multi-domain semantic segmentation: 
i) training on datasets 
with inconsistent labels, and
ii) designing an experimental setup 
capable of learning 
hundreds of dense logits 
on megapixel resolution.
We contribute a novel method 
for training semantic segmentation
on datasets with discrepant granularity 
and overlapping classes
by leveraging partial labels \cite{cour2011learning}
and class-wise log-sum-prob loss.
Our method builds a universal taxonomy
such that each dataset-specific class
can be expressed as a union 
of one or more universal classes.
Our universal models
can be seamlessly trained and evaluated
on each individual dataset
since probabilities of dataset classes
correspond to sums of probabilities
of universal classes.

Our method outperforms two baselines 
which ignore overlapping classes, 
as well as a recent approach \cite{lambert20cvpr} 
based on partial relabeling 
towards a closed unified taxonomy.
Computational advantages of our method become decisive 
in large-scale multi-domain experiments 
where we achieve state-of-the-art performance
on the RVC 2020 benchmark collection and
the WildDash 2 benchmark.

\section{Related Work}

We consider efficient training on a collection 
of semantic segmentation datasets 
with discrepant granularities 
and overlapping classes. 
This requires a capability to learn 
fine-grained classes on coarse-grained labels. 
Thus, we review semantic segmentation,
efficient architectures,
learning with partial labels,  
and multi-domain dense prediction.


\subsection{Semantic Segmentation}

Deep convolutional models have spurred 
substantial progress in the 
field of semantic segmentation
\cite{long2015fully,chen2017deeplab,zhao2017pyramid,cheng20cvpr}.
Recent work improves recovery
of the spatial details
lost due to downsampling
\cite{chen2017deeplab}
and incraeses the receptive field
\cite{zhao2017pyramid}.
U-Net \cite{ronneberger2015u} 
recovers the fine details
by blending deep semantics 
with spatial information 
from the early layers. 
Further work notices that the upsampling path
requires much less capacity 
than the recognition backbone
\cite{lin2017feature, kreso17cvrsuad}.
Increased subsampling 
enlarges the receptive field,
improves efficiency and 
reduces the training footprint 
\cite{chao19iccv, kreso17cvrsuad}.
Further shrinking of the training footprint
can be achieved through
gradient checkpointing \cite{bulo18cvpr, kreso20tits}.
Multi-scale processing relaxes 
the requirements on recognition capacity 
and leads to competitive results 
with efficient backbones
\cite{orsic20pr, zhao18eccv}.
HRNet \cite{wang20pami} preserves fine details 
by sustaining high-resolution representations 
along the entire convolutional pipeline.
Its improved variant uses attention pooling
to promote pixels of the 
majority class \cite{yuan20eccv}.

\subsection{Efficient inference and training}

Several recent semantic segmentation architectures
aim at high accuracy
while supporting real-time inference
\cite{chao19iccv,nie20accv,orsic20pr,hong21arxiv,zhao18eccv}.
Most of these approaches can be trained 
at a fraction of the computational budget
required by widely used methods. 
Hence, efficient architectures 
may create new opportunities
while making our research more inclusive 
and environmentally acceptable \cite{schwartz20cacm}.


\subsection{Partial labels and log-sum-prob loss}
\label{ss:partial-labels}

Log-sum-prob loss allows to train
unions of disjoint predictions
by aggregating over classes or data.
Data-wise formulation has been used 
to decrease influence of noisy labels 
\cite{zhu2019improving}.
Their loss sums probabilities 
of all labels from the $3\times3$ neighborhood
in pixels near semantic borders.
On the other hand, our loss 
performs class-wise aggregation 
by summing probabilities of all universal classes
corresponding to the coarse-grained label.
Our motivation is to allow learning
over inconsistent taxonomies,
while they alleviate
inaccurate annotations at semantic boundaries.

Learning from partial labels 
considers examples labeled 
with a set of classes 
only one of which is correct.
The original work \cite{cour2011learning} 
assumes fairly stochastic probability
$P(\set Z|\vec x, y)$
of false labels $\set Z$
given the datum $\vec x$
and its true label $y$.
In our context, this relation is deterministic
when the dataset of the datum is known.
Different from \cite{cour2011learning}, 
we formulate learning with partial labels 
in a principled probabilistic framework
based on log-sum-prob loss. 

Recent work \cite{zhao20eccv} considers 
partial labels for 
multi-domain object detection.
Their experiments show that log-sum-prob loss
does not contribute to object detection
which is likely due to the sheer asymmetry
between positive and negative windows.
On the other hand, our experiments indicate
that log-sum-prob loss and partial labels
are an effective tool 
for cross-dataset semantic segmentation.

\subsection{Multi-domain dense prediction}

Multi-domain dense prediction involves 
training on multiple datasets 
with different taxonomies.
A simple baseline consists of 
a shared encoder and separate 
per-dataset decoders \cite{kalluri19iccv}.
This is similar to training 
on naive concatenation 
of individual datasets.
We find this to be suboptimal 
since repetition of classes
(eg.\ Vistas-bus vs Ade20k-bus)
increases the memory demands 
and decreases utility of predictions.
Instead, we prefer to train
on a shared universal taxonomy 
which contains much less classes
and supports inference on novel datasets.

A universal taxonomy can be implemented
as a hierarchy of logits 
for categories and classes \cite{liang2018dynamic,meletis18iv}.
This approach can gracefully handle
datasets with discrepant granularities.
However, on the downside,
it requires very complex training
and can not train on overlapping classes.


A universal taxonomy can also be implemented 
as a flat vector of universal logits
\cite{lambert20cvpr,zhao20eccv}.
The only such semantic segmentation approach
\cite{lambert20cvpr}
does not train on heterogeneous datasets.
Instead, they propose to adapt all datasets
towards a unified taxonomy of their own.
This requires relabeling all datasets 
which do not distinguish classes 
from their universal taxonomy,
and merging all classes 
which are more fine-grained
than the universal taxonomy.
It would be impractical 
for such unified taxonomy
to include all dataset classes,
since that would imply 
unacceptable relabelling effort.
Consequently, such design does not allow
evaluation on original datasets
nor simple introduction of datasets
with novel taxonomies.

As mentioned in \ref{ss:partial-labels} 
a recent multi-domain 
object-detection approach \cite{zhao20eccv}
considers candidate windows 
which may correspond either to the background 
or to any of the non-annotated classes.
Their attempt to solve that problem
with log-sum-prob loss was not successful.
However, classes of coarse-grained datasets
usually group visual concepts 
which are much much more similar to each other
than to a catch-all background class.
Consequently, log-sum-prob training
may have better a chance for success
in multi-domain semantic segmentation.


Different than all previous approaches,
we present a principled method
for cross-dataset and multi-domain training
of semantic segmentation models.
We assemble a universal flat taxonomy
from an almost unconstrained
collection of datasets.
The recovered universal taxonomy allows
to seamlessly train and evaluate
a universal probabilistic model
on unchanged individual datasets
by relying on partial labels 
and log-sum-prob loss.
The resulting models are suitable 
for universal semantic segmentation
in the wild.

\section{Multi-domain semantic segmentation}
\label{sec:method}

We present a principled method 
for simultaneous training 
on semantic segmentation datasets
with incompatible labeling policies.
Our method builds a universal taxonomy
which gathers and refines 
semantics of individual datasets
by expressing each dataset class 
as a union of universal classes.
Our models learn to predict universal classes
while fully supporting training and evaluation 
according to individual datasets.
We take care to contain 
the memory footprint of our models
since effective multi-domain learning 
requires large crops and large batches.

\subsection{Terminology and notation}

We consider semantic classes $c$ as 
sets of pixels in all possible images
and use set notation 
to express relations between them.
For example: Vistas-sky = CS-sky,
CS-road $\supset$ Vistas-manhole,
VIPER-truck $\cap$ Vistas-car = WD-pickup,
CS-road $\perp$ CS-car 
$\Rightarrow$ 
CS-road $\cap$ CS-car = $\emptyset$.
Note that WD and CS stand for 
WildDash 2 and Cityscapes, respectively.

A set of mutually disjoint 
semantic classes 
defines a flat taxonomy 
$\set S_d = \{ c_t \}, 
 c_i \perp c_j,
 \forall i,j\colon i\neq j$.
We note that a union of flat taxonomies
is not guaranteed to be a true taxonomy 
since members may have non-empty intersections.
For example, $ 
  {\set S}_\text{VIPER}
  \cup
  {\set S}_\text{Vistas}$ 
is not a true taxonomy since
VIPER-truck $\not\perp$ Vistas-car
(cf.~Fig.~\ref{fig:intro}).
  
A semantic segmentation dataset 
consists of images and 
corresponding dense labels: 
$\set D = \{(\vec x_k, \vec y_k) \}$.
The labels assign semantic classes $c \in \set S_d$.

\subsection{Baseline approach: naive concatenation}
\label{ss:concat}

We consider training a semantic segmentation model
on a compound dataset $\bigcup_d \set D_d$.
This can not be accomplished
by relying on individual taxonomies $\set S_d$, 
since that would either entail information loss
(eg.\ remapping Vistas-manhole to CS-road),
or require expensive relabeling
(eg.\ remapping some pixels
from CS-road to Vistas-manhole).
 
We propose a simple baseline approach
which creates a training taxonomy 
as a \emph{naive concatenation}
of all individual taxonomies $\set S_d$. 
Each training taxonomy is mapped
onto the corresponding subset 
of universal labels.
For instance, cars from Cityscapes 
are mapped to the class Cityscapes-car, 
while cars from Vistas 
are mapped to Vistas-car.
The predictions are produced
by a single softmax 
over $\sum_d|\set S_d|$ logits,
which allows training
with the cross-entropy loss.

However, the described approach 
does not produce a true taxonomy,
and thus leads to contention 
between overlapping logits
(e.g. CS-road vs Vistas-road). 
This wastes model capacity 
due to need for dataset recognition
instead of promoting generalization 
by principled dealing 
with ambiguous supervision. 
Additionally, the class replication 
increases the number of logits. 
This may hurt recognition of rare classes 
and exhaust GPU memory since the predictions 
are produced on a large resolution. 
Finally, inference has to deal with
starvation of overlapping logits
which we alleviate with
post-inference mappings
(cf.\ \ref{ss:univ-evaluation}).

\subsection{Baseline approach: partial merge}
\label{ss:partial-merge}

Naive concatenation can be improved
by merging classes which match exactly;
for example, Cityscapes-sky and Vistas-sky 
can be merged into the universal class sky.
Classes which can not be matched
in a 1:1 manner remain separate 
(e.g. Cityscapes-road, Vistas-road
and Vistas-marking).
This is similar to naive merge from \cite{lambert20cvpr},
however we merge according to semantics 
instead of symbolic names
and use post-inference mappings
as described in \ref{ss:univ-evaluation}.
This may reduce the amount of capacity
needed to distinguish between datasets
due to eliminated contention 
between the merged classes. 
Nevertheless, partial merge also 
fails to produce a true taxonomy 
since overlapping classes remain unresolved.
Hence, we encounter the same problems  
as presented in \ref{ss:concat}
although to a lesser extent.

\subsection{Creating a universal taxonomy}
\label{ss:universal}

We address disadvantages 
of the two baselines 
by building a flat universal taxonomy $\set U$
which will allow 
i) training on overlapping labels 
(cf.\ \ref{ss:nllplus}), and
ii) seamless dataset-specific 
(cf.\ \ref{ss:univ-evaluation}) or 
dataset-agnostic evaluation. 
The desired universal taxonomy $\set U$
should have the following properties.
First, its classes should encompass
the entire semantic range
of individual datasets:
$\bigcup_{\cls u\in\set U} \cls u = 
 \bigcup_d \set S_d$.
Second, universal classes should be disjoint:
$\forall\cls u, \cls u'\in\set U
 \colon 
 \cls u \perp \cls u'$.
Third, no universal class
may have a non-empty difference
towards an intersecting 
dataset-specific class:
$\forall 
 \cls u \in \set U,
 \cls c \in \bigcup_d \set S_d
 \colon
 (\cls u \perp \cls c) 
 \vee
 (\cls u \subseteq \cls c)$.
Such taxonomy will allow us to define mappings 
$m_{\set S_d} \colon \set S_d \to 2^{\set U}$
from dataset-specific classes
to subsets of the universal set.

We propose 
to recover the universal taxonomy $\set U$
by iteratively transforming
a multiset $\set M$ that contains
dataset specific classes from all datasets. 
We apply the rules following rules until
all of the classes are disjoint.
\begin{enumerate}
\itemsep0em
\item
  If two classes $c_i$ and $c_j$ match
  exactly, we merge them into a new
  class $c'$ and map $c_i$ and $c_j$ to $c'$.
  \\\emph{Example}: 
  Since WD-sky and CS-sky are equal, 
  we merge them into a class sky,
  and define new mappings:
    WD-sky $\mapsto$ sky,
    CS-sky $\mapsto$ sky.
\item
  If a class $c_i$ is 
  a superset of a class $c_j$,
  then  $c_i$ is
  mapped to $\cbr{\cls c_j, \cls c_i'}$
  where $\cls c_i'=\cls c_i\setminus \cls c_j$.
  \\\emph{Example}: 
  KITTI-car is a superset of car
  because it contains vans.
  We therefore add van as a new class and create
  a mapping 
  KITTI-car $\mapsto$ 
    \{car, van\}.
\item
  If two classes overlap,
  $(\cls c_i \not\perp \cls c_j)$
  $\wedge$ 
  $(\cls c_i \setminus \cls c_j \neq \emptyset)$
  $\wedge$ 
  $(\cls c_j \setminus \cls c_i \neq \emptyset)$,
  $c_i$ and $c_j$ are replaced with three
  new disjoint classes
  $\cls c_i' = \cls c_i\setminus \cls c_j$, 
  $\cls c_j' = \cls c_j\setminus \cls c_i$, 
  and $c' =\cls c_i \cap \cls c_j$. 
  Class $\cls c$ is mapped to 
  $\cbr{\cls c_i', \cls c'}$,
  while $k$ is mapped to
  $\cbr{\cls c_j', \cls c'}$.
  \\\emph{Example}: 
  VIPER-truck contains trucks and pickups
  while Ade20k-truck contains trucks and trailers.
  We therefore replace VIPER-truck and Ade20k-truck
  with truck, pickup and trailer and create
  the following mappings:
  VIPER-truck $\mapsto$ 
    \{truck, pickup\} and
  Ade20k-truck $\mapsto$ 
    \{truck, trailer\}
\end{enumerate}

\subsection{Mapping predictions to target classes}
\label{ss:univ-evaluation}

In practice, universal models 
are evaluated on particular
dataset-specific taxonomies.
We therefore design 
evaluation mappings
from dataset specific taxonomy $\set S_d$
to  the training class set $\set T$.
We extend $\set S_d$
with a void class. 
In our experiments, 
$\set T$ is either one of the two baselines
or the universal taxonomy $\set U$, 
while $\set S_d$ corresponds
to the particular dataset 
(e.g.\ $\set S_d = \set S_{\text{CamVid}}$).
We can create evaluation mappings
according to the following rules: 
\begin{enumerate}
\item 
 Each dataset-specific class is mapped to
 all training classes it overlaps with:
 $c^{S_d}\mapsto \{c^T \in \set T\colon c^{S_d} \not\perp c^T\}$.
\item 
 The void class is mapped to
 all training classes which 
 do not overlap with 
 any of the dataset-specific classes:
 $\mathrm{void}\mapsto \{c^T \in \set T\colon \forall c^{S_d}\in\set S_d, c^{S_d} \perp c^T\}$.
\end{enumerate}
We can use the created mappings
to calculate the classification score 
for each dataset-specific class in 
each pixel:
\begin{align}
  \mathrm{S}({\rvar y}^{S_d} = 
    \cls c | \vec x) 
  = 
  \sum_{\cls t \in m_{\set S_d}(c)}
      \P({\rvar y}^\set T
         = \cls t| \vec x) \; .
\end{align}

In the case of the universal taxonomy,
the above score corresponds to probability
since each universal class occurs in the
mapping set of
exactly one dataset-specific class.
In the case of the two baselines,
the score has to be renormalized 
because training classes may overlap 
with several dataset-specific classes,
e.g.\ $\mathrm{S}$(Vistas-pothole$|\vec x$) 
  = 
  $\mathrm{P}$(cs-road$|\vec x$) + $\mathrm{P}$(vistas-pothole$|\vec x$), 
  $\mathrm{P}$(Vistas-pothole$|\vec x$) = 
  $\mathrm{S}$(Vistas-pothole$|\vec x$) / 
    $\sum\mathrm{S}(c^{S_d}_j|\vec x)$.
Thus, creation of post-inference mappings 
for the two baselines 
requires almost the same effort
as the design of our universal taxonomy.

\subsection{Learning a universal model with partial labels}
\label{ss:nllplus}

We propose to learn cross-dataset models
on the universal taxonomy $\set U$
in order to properly address overlapping labels 
while fully exploiting all available supervision.
The resulting universal models will overcome 
all disadvantages of the two baselines.
In particular, they will be 
capable to exploit partial supervision
while being applicable in the wild.

Let us model the probability 
of universal classes
as per-pixel softmax 
over $\envert{\set{U}}$ logits
$\vec s$.  
Let the random variable $\rvar y$
denote a universal model
prediction at that pixel,
and let $\vec p$ denote the softmax output.
Then we can write:
$p_{u}
  = \softmax(\vec s)_{u}
  = \P(\rvar y
      \cls = u| \vec x)$.
Hence, the negative log-likelihood 
of the probability of a
dataset-specific class $c$
can be expressed as:
\begin{align}
  \label{eq:nllplusdef}
  \mathcal{L}^\text{NLL+}
    (\vec p, \cls c)
    =
  - \ln 
    \sum_{\cls u \in m_{\set S_d}(\vec c)}
    p_{u} \;.
\end{align}
Note that the sum of $p_u$ 
can be considered as probability 
since our universal classes 
are disjoint by design.
%
We denote the resulting 
log-sum-prob form as NLL+:
 negative log-likelihood 
 of aggregated probability.
This loss becomes the standard 
negative log-likelihood when 
$\envert{m_{\set S_d}(c)} = 1$.

In order to better understand 
the NLL+ loss (\ref{eq:nllplusdef}) 
we analyze its partial derivatives  
with respect to logits $\vec s$:
\begin{equation}
  \label{eq:nllplus}
  \frac{\partial \mathcal{L}}
      {\partial s_u} 
  =
  p_u
  - 
  \enbbracket{\cls u \in m_{\set S_d}(\cls c)}
  \frac{\exp{ s_u }}
    { \sum_{\cls k \in m_{\set S_d}(\cls c)} 
      \exp{ s_k } 
    }
    \;.
\end{equation}
We observe that gradients 
w.r.t.~incorrect logits 
are positive and the same 
as in the standard case,
since then the second term is zero.
Furthermore, the gradients 
w.r.t.~labeled logits
are always negative 
since the two terms 
have equal numerators,
while the second term 
has a smaller denominator.
If there are two correct logits,
the larger one will have an exponentially larger 
numerator of the second term,
while the denominator will be the same.
Hence the difference between 
the correct logits 
is likely to increase after the update.

This analysis indicates that NLL+
favours peaky sub-distributions
over correct logits: 
the model is compelled to pick 
only one among them.
In the limit, when one of the labeled logits
becomes much stronger than its peers,
its gradients become equal 
to the standard supervised case,
provided that the chosen logit is correct.
This suggests that partial labels
may help whenever there exists 
some learning signal 
that supports recognition of
individual universal classes.

Our GPU implementation represents 
each dataset mapping 
$m_{\set S_d}$ as an 
$\envert{\set S_d}\times\envert{\set U}$
binary matrix with unit column sums.
Probabilities of dataset-specific classes
are computed by multiplying this matrix 
with universal probabilities.

\subsection{Efficient multi-domain semantic segmentation}

Multi-domain semantic segmentation
is an extremely computationally intensive task
since the complexity of training
may be in the petaFLOP range.
The training footprint of the model 
is an important factor since 
the model performance improves 
with large batches, 
large training crops
and increased number of channels 
along the upsampling path.
These considerations are a compelling reason 
to prefer training on universal classes
instead of on less principled options
such as the two baselines presented before.

This discussion suggests that
efficient approaches may represent 
a sensible choice for such problems.
We therefore base our experiments
on pyramidal SwiftNet \cite{orsic20pr}.
All our models use three image scales
and produce dense logits 
at 8$\times$ subsampled resolution.
These logits are bilinearly 
upsampled to the full resolution.

We train our models with a compound loss
which corresponds to the product
of the boundary-aware (BA) factor
and the NLL+ loss (\ref{eq:nllplusdef}).
The BA modulation factor prioritizes 
poorly classified pixels
and pixels at boundaries
\cite{orsic20pr}:
\begin{align}
  \label{eq:loss}
  \mathcal{L}^\text{BA}(\vec p, c) &= - 
   \alpha \cdot e^{\gamma (1 - \hat{p})}
   \ln \hat{p}, \;
 \text{where} \;
 \hat{p} = \!\!\!\!\!\!
   \sum_{u \in m_{\set S_d}(c)}
     \!\!\!\!\!p_{u} \; .
\end{align}

We note that memory efficency
becomes extremely important
when training with 
several hundred logits
and large crops.
Even a humble inference 
becomes a challenge
in 2$\times$ upsampled Vistas images (24MPx)
used in multiscale inference
since then the logit tensor 
requires almost 20GB RAM.
We avoid caching 
multiple activations at full resolution
by implementing the boundary aware 
NLL+ loss (\ref{eq:loss})
as a layer with custom backprop.
This ensures a minimal increase 
of the training footprint 
with respect to the standard NLL formulation. 

\section{Experiments}

We present semantic segmentation experiments
on models based on pyramidal SwiftNet \cite{orsic20pr}.
Most experiments are based on ResNet-18 \cite{he16cvpr}, 
except our RVC models
which use DenseNet-161 \cite{huang19pami}.
All our models are pre-trained on ImageNet.
We train by optimizing 
the BA loss (\ref{eq:loss})
(either NLL or NLL+) with Adam.
We attenuate the learning rate
from $5\cdot10^{-4}$ to $6\cdot10^{-6}$
by cosine annealing.

We train our validation models 
by early stopping with respect to 
the average mIoU across all datasets.
Most of our models may emit predictions 
which cannot be mapped 
to any of the evaluation classes. 
For instance, Vistas-ego-vehicle 
is not labeled on Cityscapes.
We map such predictions to void
and count them as false negatives 
during calculation of the mIoU score.

We train on random square crops
which we augment through
resizing from 0.5$\times$ to 2$\times$
and horizontal flipping.
The default crop size is 768.
Our mini-batches favor images 
with multiple class instances
and rare classes
\cite{bulo18cvpr}.
We also reweight the images
so that the sum of image weights
is equal for all of the datasets.
This encourages fair representation
of datasets and classes during training.

\subsection{Comparison with two baselines}
\label{ssec:exp-baseline}

We compare our universal taxonomy 
with two baselines: 
naive concatenation (cf.\ \ref{ss:concat})
and partial merge (cf.\ \ref{ss:partial-merge}).
%
Both baselines require 
complex post-inference mapping 
before evaluation on particular datasets.
For example joint training
on Cityscapes and Vistas requires:
\{cs-sign, vistas-sign-front, 
  vistas-sign-back, vistas-sign-frame\} 
$\mapsto$ CS-sign.
Some universal logits will be mapped
to more than one evaluation class, eg:
\{cs-road, vistas-zebra\} 
  $\mapsto$ Vistas-zebra
  and
  \{cs-road, vistas-pothole\} 
  $\mapsto$ Vistas-pothole.
The merged training classes 
are 1:1 mapped 
to evaluation classes, eg;
sky $\mapsto$ CS-sky and 
sky $\mapsto$ Vistas-sky.

We consider joint training 
on two pairs of road-driving datasets 
with incompatible taxonomies. 
We train for 100 epochs with batch size 18 
which is the maximum that can fit 
into one Tesla V100
for the naive concatenation models.
Our first experiment considers Cityscapes and Vistas. 
In this case, our universal taxonomy 
(cf.\ \ref{ss:universal}) 
corresponds to the Vistas taxonomy 
since it is a strict refinement of Cityscapes. 
Our universal taxonomy has 65 classes,
which is less than the naive union (93)
and the partial merge (72).

Our second experiment considers 
joint training on Vistas and WildDash 2 
(WD2) \cite{zendel18eccv}.
We form WD2 minival 
by collecting the first 572 images
and use all remaining images as WD2 minitrain.
This setup is different from the first one 
since WD2 is much more diverse than Cityscapes. 
Hence, the baselines are going to have
a harder time to recognize particular datasets 
according to the camera type or location.
Additionally, WD2 has more classes than Cityscapes
and a finer granularity of car types than Vistas.
Consequently, our universal taxonomy has 67 classes
which is less than the naive concatenation (98), 
and the partial merge (76).

Table \ref{table:concatvsnllplus} shows 
that the proposed universal taxonomy 
succeeds to outperform both baselines
in spite of reduced capacity
due to less logits.
The advantage is especially evident 
in the case of joint training on WD2 and Vistas.
We hypothesize that this occurs
due to reduction of contention
between overlapping classes,
which allows the model capacity 
to be more efficiently exploited. 
Our universal taxonomy also outperforms
the baselines on City-Vistas
although the advantage is smaller 
than on WD2-Vistas.
We explain this as follows.
First, all Cityscapes images have been acquired 
across a contained geographical region 
with the same camera model. 
This makes dataset detection an easy task
and hence alleviates class contention
within the baselines.
Second, this setup entails 
a smaller difference between our taxonomy 
and the partial merge
(City-Vistas: 7 logits vs 
 WD2-Vistas: 9 logits).
We notice that our universal taxonomy 
outperforms individual training
(Cityscapes: 0.5pp, Vistas: 0.1pp \cite{orsic20pr}),
as well as that training on WD2-Vistas
slightly reduces Vistas performance.
This indicates that WildDash 2
requires more capacity than Cityscapes.
\begin{table}[htb]
  \centering
  \begin{tabular}{lcccccc}
    \multirow{2}{*}{Taxonomy} 
             && \multicolumn{2}{c}{(City-Vistas)} &&
                \multicolumn{2}{c}{(WD2-Vistas)}
    \\
             && City  & Vistas&& 
                WD2   & Vistas
    \\
    \toprule
    naive concat   && 76.8          & 44.6&&
                55.3          & 43.1
    \\
    partial merge && 77.1 & 44.5&&
                54.7 & 44.1
    \\
    universal (ours) && 77.0 & \textbf{44.9}&&
                \textbf{56.2} & \textbf{44.4}
    \mytabsep
\end{tabular}
\caption{Evaluation of joint training 
  on Cityscapes and Vistas (City-Vistas),
  as well as on WildDash 2 and Vistas (WD2-Vistas).
  Our universal taxonomy outperforms 
  both baselines.
 }
 \label{table:concatvsnllplus}
\end{table}


\subsection{Cross-dataset evaluation}
\label{ssec:exp-mseg}

We explore models 
from Table \ref{table:concatvsnllplus}
on novel road-driving datasets: 
CamVid test \cite{badrinarayanan17pami}, 
KITTI train \cite{geiger13ijrr}, 
BDD val \cite{yu18bdd} and 
IDD val \cite{varma19wacv}.
We also evaluate the City-Vistas model 
on WildDash V2 minival
and the Vistas-WD2 model on Cityscapes val.
As before, we design mappings 
from each of the six training taxonomies
to each evaluation taxonomy, 
and consider all unmapped logits 
as class void.

Tables \ref{table:camvid-KITTI} and 
\ref{table:camvid-KITTI-wd}
show
that our taxonomy 
outperforms naive concatenation 
and partial merge on most foreign datasets,
while in the remaining few cases
the difference is within variance.
Note that KITTI has only 200 images
and is very similar to Cityscapes.
As before, our contribution is greater
on WD2-Vistas than City-Vistas.
We connect that with Cityscapes peculiarities
which discourage generalization,
as discussed in \ref{ssec:exp-baseline}.

\begin{table}[hbt]
  \centering
  \begin{tabular}{lcccccc}
    Model & WD2 & CV & KITTI & BDD & IDD \\
    \toprule
    naive concat & 43.3 & 74.1 & 58.9 & 56.7 & 42.4
    \\
    partial merge & 43.8 & 73.9 & 59.4 & 57.0 & 43.0
    \\
    universal (ours) &  43.9 & \textbf{75.3} & \textbf{60.5} & \textbf{58.0} & 42.8
    \mytabsep
\end{tabular}
\caption{Cross-dataset evaluation 
  of joint training on Cityscapes and Vistas.
  We evaluate the three models from Table  \ref{table:concatvsnllplus}
  on WildDash 2 mini val, 
  CamVid test, KITTI, BDD val, and IDD val. 
 }
 \label{table:camvid-KITTI}
\end{table}

\begin{table}[htb]
  \centering
  \begin{tabular}{lcccccc}
    Model & CS & CV & KITTI & BDD & IDD \\
    \toprule
    naive concat & 69.0 & 72.7 & 53.6 & 56.1 & 41.6
    \\
    partial merge & 69.8 & 72.4 & 53.5 & 57.1 & 41.9
    \\
    universal (ours)  & \textbf{71.4} & \textbf{74.9} & 53.0  & \textbf{59.0} & \textbf{42.6}
    \mytabsep
\end{tabular}
\caption{Cross-dataset evaluation 
  of joint training on Vistas and WildDash 2.
  We evaluate the three models from Table  \ref{table:concatvsnllplus}
  on Cityscapes val, 
  CamVid test, KITTI, BDD val, and IDD val.  }
 \label{table:camvid-KITTI-wd}
\end{table}


\subsection{Comparison with partial manual relabeling}

\begin{table*}[htb]
  \centering
  \begin{tabular}{llccccccc}
    Evaluation protocol & Taxono
      & Ade20k & BDD & Cityscapes 
      & COCO & IDD & SUN RGBD & Vistas \\
    \toprule
    \multirow{2}{*}{Original}
      & Universal (ours)  & \textbf{31.0} & 58.5
      & 72.6 & \textbf{35.4} & \textbf{54.4} 
      & \textbf{41.7} & \textbf{39.1}
    \\
      & MSeg & 23.3 & \textbf{59.4} 
      & 72.6 & 30.3 & 42.6 & 40.2 & 26.1
    \\
    [0.5em]
    \multirow{2}{*}{MSeg}
      & Universal (ours)  & 34.5 & 58.5 & 72.6 & \textbf{36.3} & 53.0 & \textbf{41.7} & \textbf{45.4}
    \\
      & MSeg & 34.3 & \textbf{59.4} & 72.6 & 34.9 
      & \textbf{55.6} &  40.2 & 43.6
    \mytabsep
\end{tabular}
\caption{Multi-domain experiments with
  SNp-RN18 on the seven MSeg datasets
  \cite{lambert20cvpr}.
  We train a NLL+ model on original labels,
  and compare it with a NLL model
  which trains on manually relabeled images
  according to the MSeg taxonomy \cite{lambert20cvpr}. 
  Both models are evaluated on validation subsets 
  of Ade20k, BDD, Cityscapes, Coco, 
  IDD, SUN RGB-D and Vistas.
  We consider all unmapped logits 
  as class void.
 }
 \label{table:univ-mseg}
\end{table*}

We consider 7 datasets from different domains: 
Ade20k \cite{zhou17cvpr},
BDD \cite{yu18bdd}, 
Cityscapes \cite{cordts16cvpr}, 
Coco \cite{lin14eccv}, 
IDD \cite{varma19wacv},
SUN RGBD \cite{song15cvpr}
and Vistas  \cite{neuhold17iccv}.
Previous work has partially
relabeled these datasets in order to make
them fully compatible with a custom 
unified taxonomy called MSeg \cite{lambert20cvpr}.
However, they had to omit 100 classes in 
order to contain the relabeling effort.
We explore this trade-off by
comparing identical models trained with:
i) our universal taxonomy, original datasets
  and 294-way NLL+ loss,
and ii) MSeg taxonomy, relabeled datasets,
and 194-way NLL loss.

We resize each training image 
so that its smaller side is 1080 pixels, 
sample random square crops between 
256$\times$256 and 1024$\times$1024,
and resize them to 512$\times$512.
We train both models on a single Tesla V100 
for 20 epochs with batch size 20.
We validate the two models according 
to the following two protocols.
The MSeg protocol only evaluates the 194 classes 
from the MSeg taxonomy \cite{lambert20cvpr},
while the original protocols 
evaluate all classes.

The top section of Table \ref{table:univ-mseg}
presents evaluation 
according to the original protocol. 
Our approach prevails on most datasets
due to capability to predict all 294 classes. 
The bottom section of the table shows that
training with NLL+ remains competitive
even when we evaluate only on the 194 classes
covered by the MSeg taxonomy. 
This suggests that our universal taxonomy 
represents a flexible alternative 
to custom taxonomies,
especially in view of adding new datasets 
to the training collection 
(cf.~\ref{ssec:exp-rvc}). 


\subsection{Recognition of unlabeled concepts}

This experiment explores whether our method 
is able to recognize unlabeled concepts.
We partition Cityscapes train into two splits
with approximately equal number of images
and similar overall distribution of classes.
The first split gathers 
images from Aachen to Hanover.
It relabels all trucks, buses and cars 
into the class called 'four-wheels-vehicle'.
The second split gathers 
images from Jena to Zurich and 
relabels bicycles, motorcycles and cars
into the class 'personal-vehicle'.
Both splits have 17 classes.
Buses, trucks, motorcycles and bicycles
are labeled as standalone classes
in only one of the two splits
(roughly, half of the images),
while cars are never labeled 
as a standalone class.
Our universal taxonomy 
corresponds to 19 Cityscapes classes.
The dataset class four-wheel-vehicle
maps to \{car, bus, truck\}.
The dataset class personal-vehicle
maps to \{car, bicycle, motorcycle\}.
All other dataset classes map to themselves.

We validate four models 
for 19-way dense prediction.
The baseline model ignores
all pixels of aggregate classes
and uses the standard NLL loss.
Two models train on aggregate classes
through our NLL+ loss \eqref{eq:loss} 
and its variant which replaces the sum with max.
The oracle uses the original Cityscapes train.
We train all four models for 250 epochs
with batch size 14 on GTX1080 
while oversampling trains.

Table \ref{table:novelconcepts} reveals 
that NLL+ performs much closer
to the oracle than to the NLL baseline.
There is a substantial advantage 
for classes with only half non-ambiguous labels
(bus, truck, motorcycle, bicycle),
and especially on cars 
which are recognized as a novel concept
at the intersection of two training labels.
NLL-max \cite{zhao20eccv}
performs only slightly better 
than the NLL baseline,
while NLL+ appears as a method of choice
in presence of class overlap.
The two baselines succeed to detect cars
due to post-inference mapping, 
however NLL+ outperforms them substantially.

\begin{table}[htb]
\begin{center}
\begin{tabular}{l@{\quad}
    r@{\quad}r@{\quad}r@{\quad}
    r@{\quad}r@{\quad}r}
  Model & \faCar & \faBus & \faTruck & 
      \faMotorcycle & \faBicycle & mIoU
  \\
  \toprule
  NLL baseline &
    0 & 54.2 & 43.9 & 32.5 & 60.3 & 58.7
  \\
  NLL-max  &
    0    &  9.6 & 40.8 &  1.6 & 75.7 & 61.8
  \\
  naive concat & 91.1 & 61.4 & 42.2  & 39.0 & 72.3 & 67.6
  \\
  partial merge & 92.4 & 54.3 & 55.1 & 39.0 & 74.3 & 71.4
  \\
  NLL+  &
    \textbf{93.6} & \textbf{73.3} & \textbf{66.6} & \textbf{46.4} & 75.4 & \textbf{74.3}
  \\
  \midrule
  oracle  &
    94.4 & 82.9 & 72.9 & 62.2 & 76.5 & 76.2
  \mytabsep
\end{tabular}
\caption{Training on relabeled Cityscapes:
  one split aggregates cars, buses and trucks,
  while the other aggregates
  cars, bicycles and motorcycles.
  We validate NLL+ training, its NLL-max variant, 
  and the NLL baseline 
  which ignores all aggregate labels.
}
\label{table:novelconcepts}
\end{center}
\end{table}


\subsection{Large-scale multi-domain challenge}
\label{ssec:exp-rvc}

We present a large-scale multi-domain experiment 
on a benchmark collection proposed at 
the Robust Vision Challenge RVC 2020.
The challenge requires submitting 
a single semantic segmentation model
with less than 300 logits
to seven benchmarks:
Ade20k, 
Cityscapes, 
KITTI, 
Vistas, 
Scannet \cite{dai17cvpr}, 
Viper,  
and WildDash 2. 

\newcommand{\myheight}{2.6cm}
\begin{figure*}
    \centering
    \includegraphics[height=\myheight]{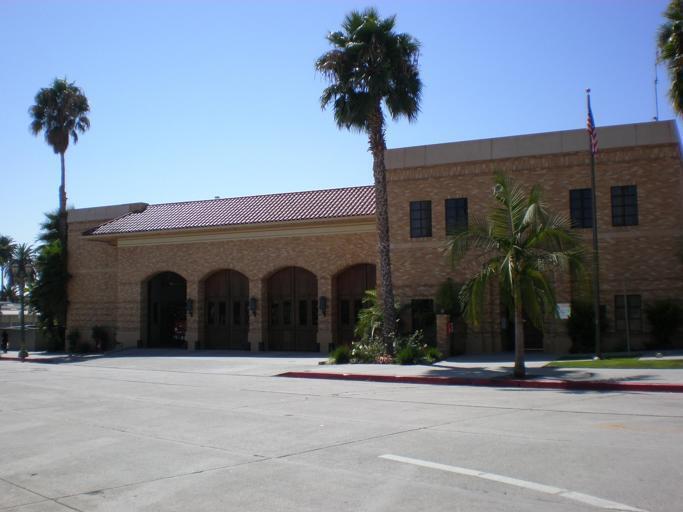}
    \includegraphics[height=\myheight]{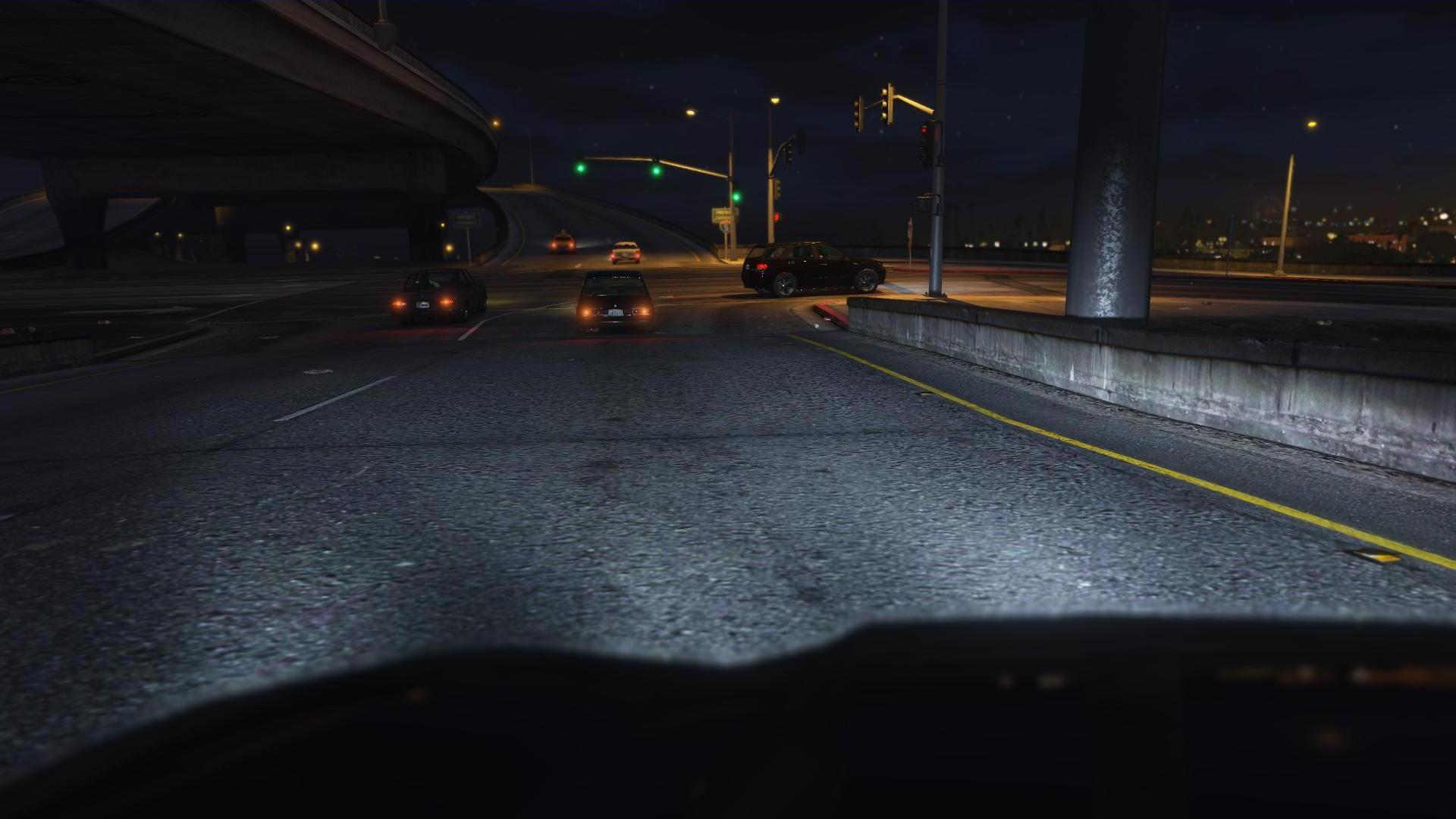}
    \includegraphics[height=\myheight]{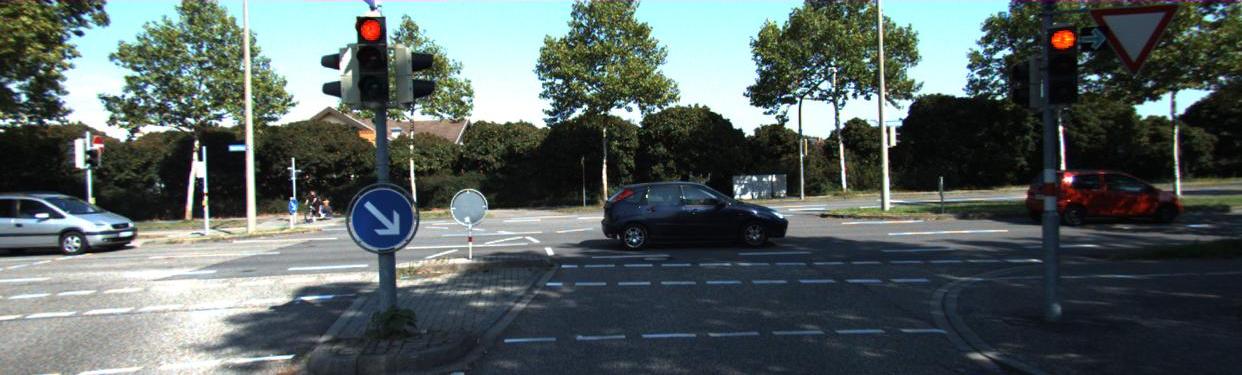}
    \\
    \includegraphics[height=\myheight]{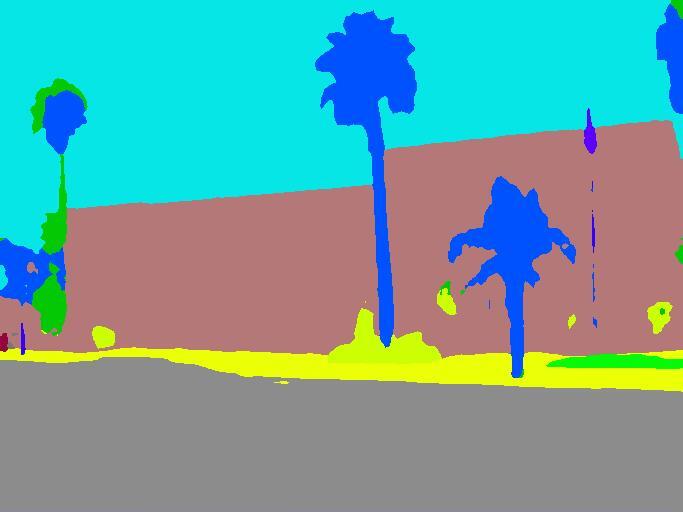}
    \includegraphics[height=\myheight]{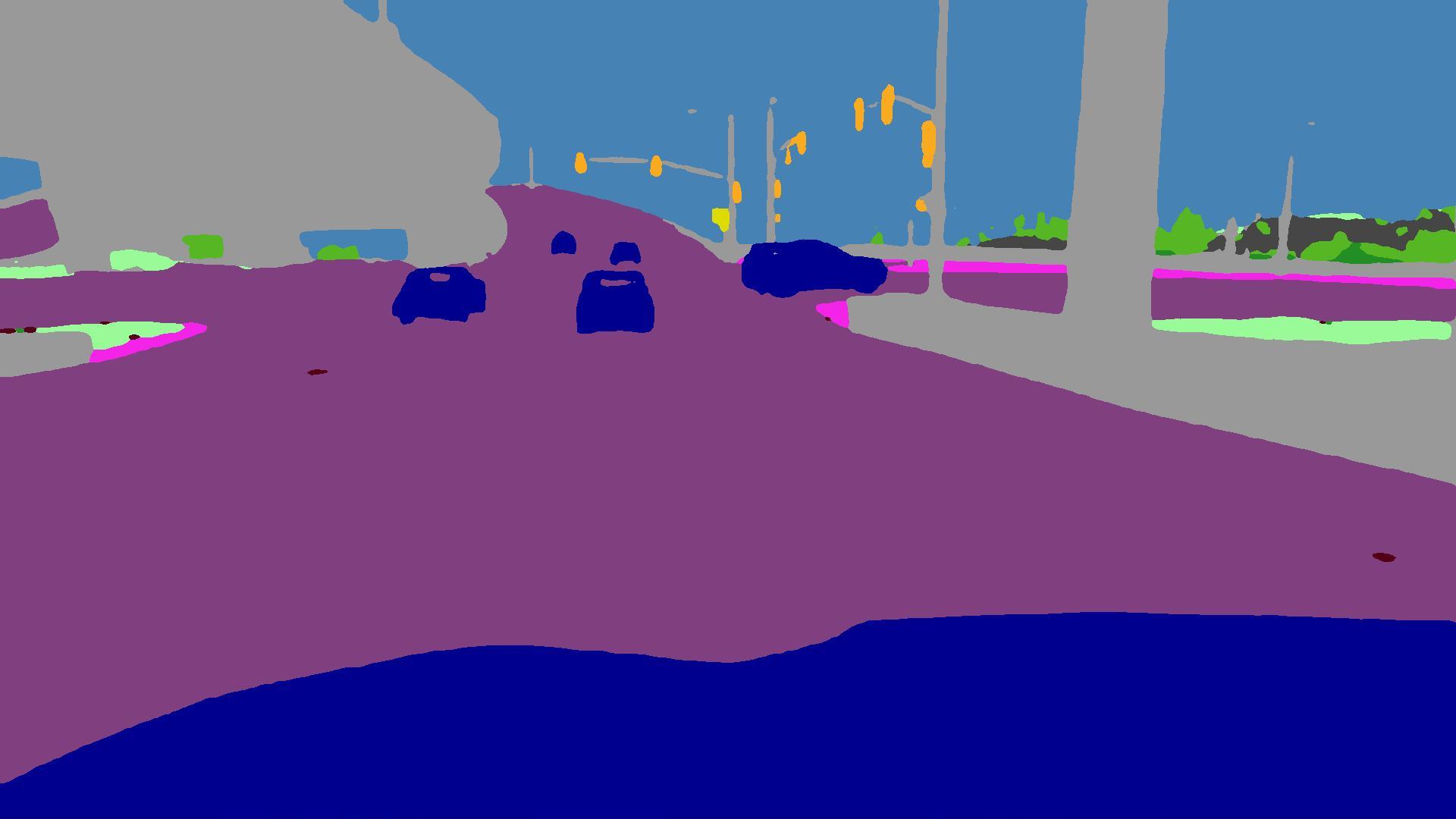}
    \includegraphics[height=\myheight]{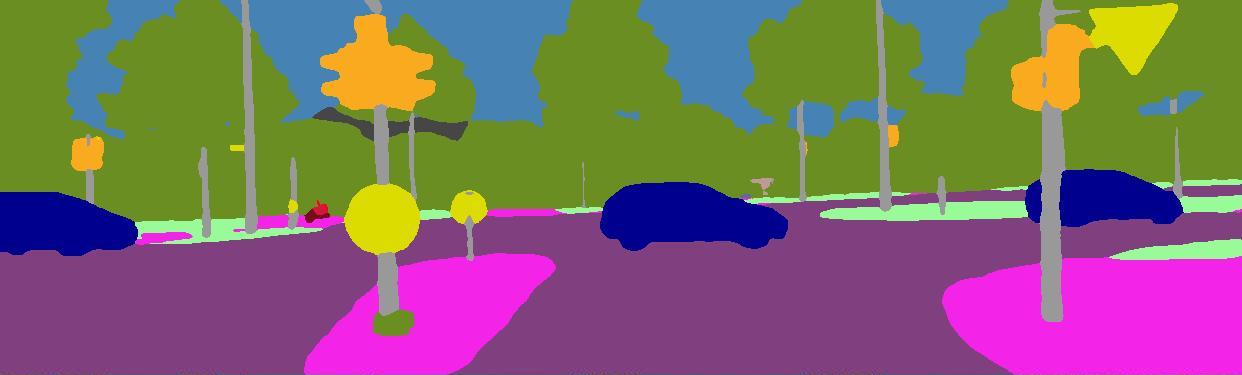}
    \\
    \includegraphics[height=\myheight]{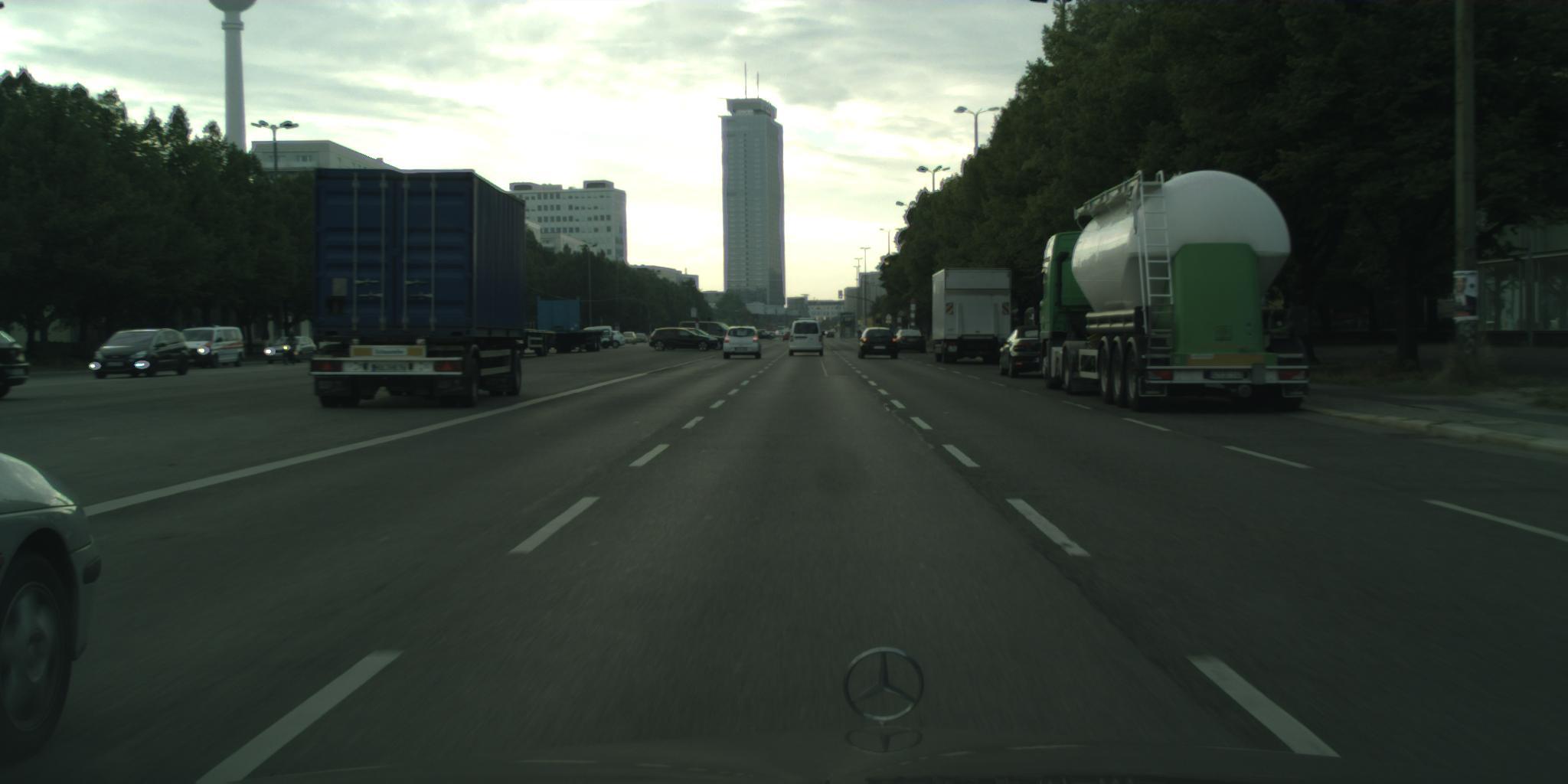}
    \includegraphics[height=\myheight]{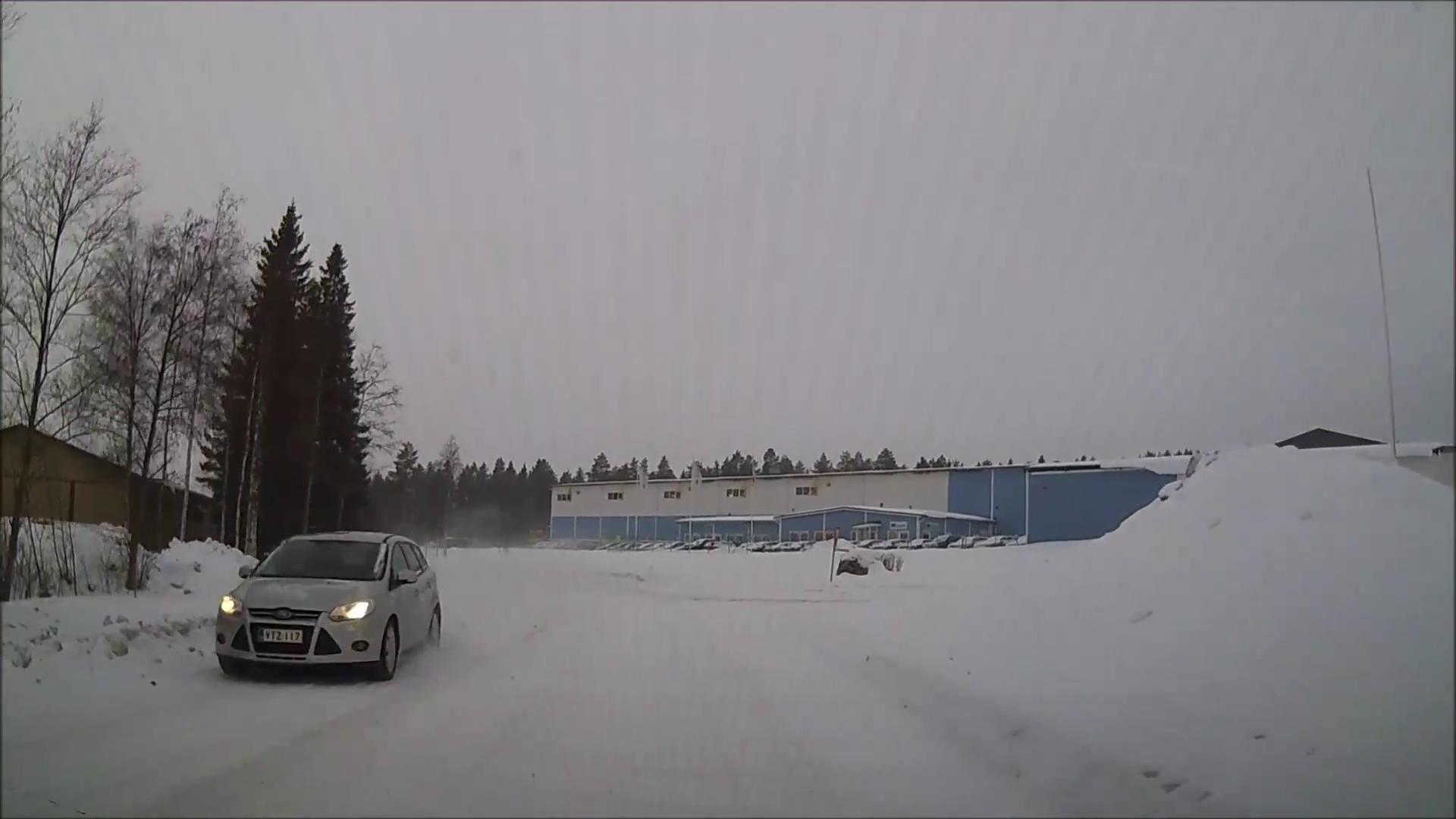}
    \includegraphics[height=\myheight]{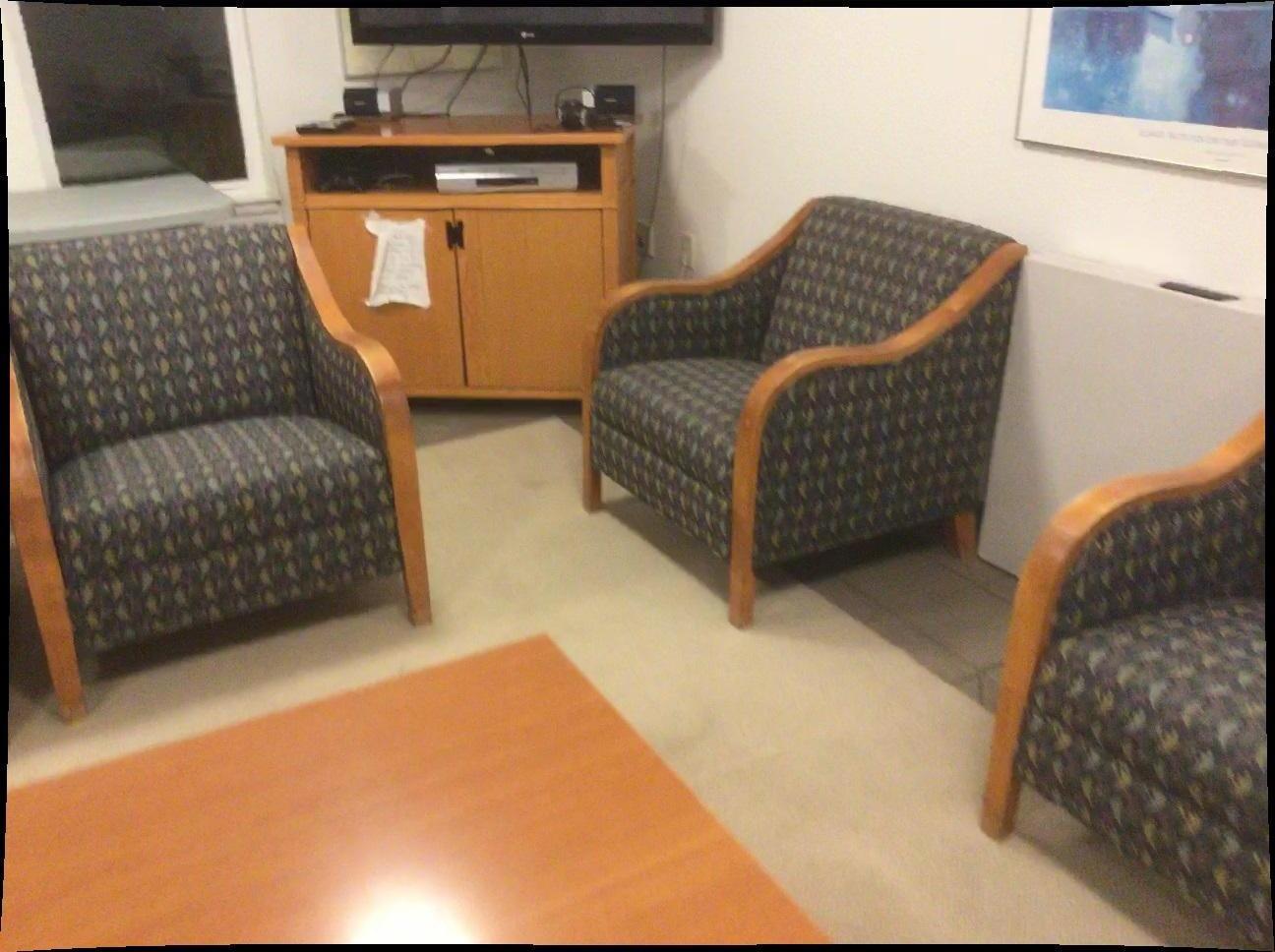}
    \includegraphics[height=\myheight]{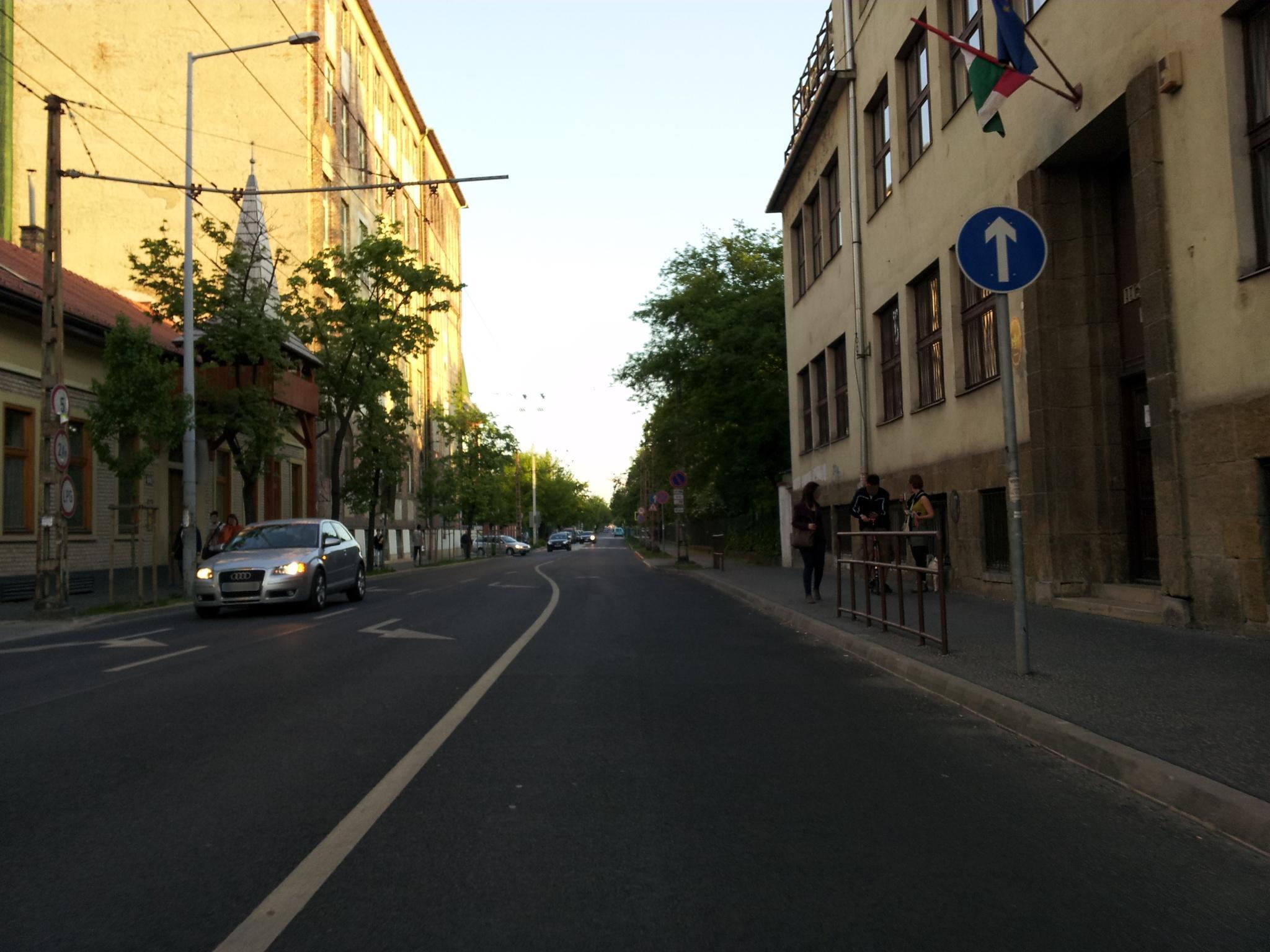}
    \\
    \includegraphics[height=\myheight]{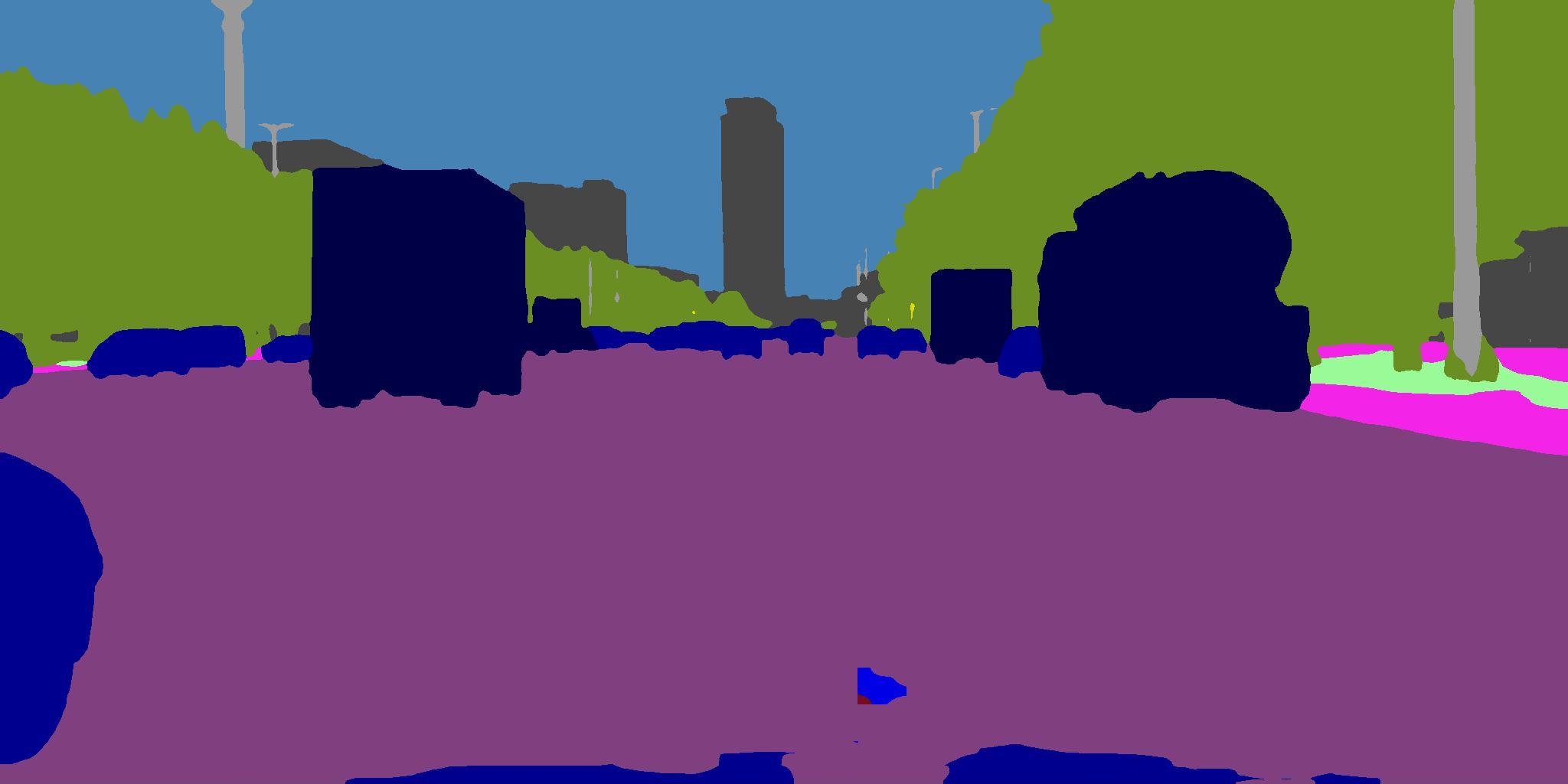}
    \includegraphics[height=\myheight]{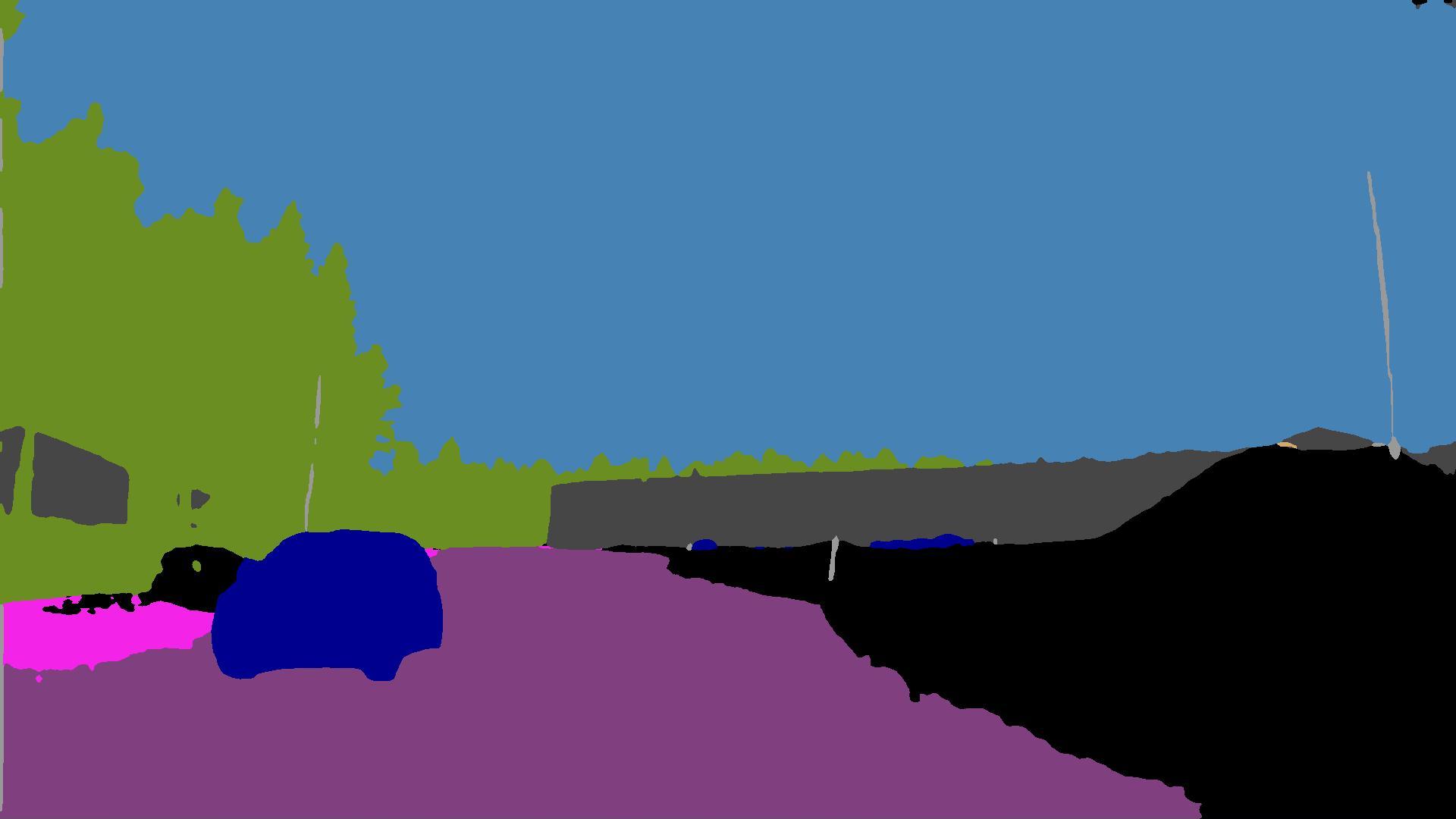}
    \includegraphics[height=\myheight]{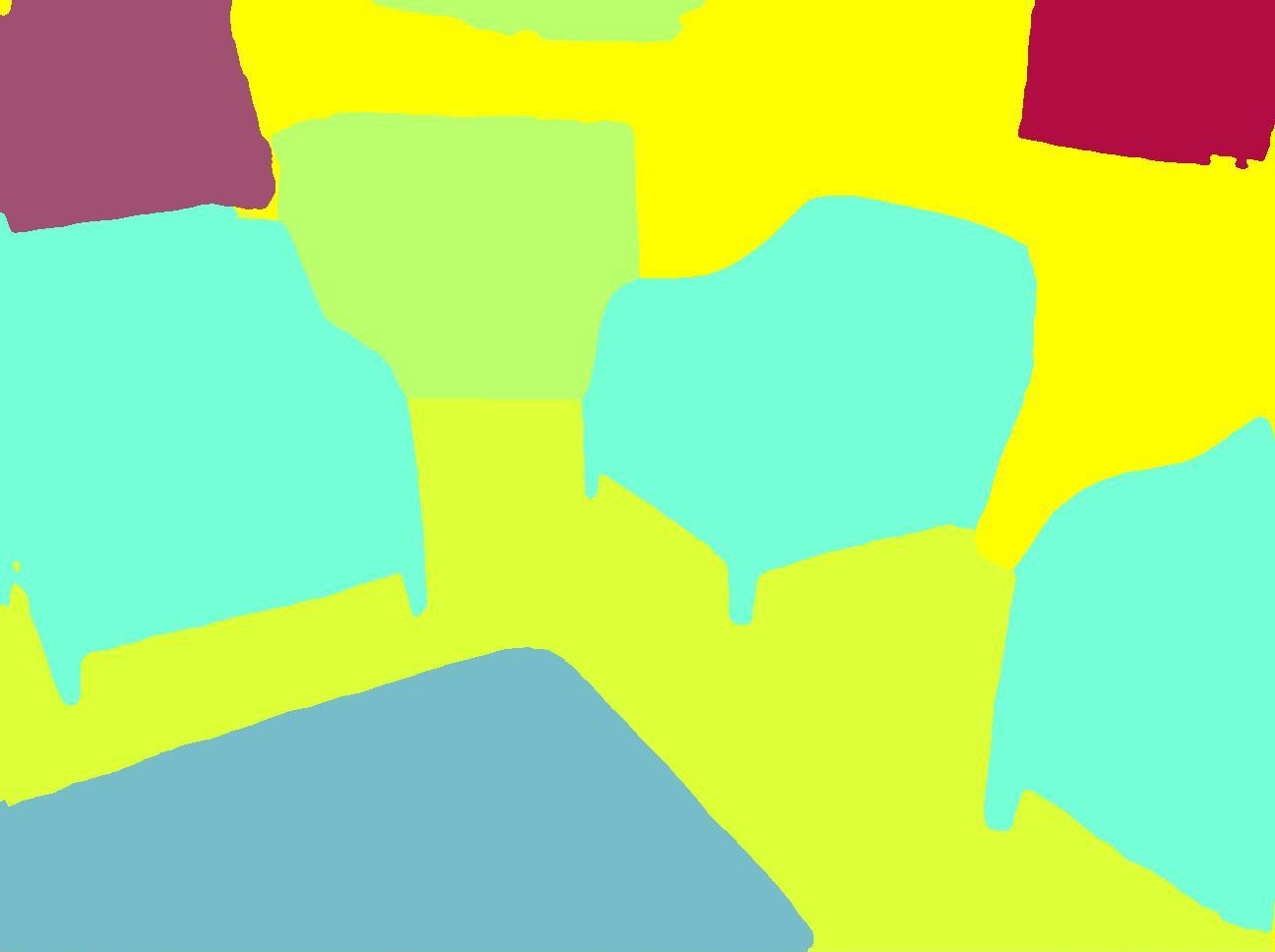}
    \includegraphics[height=\myheight]{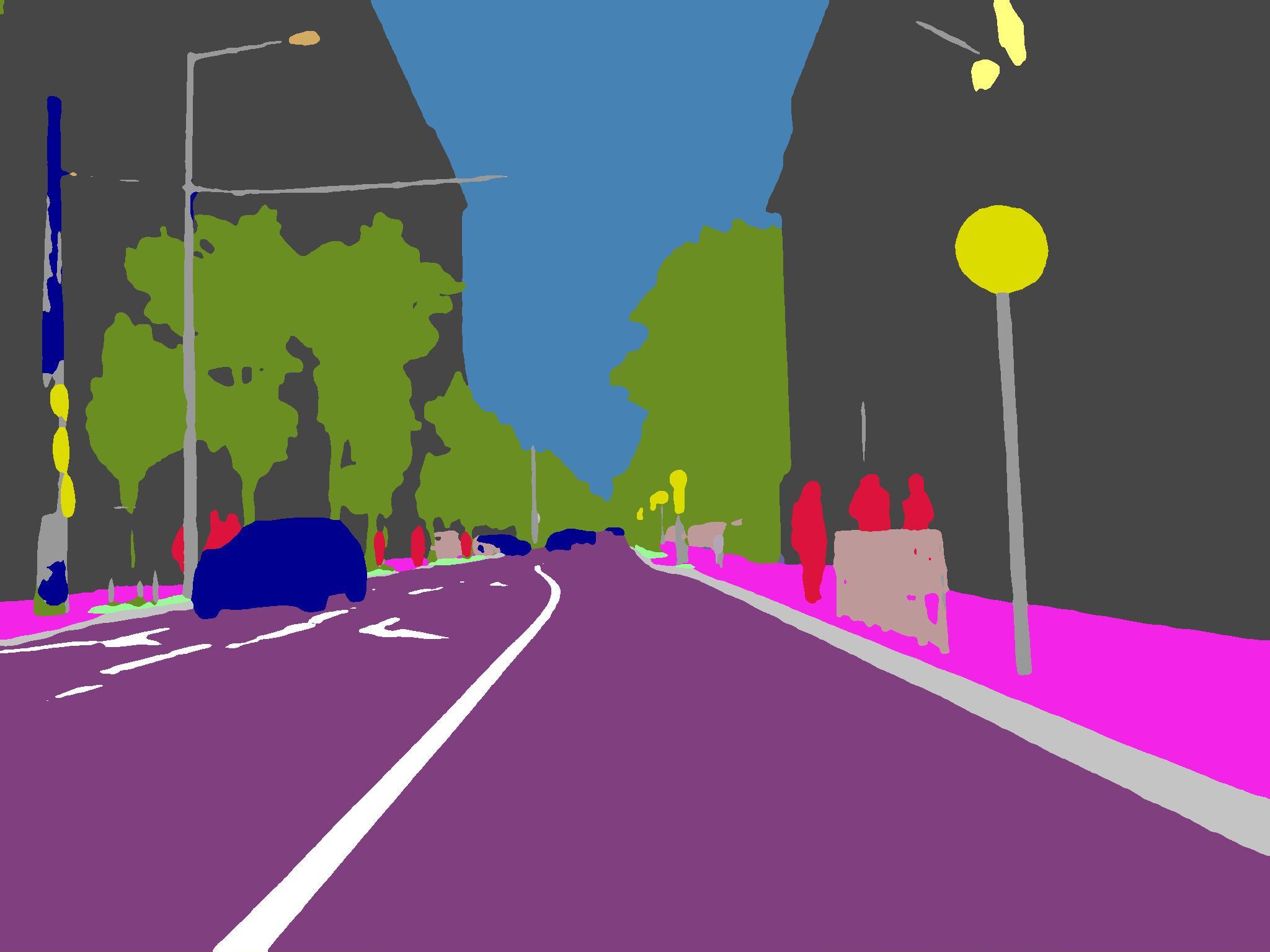}
     \\
    \caption{Qualitative performance of our SNp-DN161 submission to RVC 2020.
    Rows 1 and 3 show input images 
    while rows 2 and 4 show the model predictions. 
    Images belong to (top to bottom, left to right) ADE20k, Viper, KITTI, Cityscapes, WildDash, ScanNet and Vistas.}
    \label{fig:rvc}
\end{figure*}

We recover a taxonomy of 192 universal classes
according to the procedure from \ref{ss:universal}.
It turns out that Vistas labels 
vehicle windows as vehicles, 
while VIPER labels these pixels
with what is seen behind.
A consistent resolution of this issue
would effectively double 
the number of logits
since most universal classes 
would require addition of their twin class
(eg.~person vs person\_through\_glass).
Consequently, we introduce 
simplifying assumptions such as 
VIPER-car = Vistas-car and 
Vistas-car $\perp$ VIPER-person
in order to reduce 
the training footprint 
of the universal model.

We increase the model capacity by
using the DenseNet-161 \cite{huang19pami} backbone
and setting the upsampling width to 384 channels.
We decrease the memory footprint 
with custom backprop 
and gradient checkpointing
\cite{bulo18cvpr} 
in order to enable training on batches 
of 8 768$\times$768 crops per GPU.

We train on seven datasets 
of the RVC 2020 collection 
on 6 Tesla V100 32GB for 100 epochs.
We set the boundary modulation
to 1 (minimum) for all ScanNet crops
in order to alleviate noisy labels.
We evaluate on six scales 
and two horizontal flips.
Table \ref{tab:results} compares 
our model SNp\_DN161
with the two valid 
submissions to RVC 2020 \cite{orsic20arxiv}.
Fig.~\ref{fig:rvc} presents 
qualitative performance of our model.

\begin{table}[htb]
  \centering
  \begin{tabular}{l@{\;\;}
      c@{\;\;}c@{\;\;}c@{\;\;}
      c@{\;\;}c@{\;\;}c@{\;\;}c}
    Model & 
    ADE & CS & KIT & MV & SN & VIP & WD 
    \\
    \toprule
     MSeg1080 
     &
       \textbf{33.2} & \textbf{80.7} & 
       62.6 & 34.2 & 48.5 & 40.7 & 35.2 
     \\ 
    SN\_rn152pyrx8 &
      31.1 & 74.7 & 63.9 & 
      40.4&
      \textbf{54.6}& 
      62.5&
      45.4
    \\
    [0.5em]
    SNp-DN161  &  
      30.8 & 77.9 & \textbf{68.9} & \textbf{44.6} & 53.9 & \textbf{64.6} &\textbf{46.8} \\
    \mytabsep  
  \end{tabular}
  \caption{Performance evaluation 
    on RVC 2020 benchmarks
    Ade20k (ADE), Cityscapes (CS), 
    KITTI (KIT), Vistas (MV), 
    ScanNet (SN), VIPER (VIP) and
    WildDash 2 (WD).
    We compare the two valid
    RVC submissions (top)
    with our model (bottom).
  }
  \label{tab:results}
\end{table}

We note that our model significantly outperforms
the baseline submission MSeg1080\_RVC
\cite{lambert20cvpr}
which is unable to predict any classes
outside their closed unified taxonomy.
This suggests advantage of
open universal taxonomies such as ours,
similarly as in \ref{ssec:exp-mseg}.
We believe that multi-domain training
is instrumental for achieving
robust performance in the real world.
This is confirmed by the 
state-of-the-art performance
of our RVC model on the WildDash 2 benchmark
which is currently the only dataset
which specifically targets
expected points of failure 
of dense prediction models.

\section{Conclusion}

We have presented a principled method
for cross-dataset training 
of semantic segmentation models. 
Our method allows 
seamless training and evaluation 
on multi-domain datasets 
with discrepant granularity 
and overlapping classes.
Unlike simpler approaches 
such as multi-head prediction 
with shared features
and dataset detection,
our method recovers a universal taxonomy
which transcends semantics 
of individual datasets.
This allows transparent inference
on novel datasets and promises
thriving performance in the wild.

Our method expresses each dataset class 
as a union of disjoint universal classes. 
Hence, probabilities of dataset classes
correspond to sums of probabilities 
of universal classes.
This allows us to train a universal model
on dataset-specific labels
through negative log-likelihood 
of aggregated probability
which we refer to as  NLL+.
Such training can be viewed
as learning from partial labels
since we train fine-grained universal classes
on coarse-grained labels.

We compare our universal models
to two baselines which ignore
the overlap between
dataset-specific logits.
Our universal models 
outperform these two baselines
in within-dataset generalization
and cross-dataset validation.
Experiments on relabeled Cityscapes
indicate that our method is able to
transcend semantics of individual datasets
by learning an unlabeled universal concept
at the intersection of 
overlapping dataset classes.

Validation on the MSeg collection
shows that our method is competitive
to standard NLL training
on manually relabeled data.
Our advantage becomes substantial 
when evaluating on original datasets
where unified taxonomies have to drop classes
in order to alleviate the relabeling effort.

Our method achieves
the best aggregate performance
on the 7 benchmarks from 
the RVC 2020 challenge.
The corresponding universal taxonomy 
has only 192 classes,
which is a considerable improvement 
over 311 classes in the naive concatenation.
We see such multi-domain training
as a method of choice for machine vision 
in uncontrolled environments.
Indeed, our RVC models
set a new state of the art 
on WildDash 2 ---
the toughest road-driving benchmark, 
while outperforming 
other approaches by a large margin.

The recovered universal models 
can be used as a tool for evaluating 
automatically recovered taxonomies.
The authors would like 
to encourage further work 
on advanced evaluation datasets
that will reward cross-dataset training
and promote development of smarter
and more robust visual systems.
An ideal future dataset should combine 
rich taxonomies with domain shift,
anomalous objects, out-of-domain context, 
and adversarial input.
Other suitable avenues for future work include
automatic recovery of universal taxonomies,
extending the developed framework towards 
outlier detection and open-set recognition,
and exploring knowledge-transfer potential 
of multi-domain universal models.

{\small
\bibliographystyle{ieee_fullname}
\bibliography{egbib}
}

\newpage
\begin{appendices}
\appendix
\appendixpage
\addappheadtotoc
This supplement presents
additional validation experiments
and offers further qualitative analysis
of our submission to RVC 2020.
Additionally, it provides visualizations 
of our training and evaluation mappings
for the two baselines 
and our universal taxonomy.

We use Adam and attenuate 
the learning rate
from $5\cdot10^{-4}$ 
to $6\cdot10^{-6}$
by cosine annealing.
We use the largest batch size
that fits into the GPU memory, 
and train our models for 100 epochs.
All these experiments use 
single-scale evaluation. 

\section{Evaluation of the two baselines
in multi-domain experiments}

\begin{table*}[b]
  \centering
  \begin{tabular}{llccccccc}
    Evaluation protocol & Taxonomy 
      & Ade20k & BDD & Cityscapes 
      & COCO & IDD & SUN RGBD & Vistas \\
    \toprule
    \multirow{4}{*}{Original}
      & naive concatenation & 29.3 & 55.5
      & 68.7 & 30.8 & 52.1 & 41.1 & 35.3
    \\
      & partial merge & 30 & 58.4
      & 70.6 & 32.2 & \textbf{54.4} & 41.7 & 37.6
    \\
      & MSeg & 23.2 & 58.1  
      & 71.4 & 29 & 42.2 & 41.9 & 25.7
    \\
      & Universal (ours) & \textbf{31.3} & 56.5 
      & 71.2 & \textbf{33.7} & 53.1 & \textbf{42.5} & 37.9
    \\
    [0.5em]
    \multirow{4}{*}{MSeg}
      & naive concatenation & 32.3 & 55.5 & 68.7 & 31.5 & 53.9 & 41.1 & 41.4
    \\
      & partial merge & 33.3 & 58.4 & 70.6 & 32.9  
      & 51.8 & 41.7 & 43.9 
    \\
      & MSeg & 34.3 & 58.1  
      & 71.4 & 33.5 & 53.9 & 41.9 & 43.0 
    \\
      & Universal (ours) & 34.7 & 56.5  
      & 71.2 & \textbf{34.6} & 51.9 & \textbf{42.5}  & 43.8
    \\
    \mytabsep
\end{tabular}
\caption{Multi-domain experiments with
  SNp-RN18 on the seven MSeg datasets
  \cite{lambert20cvpr}.
  We train the baselines and 
  our universal model on original labels,
  and compare with a NLL model
  trained on manually relabeled images
  according to the MSeg taxonomy \cite{lambert20cvpr}. 
  Both models are evaluated on validation subsets 
  of Ade20k, BDD, Cityscapes, Coco, 
  IDD, SUN RGB-D and Vistas.
  We consider all unmapped logits 
  as class void.
 }
 \label{table:univ-mseg}
\end{table*}

The main paper evaluates
our universal taxonomy 
with a custom unified
taxonomy called MSeg \cite{lambert20cvpr}
in a multi-domain setup.
We repeat that experiment 
with our two baselines,
naive concatenation and 
partial merge,
which in this setup have 
469 and 307 classes.
In order to have 
a consistent training setup, 
we train all four models on a 
single Tesla V100 32GB GPU.
We use batch size 12, 
since that is the maximum 
for naive concatenation. 
We keep the rest
of the training protocol the same
as in the main paper.

The top section of Table \ref{table:univ-mseg}
presents evaluation 
according to the original protocol. 
Our universal model prevails on most datasets.
Partial merge performs comparably, 
since only 13 out of 307 classes
overlap with some other class
(our universal taxonomy has 294 classes).
The bottom section of the table shows 
a similar outcome when we evaluate 
only on the 194 MSeg classes. 
We note that even naive union 
outperforms manual relabeling 
when original evaluation protocols are used
since manual relabeling drops too many classes.

\section{Validation experiments}

  We validate several hyper-parameters
  on the Vistas dataset. 
  In all tables the columns correspond to
  training resolution (MPx),
  total batch size (BS),
  number of computational nodes
  used for training (nPU),
  and segmentation accuracy
  on Vistas val (mIoU).
 
\subsection{Validation of segmentation architectures}

Table \ref{table:backbone} shows performance
of various semantic segmentation architectures
on the validation subset of Vistas.
We present best approaches from the literature
and compare them with pyramidal SwiftNets
with different backbones.
All our models use checkpointed backbones
and were trained on 1 Tesla V100 32GB GPU.

\begin{table}[htb]
\begin{center}
\begin{tabular}{lccccc}
    Model & MPx & BS & nPU & mIoU \\
    \hline
    \hline
    Seamless \cite{porzi19cvpr} & 
      8 & 8 & 8 &50.4
    \\
    HN-OCR-W48 \cite{yuan20eccv} & 
      0.5 & 16 & 2 &50.8
    \\
    PDL-X71 \cite{cheng20cvpr} & 
      1 & 64 & 32 &55.4
    \\[0.2em]
    SNpyr-RN18 & 
      0.6 & 24 & 1 & 46.3 \\ 
    SNpyr-RN34 & 
      0.6 & 22 & 1  & 49.4 \\
    SNpyr-RN152 & 
      0.6 & 8  & 1 & 50.1\\
    SNpyr-DN121 & 
      0.6 & 22 & 1 & 49.5 \\
    SNpyr-DN161 & 
     0.6 & 17 & 1 & 52.2
    \mytabsep
\end{tabular}
\caption{Validation of semantic 
  segmentation architectures.
  All methods use single-scale evaluation,
  and train on Tesla V100 32GB GPUs
  except PDL which trains on TPUs.
  Pyramidal Swiftnets achieve
  competitive performance
  with respect to the state of the art, 
  while requiring much less
  computational resources. 
}
\label{table:backbone}
\end{center}
\end{table}

The table shows that pyramidal SwiftNets
offer competitive generalization performance
under modest computational requirements.
This makes them a good choice 
for large-scale experiments 
such as RVC 2020. 
Furthermore, larger models 
require significantly more time to
complete the training,
which may make a difference 
in large-scale multi-domain settings.

We observe that increasing the model capacity 
results in diminishing returns
(cf.~SNPyr-RN18 vs 
 SNPyr-RN34 vs SNPyr-RN152).
 SNpyr-DN161 outperforms SNPyr-RN152
in spite a slightly weaker backbone. 
We speculate that this may be 
due to larger batches.

\subsection{Validation of 
  multi-scale input}

Table \ref{table:pyr} compares 
a pyramidal model (SNPyr-DN161s3)
to its single-scale counterpart (SN-DN161s3).
Both models are based on DenseNet-161 
with 64$\times$ subsampled 
representation at the far end
which we achieve by splitting 
the 3rd dense block \cite{kreso20tits}. 
The single scale model (SN-DN161s3) uses 
an SPP module at the end of the
downsampling path at 64$\times$ 
subsampled resolution.

\begin{table}[htb]
\begin{center}
\begin{tabular}{lccccc}
    Model & MPx & BS & nPU & mIoU \\
    \hline
    \hline
    SNpyr-DN161s3 & 
      0.6 & 18 & 1 & \textbf{50.6} \\
    SN-DN161s3 & 
      0.6 & 25 & 1 & 48.1 \\
    \mytabsep
\end{tabular}
\caption{Validation of 
  pyramidal fusion.
  The multi-resolution model (SNpyr-DN161s3) 
  outperforms its single-scale counterpart
  (SN-DN161s3).
}
\label{table:pyr}
\end{center}
\end{table}

\subsection{Validation of 
  upsampling width}

Table \ref{table:upsample}
shows the influence of the upsampling
width on the segmentation performance. 
Wider upsampling improves the accuracy, 
although not significantly.

\begin{table}[htb]
\begin{center}
\begin{tabular}{lccccc}
    Model & MPx & BS & nPU & mIoU \\
    \hline
    \hline
    SNpyr-DN161s3 & 
      0.6 & 18 & 1 & 50.6 \\
    SNPyr-DN161s3-fat & 
      0.6 & 18 & 1 & \textbf{50.9}
    \mytabsep
\end{tabular}
\caption{Validation of the width 
  of the upsampling path.
  Wider upsampling path 
  achieves a slightly better accuracy.
}
\label{table:upsample}
\end{center}
\end{table}

\subsection{Validation of block splitting}
  
Table \ref{table:split}
examines the influence of 
dense-block splitting to
the generalization performance.
We observe that splitting 
the 3rd block deteriorates
the segmentation performance
when pyramidal organization is used.
We speculate that this is due
to too much subsampling within
the feature extractor.

\begin{table}[htb]
\begin{center}
\begin{tabular}{lccccc}
    Model & MPx & BS & nPU & mIoU \\
    \hline
    \hline
    SNpyr-DN161 & 
     0.6 & 17 & 1 & \textbf{52.2} \\
    SNpyr-DN161s3 & 
      0.6 & 18 & 1 & 50.6 \\
    \mytabsep
\end{tabular}
\caption{Validation of the extent
  of subsampling within a multi-scale
  feature extractor.
  Splitting the third block
  reduces segmentation accuracy. 
}
\label{table:split}
\end{center}
\end{table}

\section{Qualitative universal performance}

Figure \ref{fig:rvc} extends 
Figure 2 from the main paper by including 
qualitative universal performance 
of our SNp-DN161 model. 
The color map for the
universal label space 
combines the color maps
from Vistas and ADE20k.
We observe that the model is able 
to recognize refinements of
concepts from particular datasets.
For instance the model locates
road markings and sidewalk curbs
in images from Cityscapes, VIPER, 
KITTI and ADE20k.
It also succeeds to discriminate 
bushes and trees in KITTI and WildDash 2
although this distinction exists
only in the ADE20k taxonomy.

\begin{figure*}
    \centering
    \includegraphics[height=\myheight]{figs/ade/ADE_test_00001012_img.jpg}
    \includegraphics[height=\myheight]{figs/viper/007_00473_img.jpg}
    \includegraphics[height=\myheight]{figs/kitti/000009_10_img.jpg}
    \\
    \includegraphics[height=\myheight]{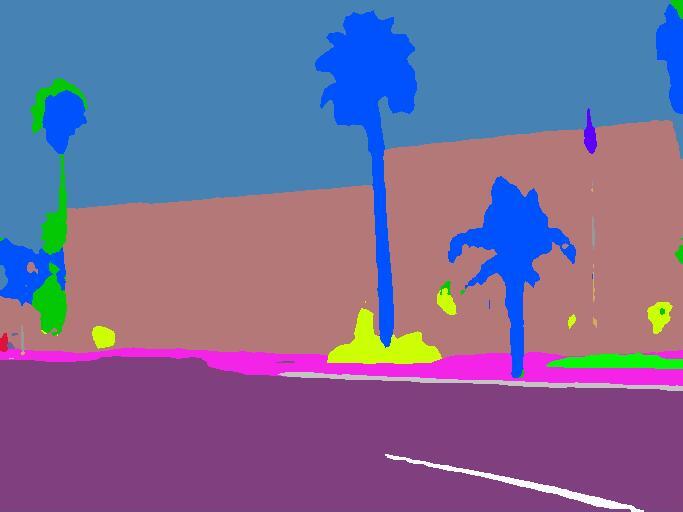}
    \includegraphics[height=\myheight]{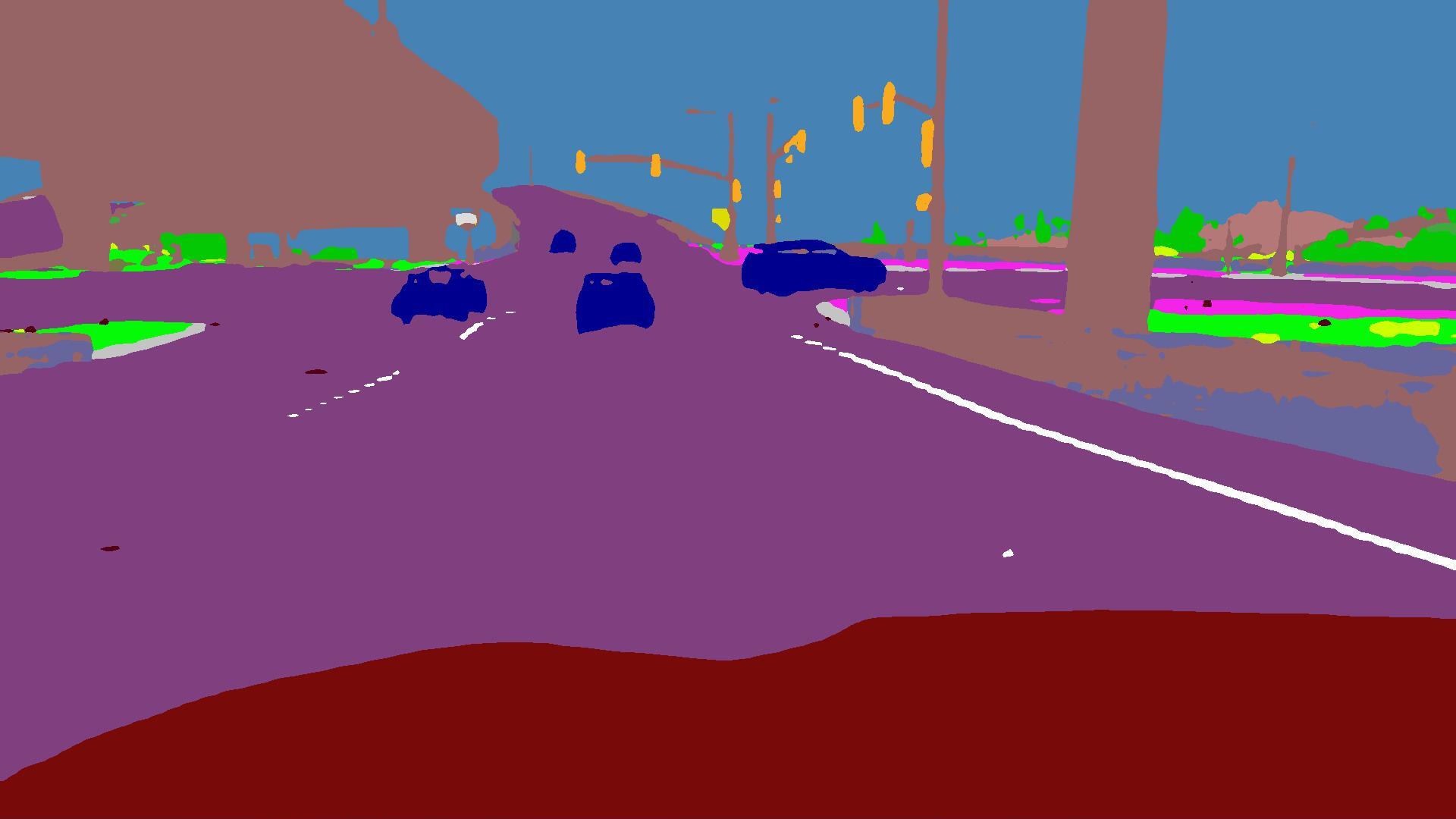}
    \includegraphics[height=\myheight]{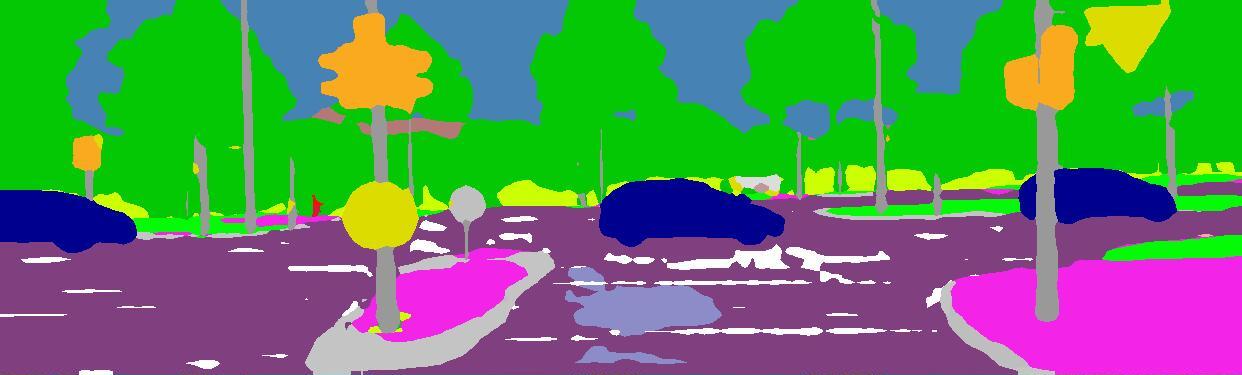}
    \\
    \includegraphics[height=\myheight]{figs/ade/ADE_test_00001012_pred.jpg}
    \includegraphics[height=\myheight]{figs/viper/007_00473_pred.jpg}
    \includegraphics[height=\myheight]{figs/kitti/000009_10_pred.jpg}
    \\
    \includegraphics[height=\myheight]{figs/city/berlin_000177_000019_leftImg8bit_img.jpg}
    \includegraphics[height=\myheight]{figs/wd/wd0093_100000_img.jpg}
    \includegraphics[height=\myheight]{figs/scannet/scene0781_00_000200_img.jpg}
    \includegraphics[height=\myheight]{figs/mvd/ZO6Xc8qwl0_S0G_ll8muhQ_img.jpg}
    \\
    \includegraphics[height=\myheight]{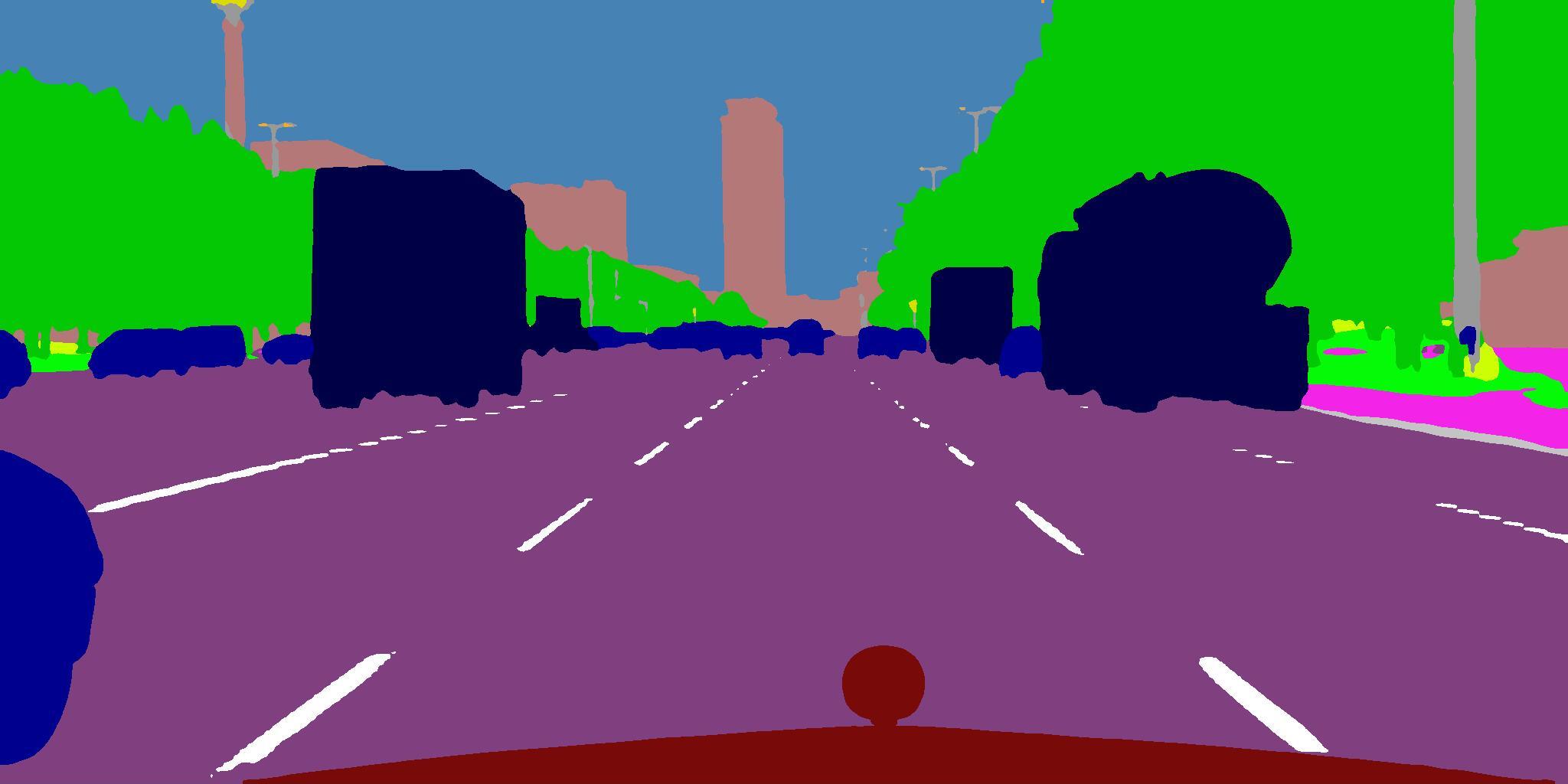}
    \includegraphics[height=\myheight]{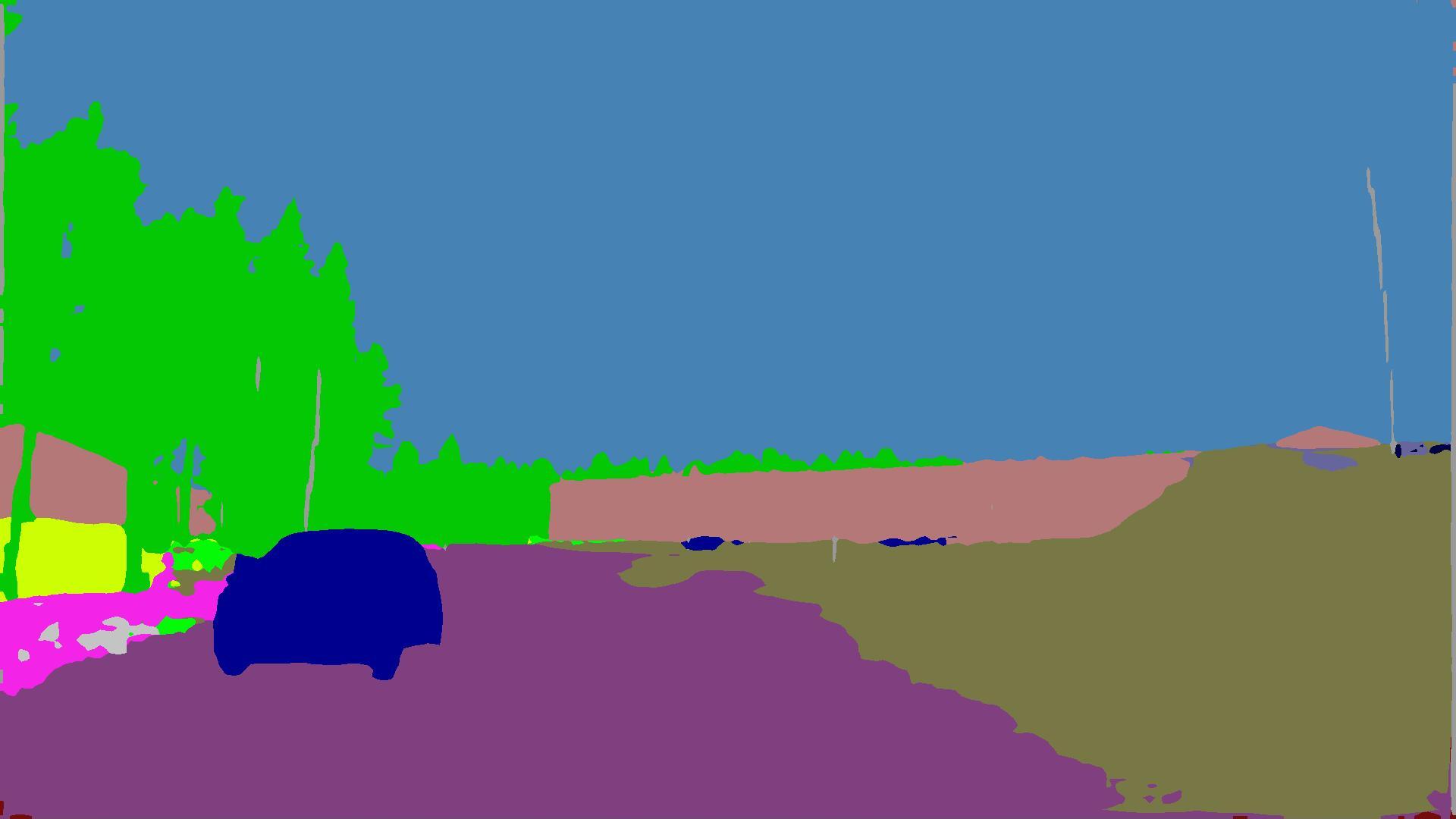}
    \includegraphics[height=\myheight]{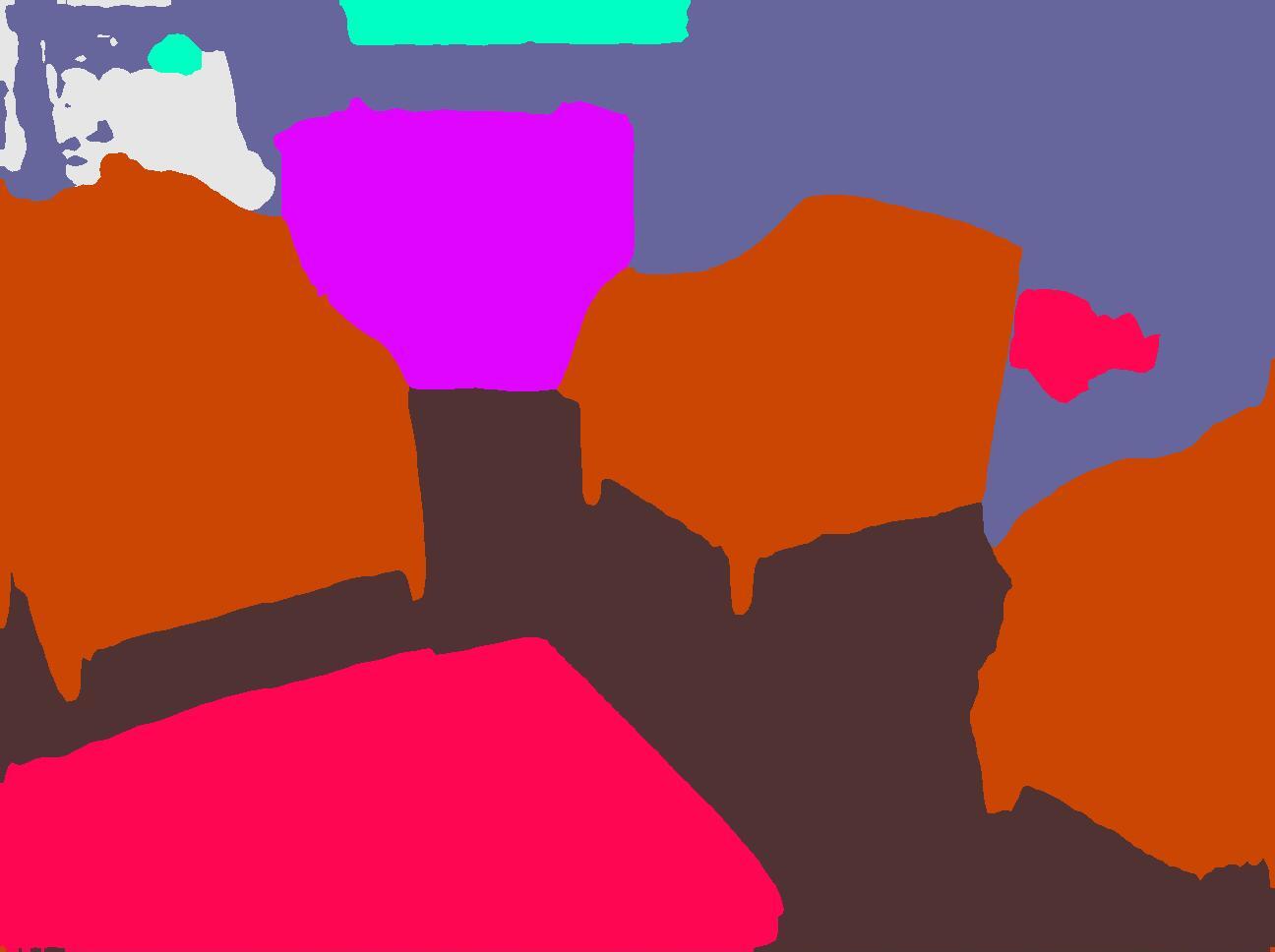}
    \includegraphics[height=\myheight]{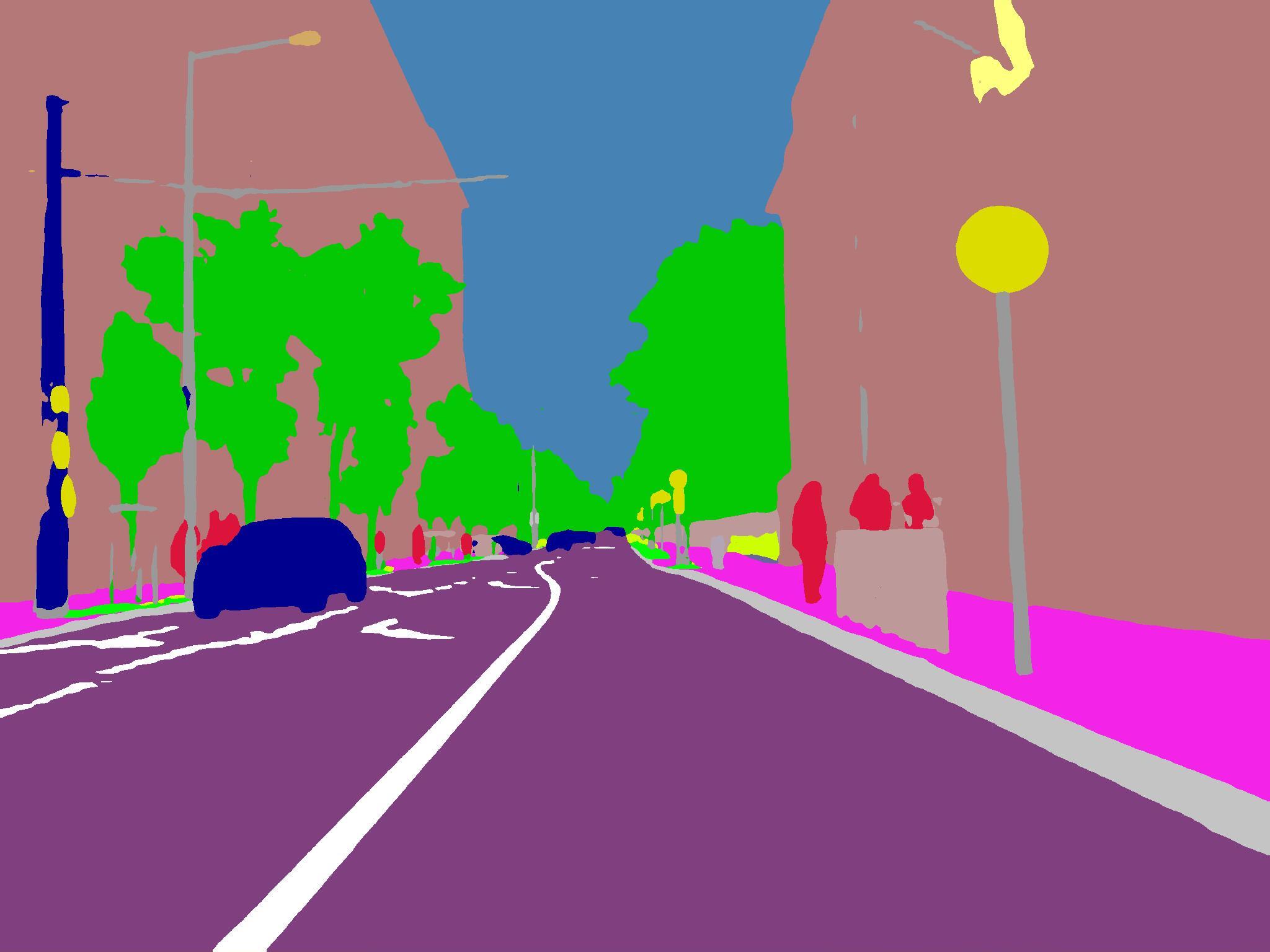}
    \\
    \includegraphics[height=\myheight]{figs/city/berlin_000177_000019_leftImg8bit_pred.jpg}
    \includegraphics[height=\myheight]{figs/wd/wd0093_100000_pred.jpg}
    \includegraphics[height=\myheight]{figs/scannet/scene0781_00_000200_pred.jpg}
    \includegraphics[height=\myheight]{figs/mvd/ZO6Xc8qwl0_S0G_ll8muhQ_pred.jpg}
     \\
    \caption{Qualitative performance of 
      our universal SNp-DN161 model 
      on test images from the seven RVC datasets.
      Rows 1 and 4 show input images,
      rows 2 and 5 show predictions 
      in the universal label space, 
      while rows 3 and 6 show 
      dataset-specific predictions. 
      Images belong to (top to bottom, left to right)
      ADE20k, Viper, KITTI, Cityscapes, 
      WildDash, ScanNet and Vistas. 
      Universal predictions find classes
      which are not labeled in 
      the corresponding dataset,
      eg.\ crosswalk, curb and road-marking 
      in the image from KITTI. 
      All road driving datasets represent vegetation 
      with a single class, but
      the universal model is able to classify those pixels
      more precisely as trees, palms and plants.
    }
    \label{fig:rvc}
\end{figure*}

\begin{figure*}
    \centering
    \includegraphics[width=0.48\textwidth]{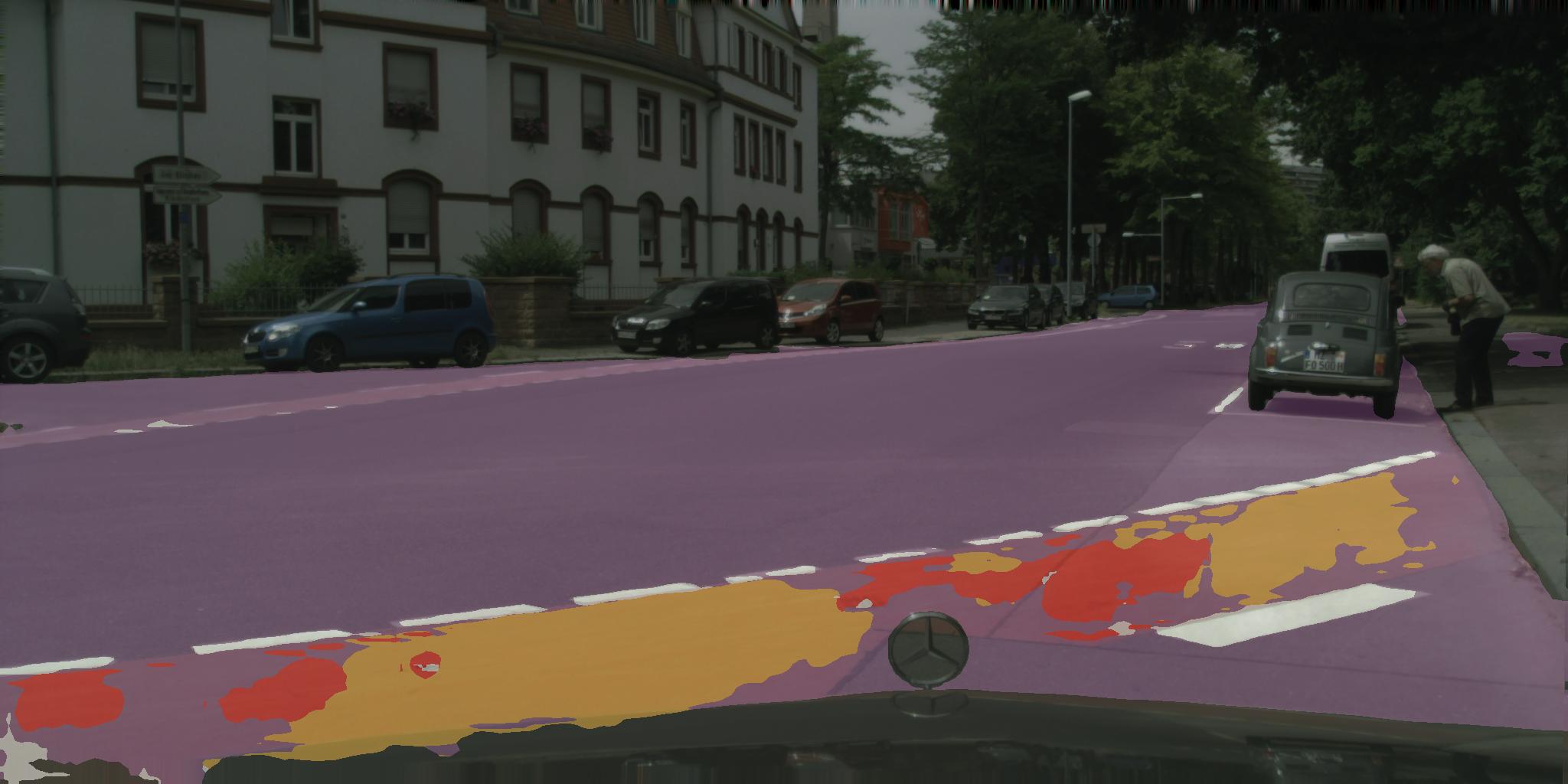}
    \includegraphics[width=0.48\textwidth]{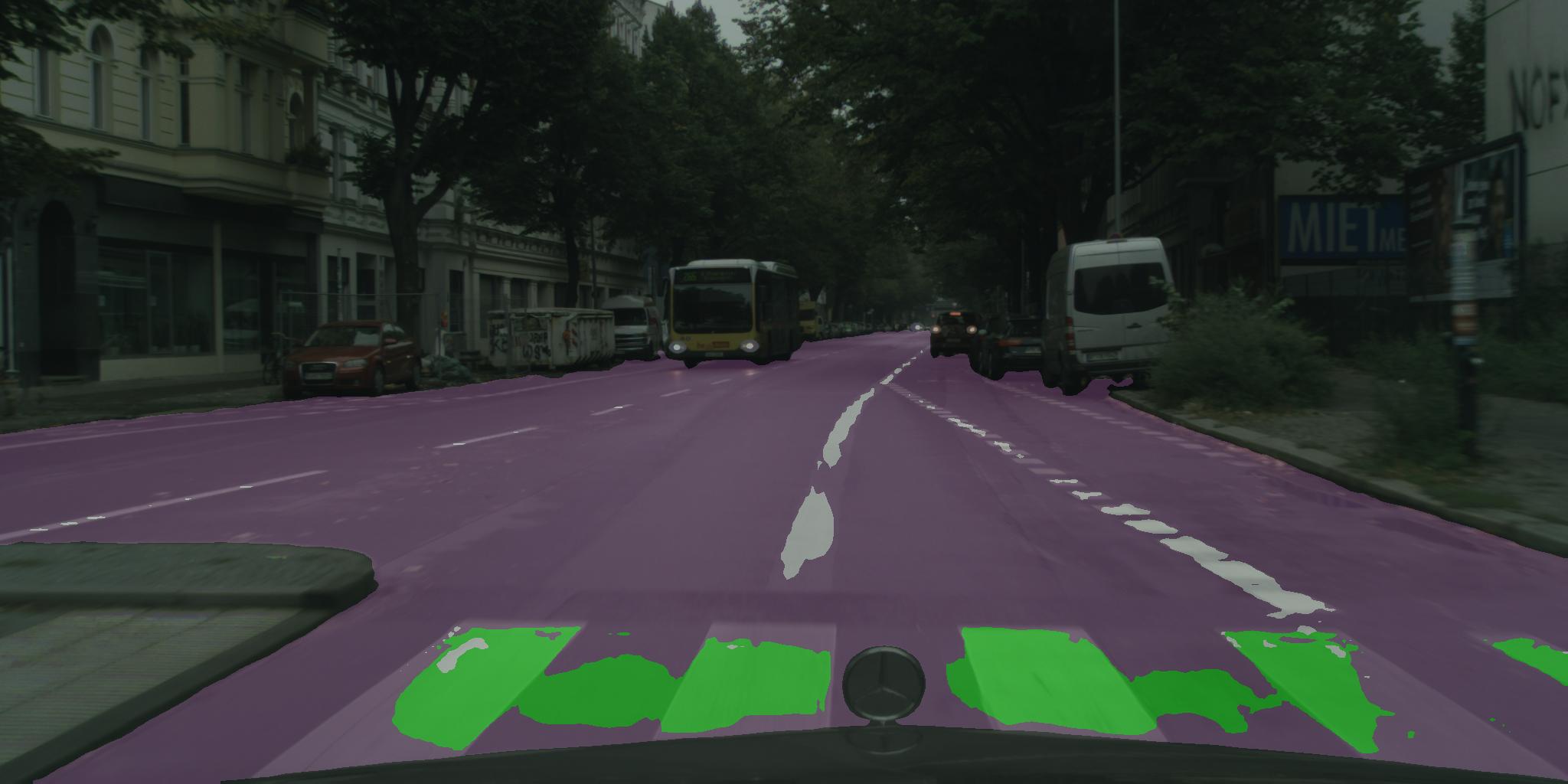}
    \\
    \includegraphics[width=0.48\textwidth]{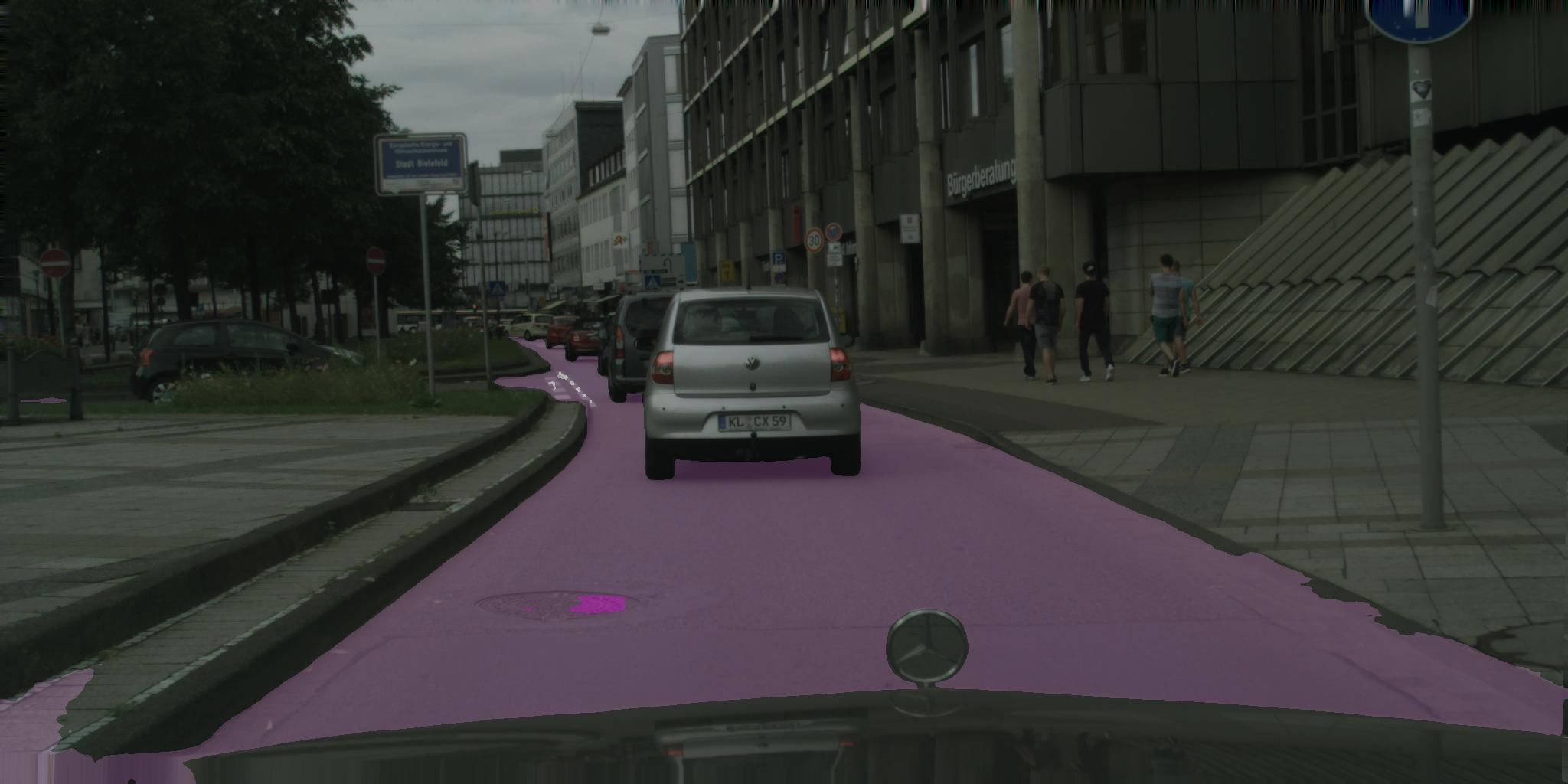}
    \includegraphics[width=0.48\textwidth]{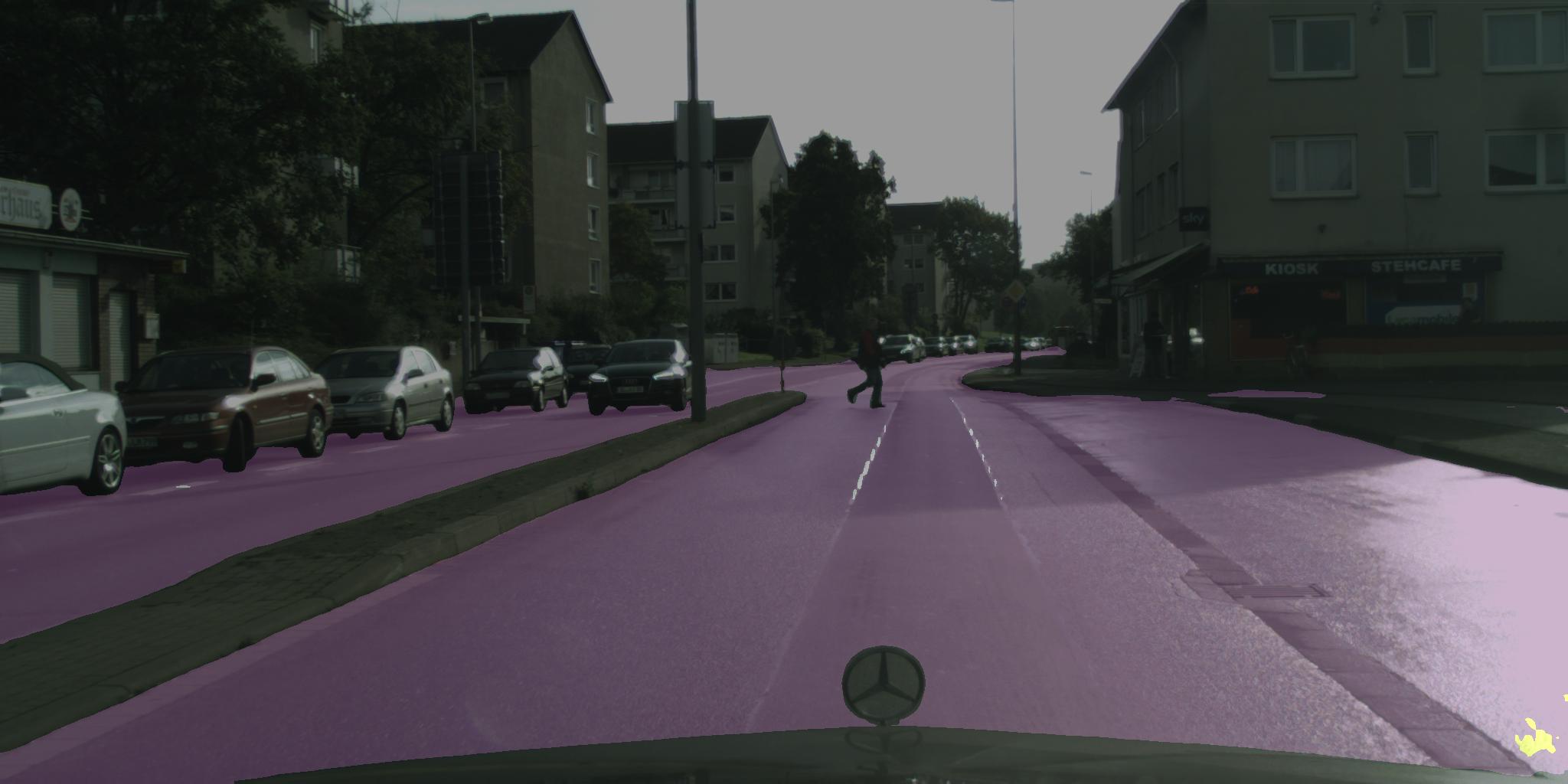}
    \caption{Qualitative performance 
      of our universal SNp-DN161 model
      on Cityscapes test.
      We overlay predictions over the input images.
      We show predictions of following universal classes:
      road (purple, all images), bike lane (yellow, top-left), 
      crosswalk (red, top-left), 
      zebra (green, top-right), 
      road marking (white, all images), 
      manhole (magenta, bottom-left) and 
      service lane (lemon yellow, bottom-right).
    }
    \label{fig:city_road_subclasses}
\end{figure*}

Figure \ref{fig:city_road_subclasses}
shows predictions of universal classes
which correspond to the class road 
in Cityscapes test.
Formally, the set of all such classes 
corresponds to 
$m_{\set S_{\mathrm{CS}}}(\mathrm{CS-road})$.
These universal classes are:
road, bike lane, crosswalk, zebra, road marking,
pothole, manhole and service lane. 
No instance of pothole was found 
in the entire Cityscapes test.
The best performing classes are 
road and road marking. 
We observe recognition of zebras 
only at close range, while otherwise 
they often get classified as road marking.
The remaining classes are usually detected 
as small clusters in the correct region. 
For example, manholes are often 
only partially segmented.

\begin{figure*}
    \centering 
    \includegraphics[width=0.33\textwidth]{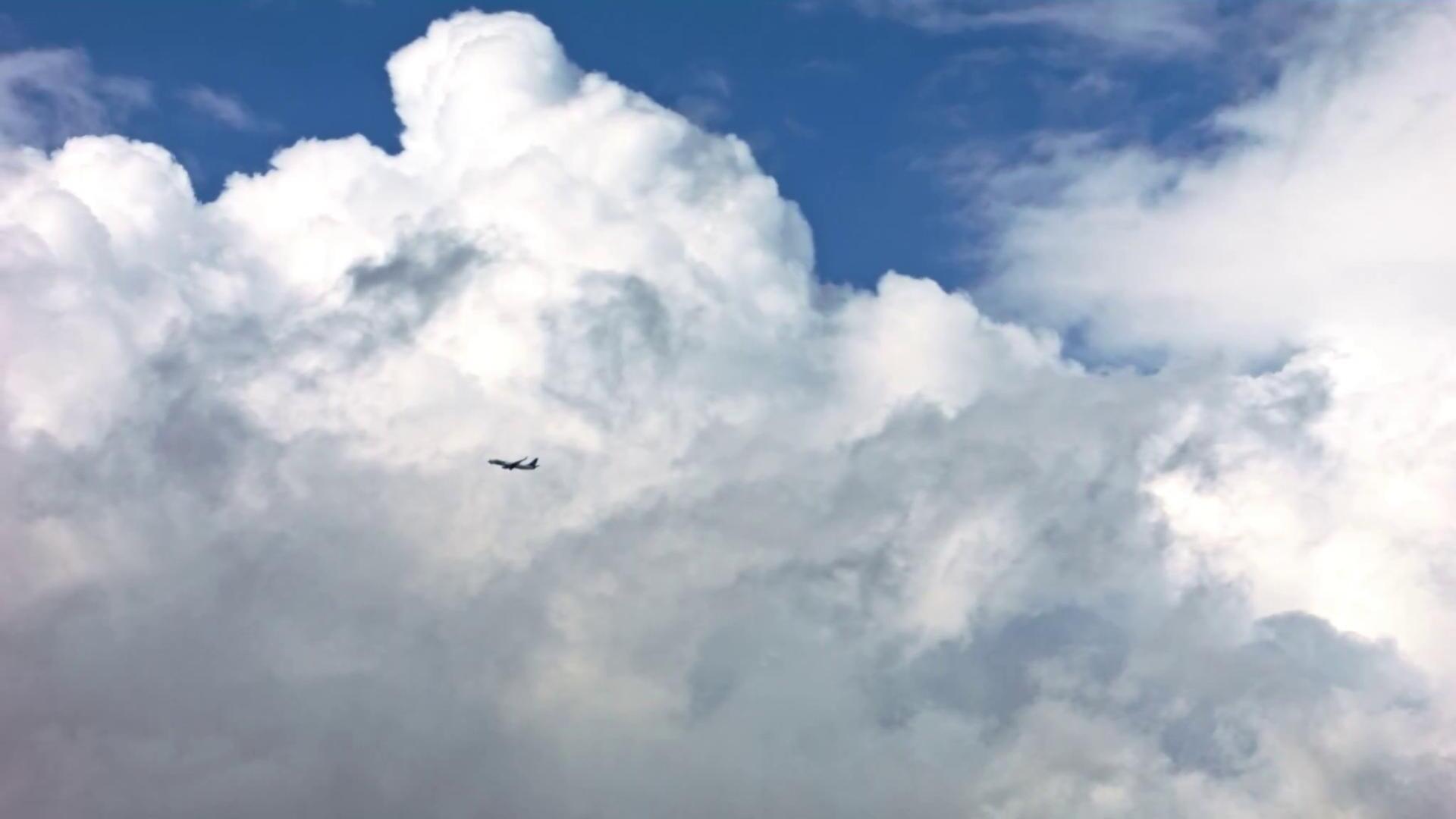}
    \includegraphics[width=0.33\textwidth]{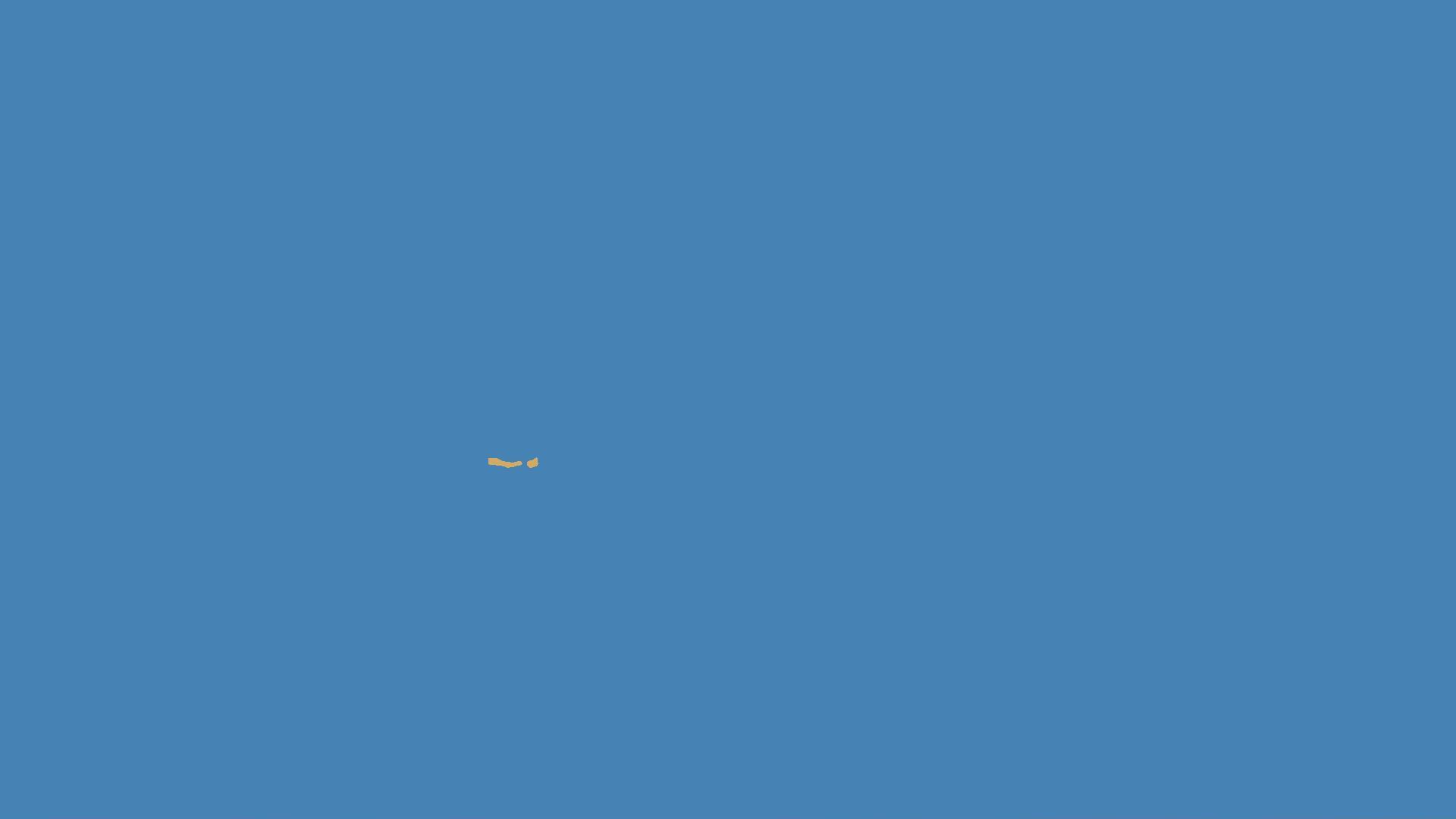}
    \includegraphics[width=0.33\textwidth]{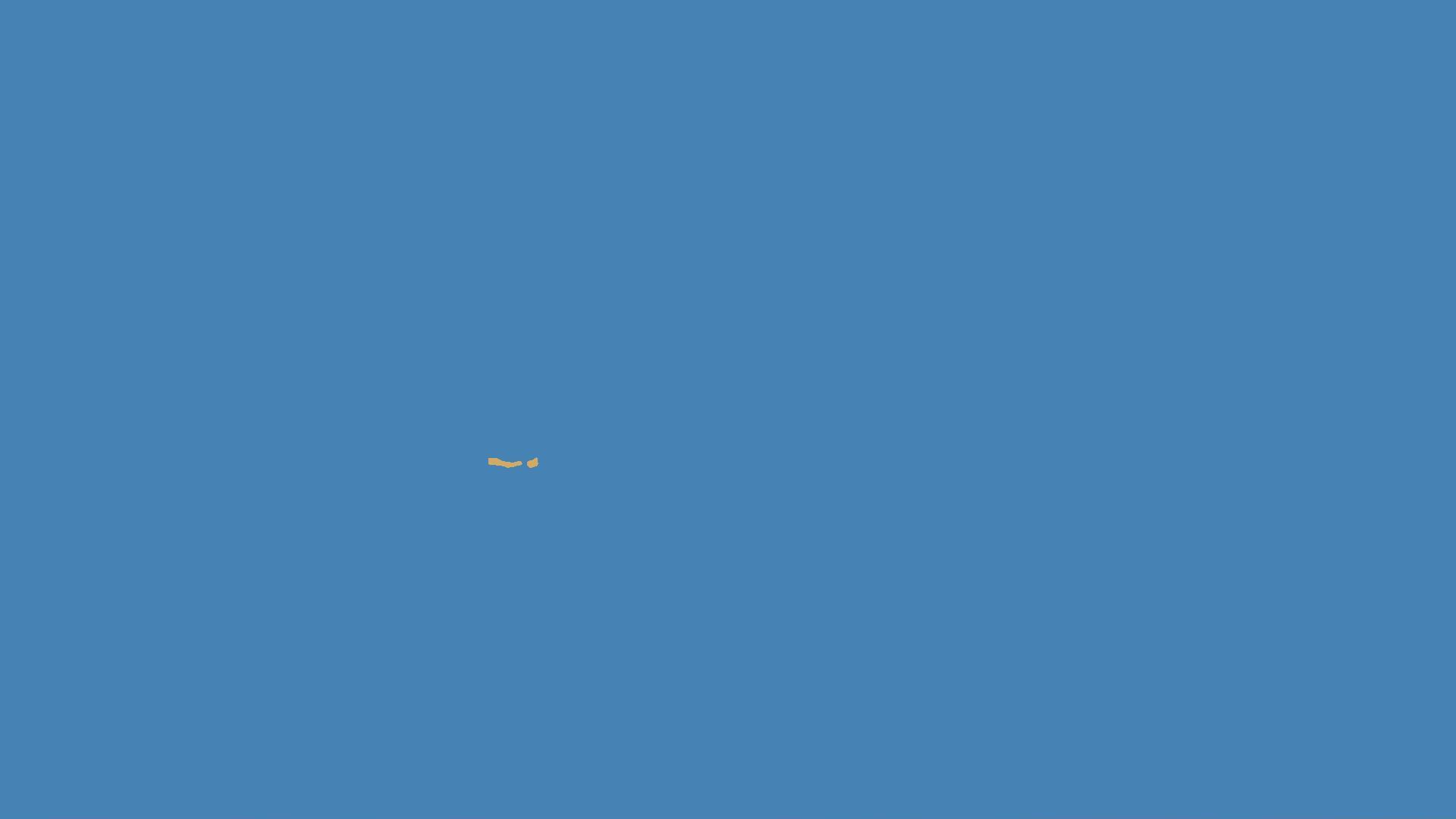}
    \\
    \includegraphics[width=0.33\textwidth]{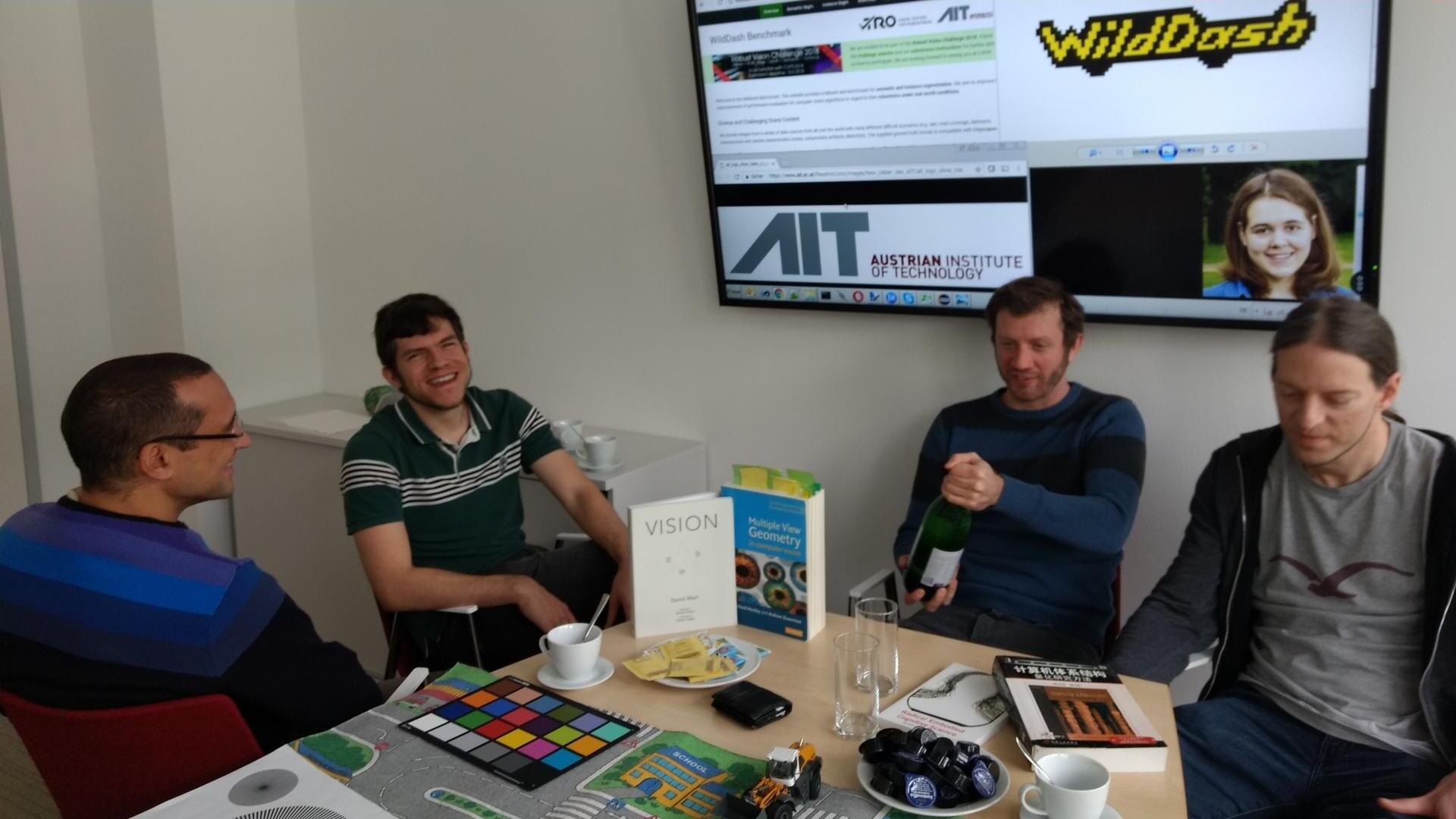}
    \includegraphics[width=0.33\textwidth]{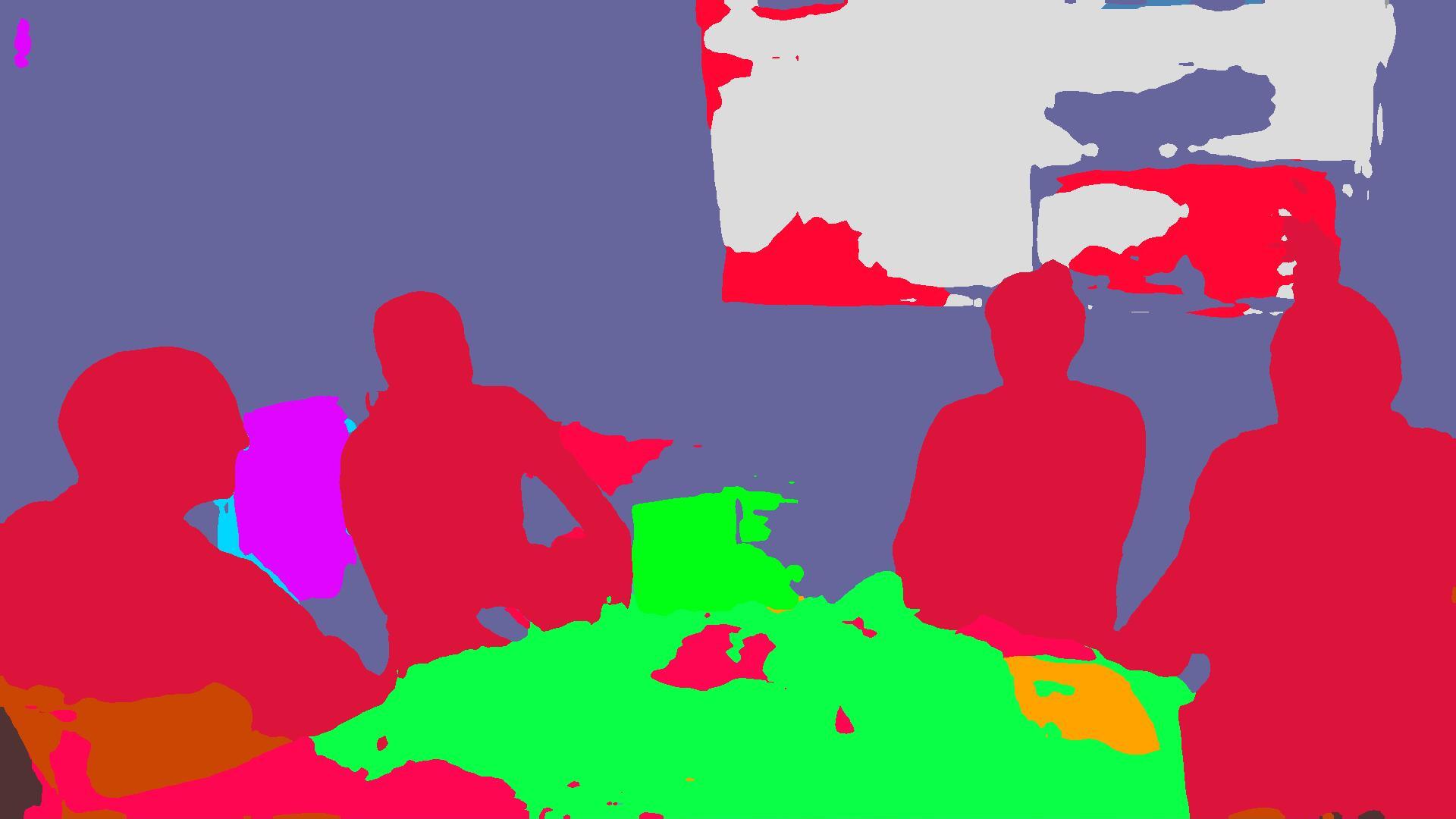}
    \includegraphics[width=0.33\textwidth]{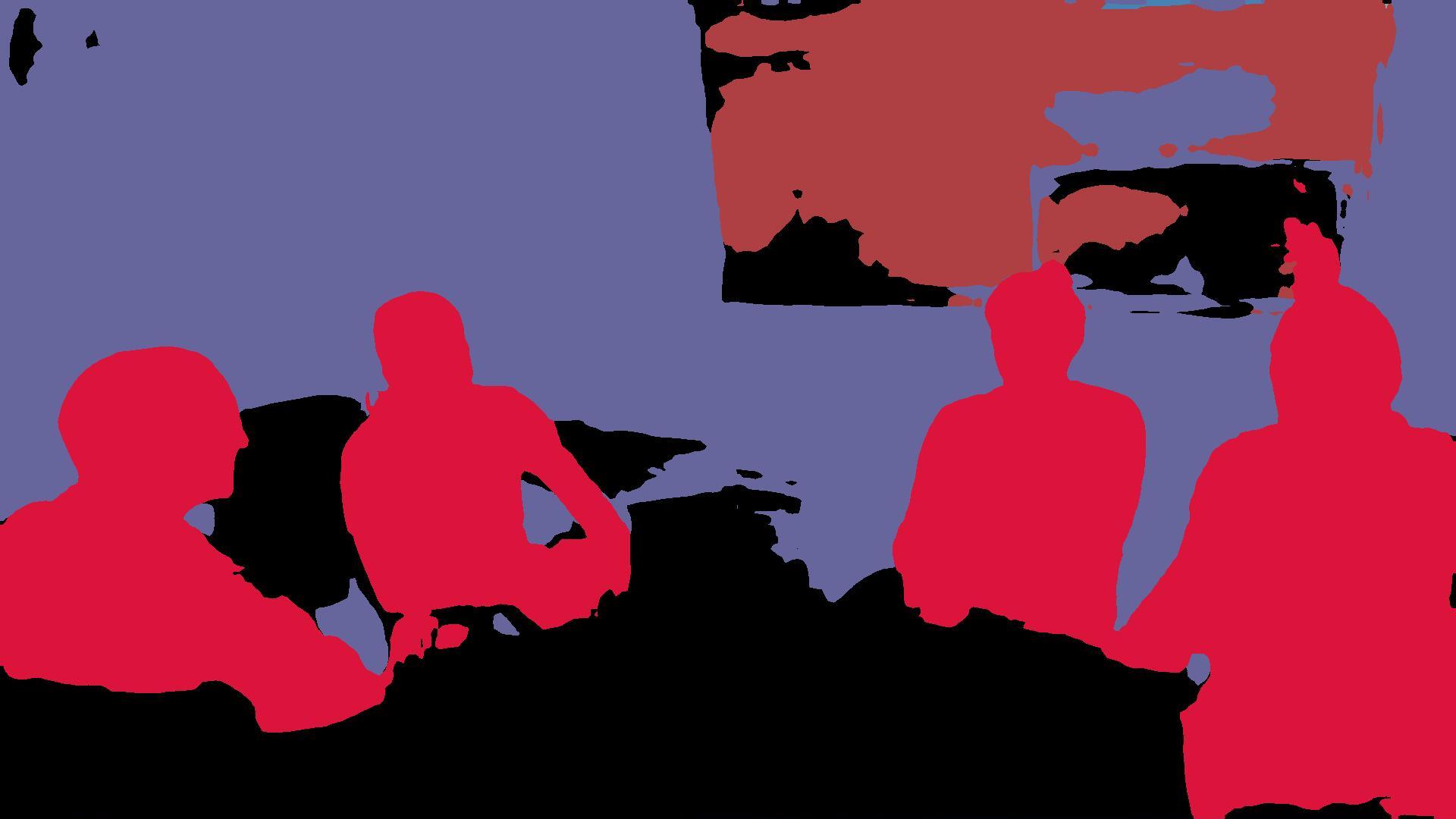}
    \\
    \includegraphics[width=0.33\textwidth]{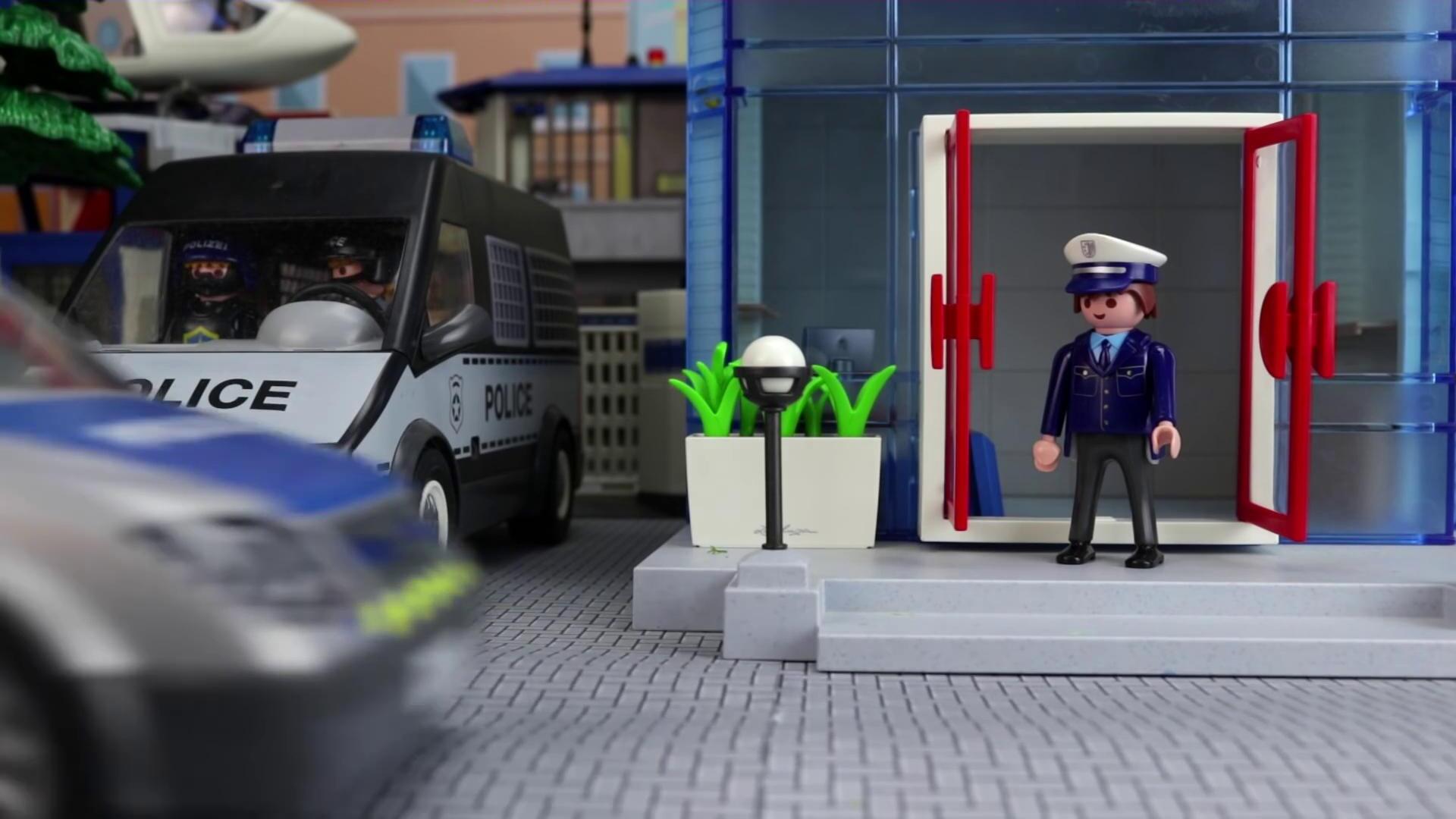}
    \includegraphics[width=0.33\textwidth]{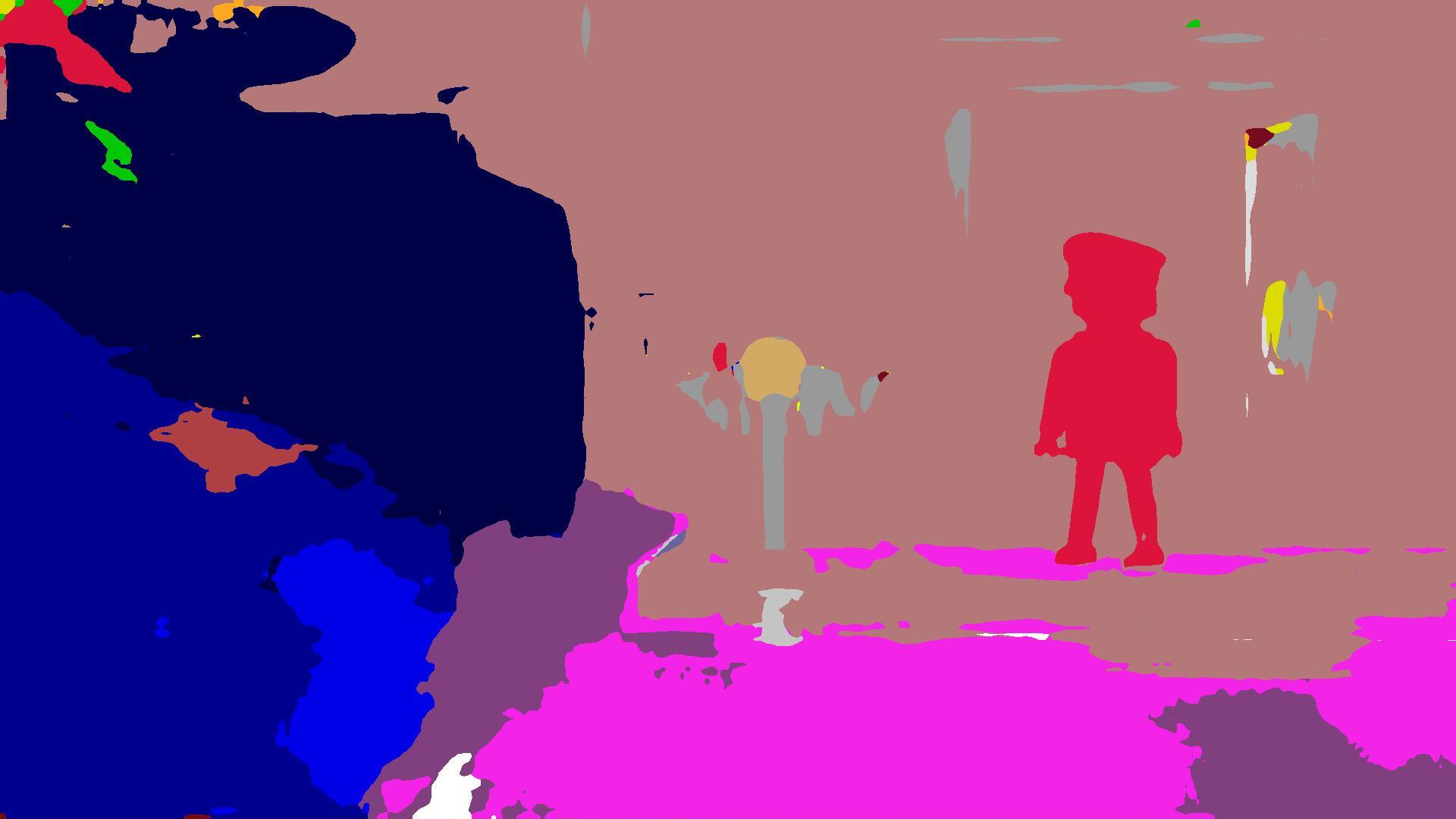}
    \includegraphics[width=0.33\textwidth]{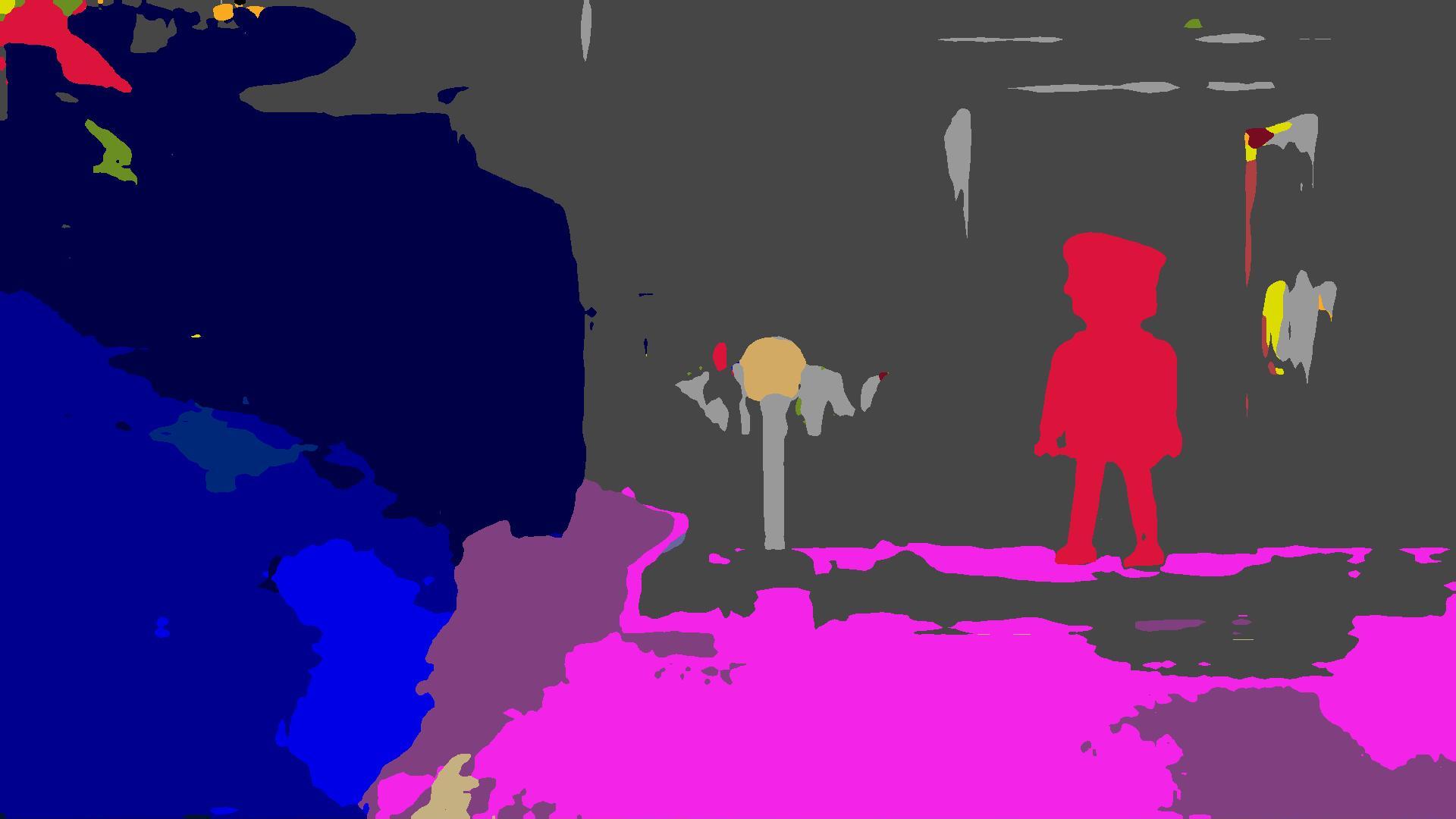}
    \\
    \includegraphics[width=0.33\textwidth]{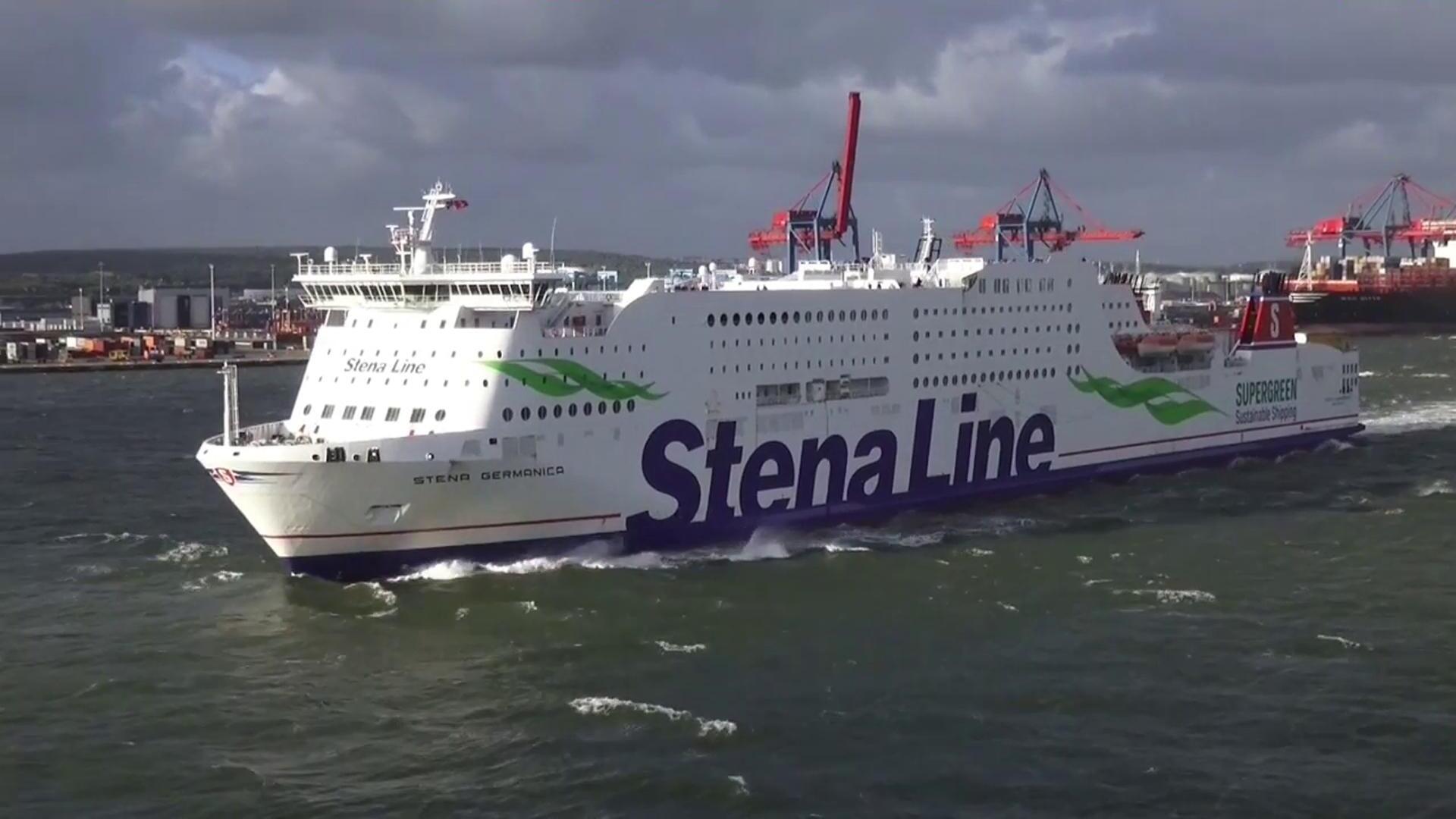}
    \includegraphics[width=0.33\textwidth]{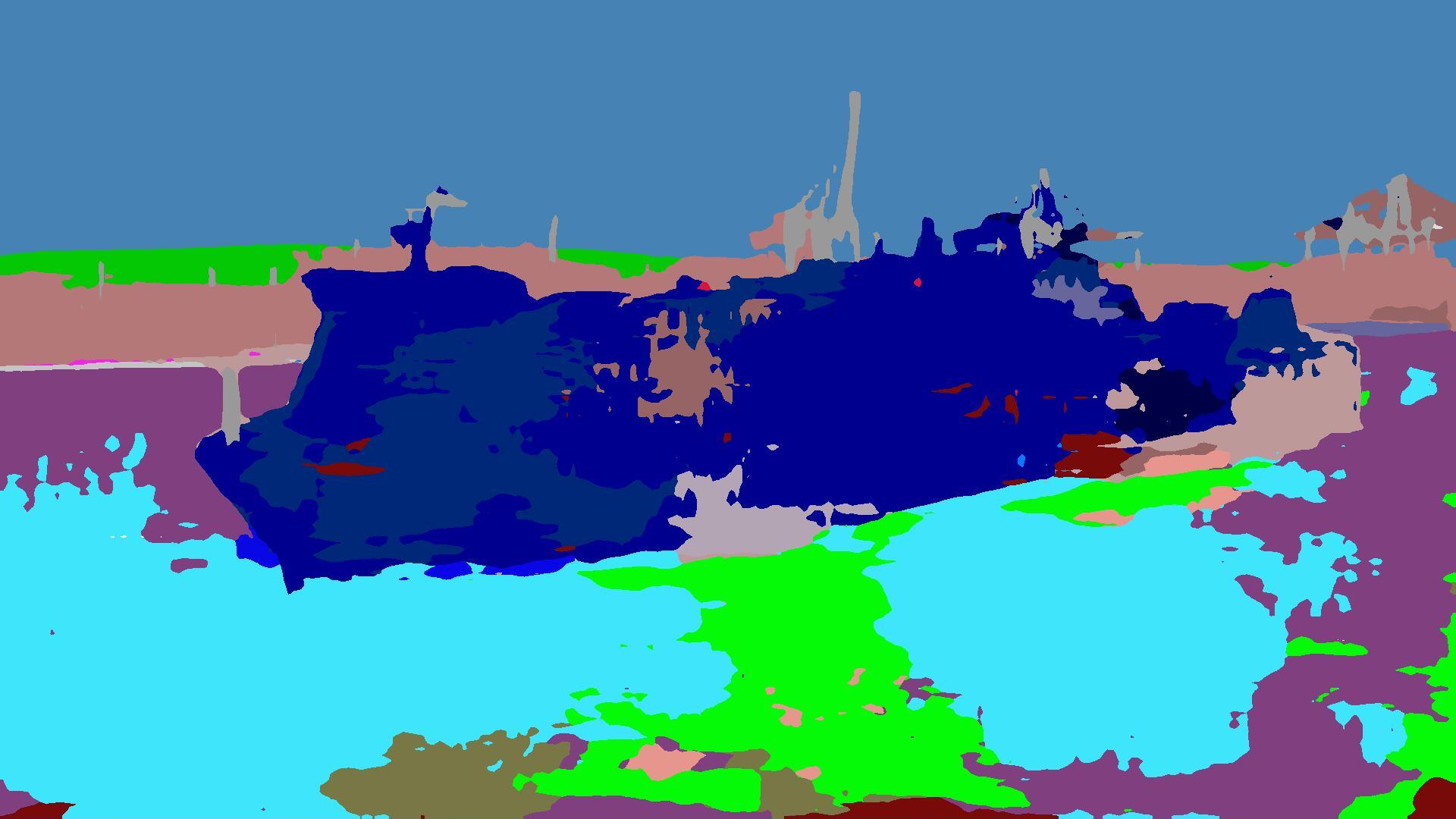}
    \includegraphics[width=0.33\textwidth]{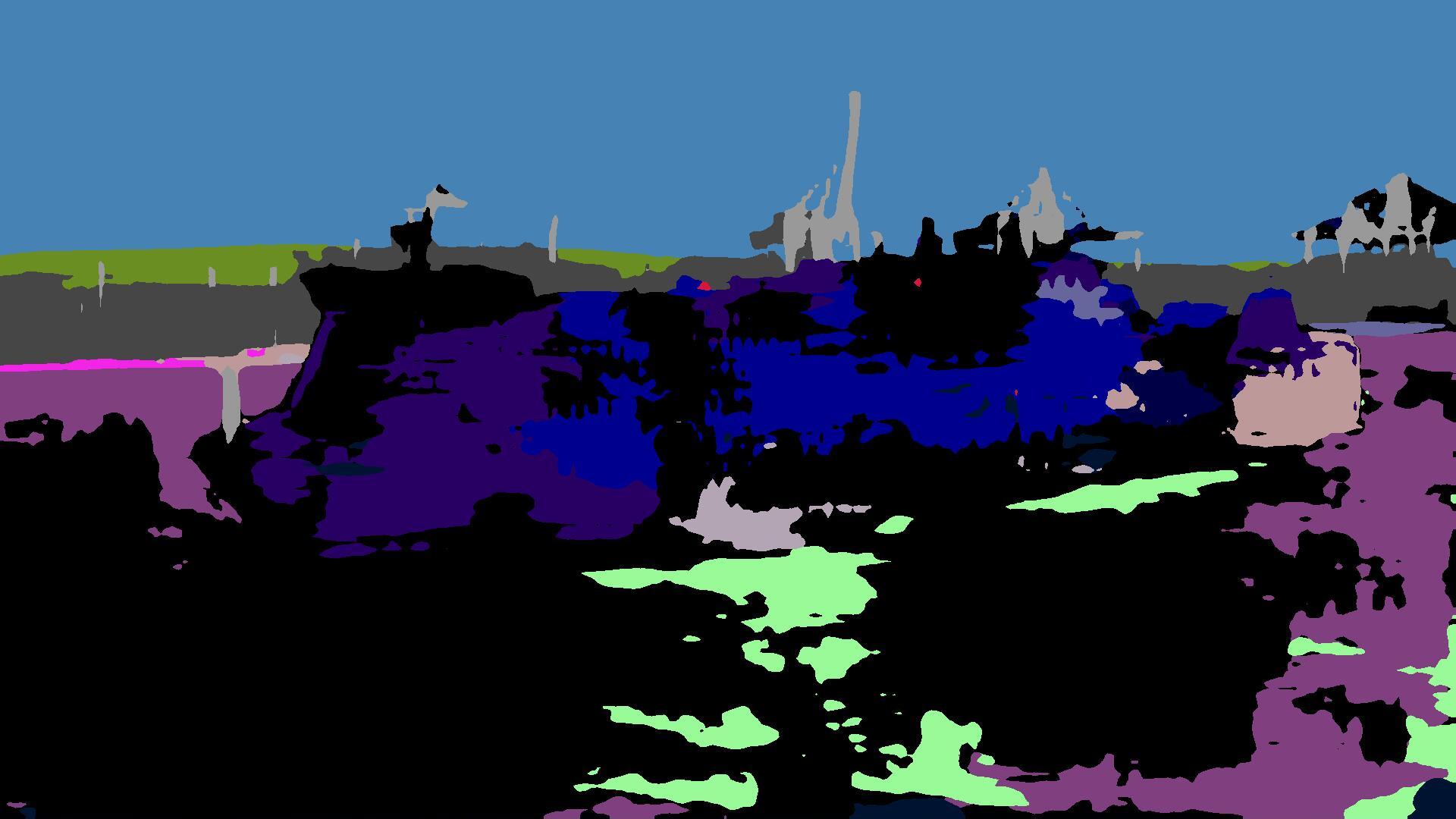}
    \\
    \includegraphics[width=0.33\textwidth]{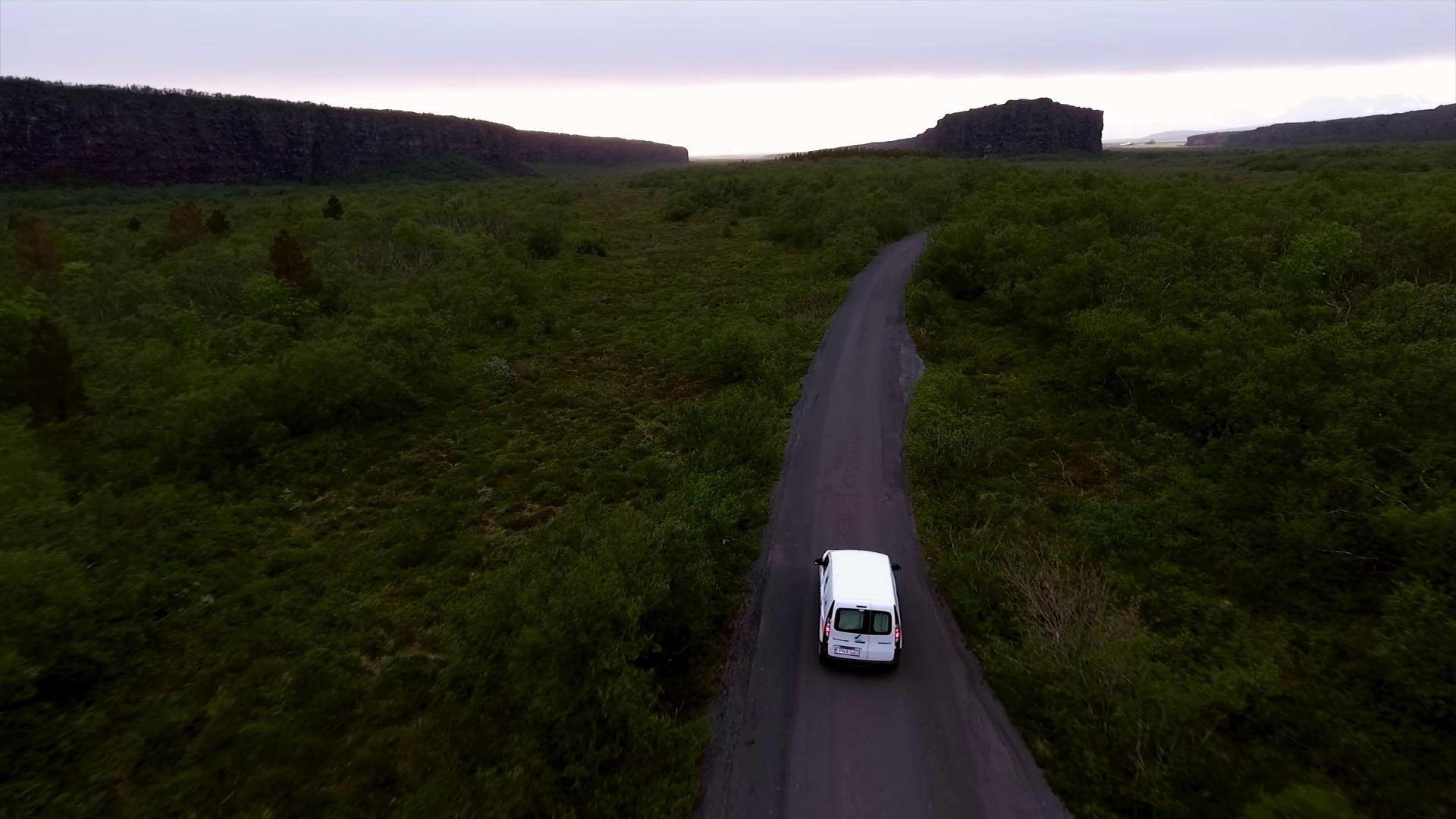}
    \includegraphics[width=0.33\textwidth]{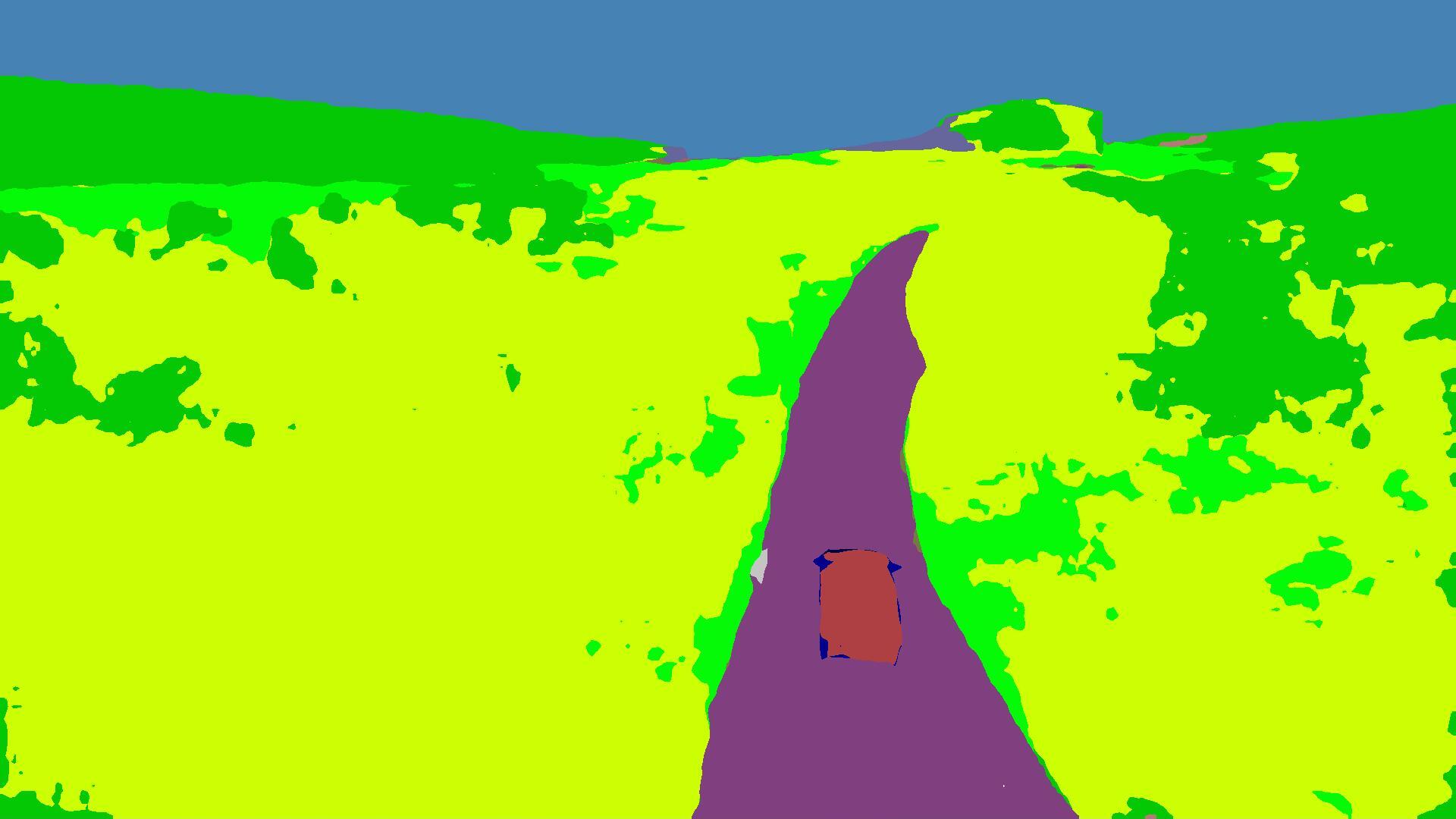}
    \includegraphics[width=0.33\textwidth]{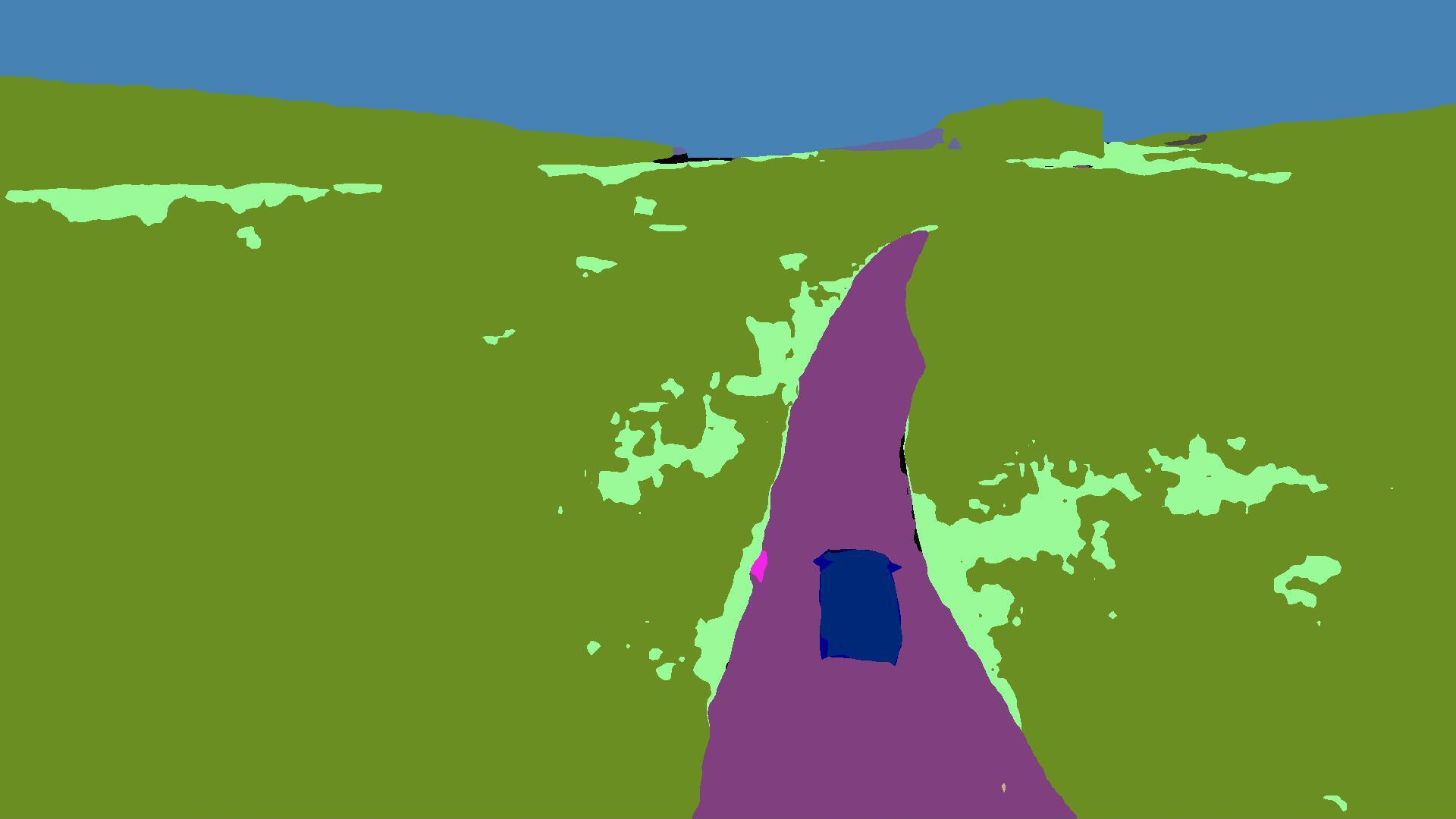}
    \caption{Performance of our universal model 
      on four WildDash 2 negative test images.
      The columns show the input image, 
      universal segmentation, 
      and the segmentation in the WildDash label space
      where class void is shown in black.
      Our model successfully recognizes 
      some non-traffic classes,
      e.g.\ table, chair, book, and cabinet (row 2), 
      or boat and water (row 5). 
      The model is robust to perspective changes (row 5)
      and exhibits fair performance in 
      presence of large domain-shift (row 3).
    }
    \label{fig:wd-negative}
\end{figure*}

Figure \ref{fig:wd-negative} shows the performance
of our model on negative images from WildDash.
These images were taken in 
non-road driving contexts (rows 1-4)
or from an unusual perspective (row 5). 
These images may contain classes 
found in traffic scenes such as people (row 2).
Note that the benchmark accepts 
either the best-case ground truth or
the void class (denoted with black).

\begin{figure*}
    \centering
    \includegraphics[width=0.33\textwidth]{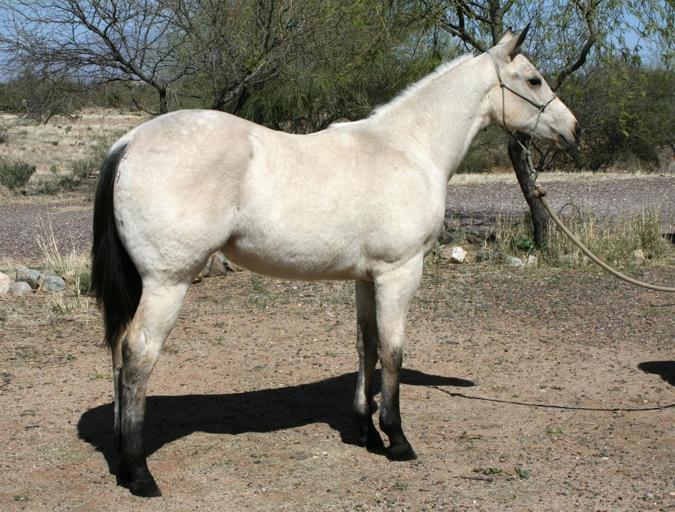}
    \includegraphics[width=0.33\textwidth]{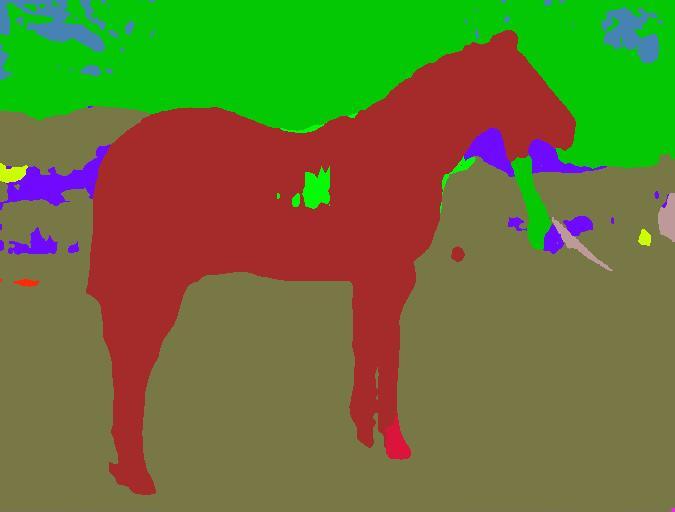}
    \includegraphics[width=0.33\textwidth]{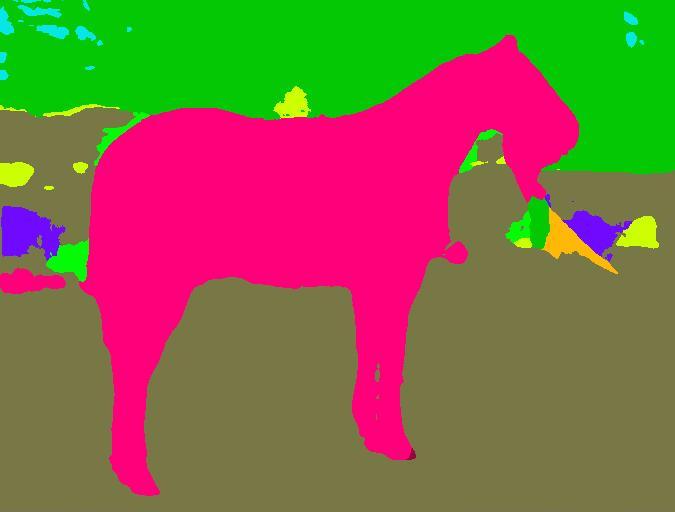}
    \caption{A failure case on ADE20k test.
      The columns show the input image, 
      universal segmentation, and 
      ADE20k segmentation. 
      The model recognizes most 
      of the horse as class bird.
      This occurs since birds and ground animals
      are annotated only in Vistas
      as extremely rare classes.
      Note that two patches are incorrectly classified 
      as grass (green) and pedestrian (red).
      These patches finaly get correctly classified
      into ADE20k-animal, since the sum of
      probabilities of classes bird and ground animal
      prevails after evaluation mapping.
    }
    \label{fig:ade-animal}
\end{figure*}

Figure \ref{fig:ade-animal}
shows an interesting failure of
our model to distinguish between different
types of animals. 
Our universal model classifies the horse 
from an ADE20k image as class bird.
This occurs since ADE20k taxonomy 
contains only the class animal
which we map to universal classes 
bird and ground-animal. 
The training signal for distinguishing
between birds and ground animals
should have come from Vistas,
but this signal was very weak
since these two Vistas classes 
are extremely rare.

\section{Mapping visualizations}

\begin{figure*}
    \centering
    \begin{subfigure}{0.95\textwidth}
    \centering
    \includegraphics[width=0.95\textwidth]{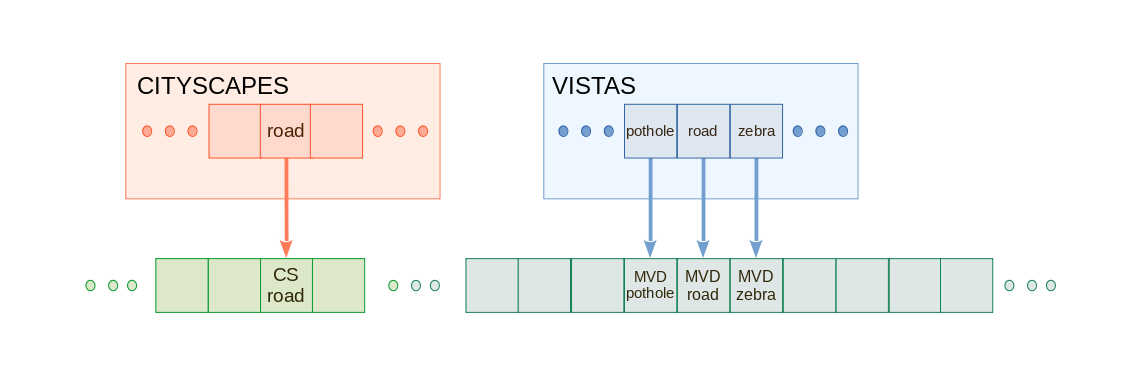}
    \caption{\\}
    \label{fig:mapping_graphs_train_nc}
    \end{subfigure}
    \\
    \begin{subfigure}{0.95\textwidth}
    \centering
    \includegraphics[width=0.95\textwidth]{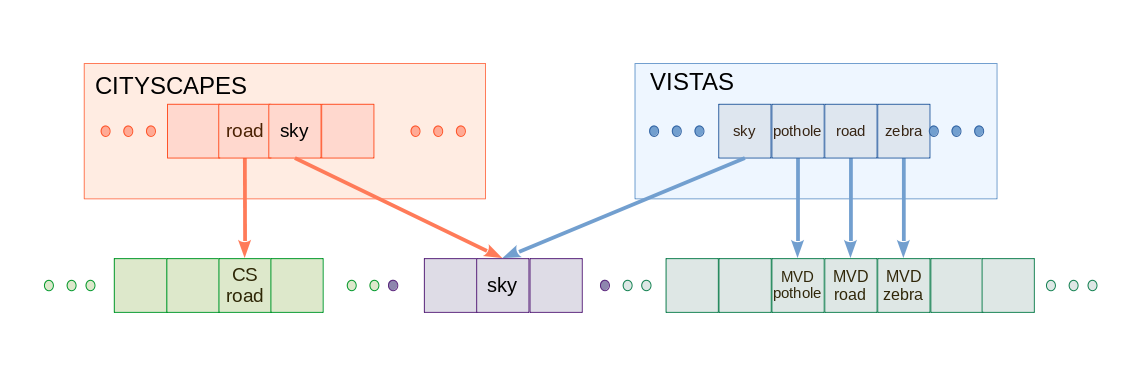}
    \caption{\\}
    \label{fig:mapping_graphs_train_pm}
    \end{subfigure}
    \\
    \begin{subfigure}{0.95\textwidth}
    \centering
    \includegraphics[width=0.95\textwidth]{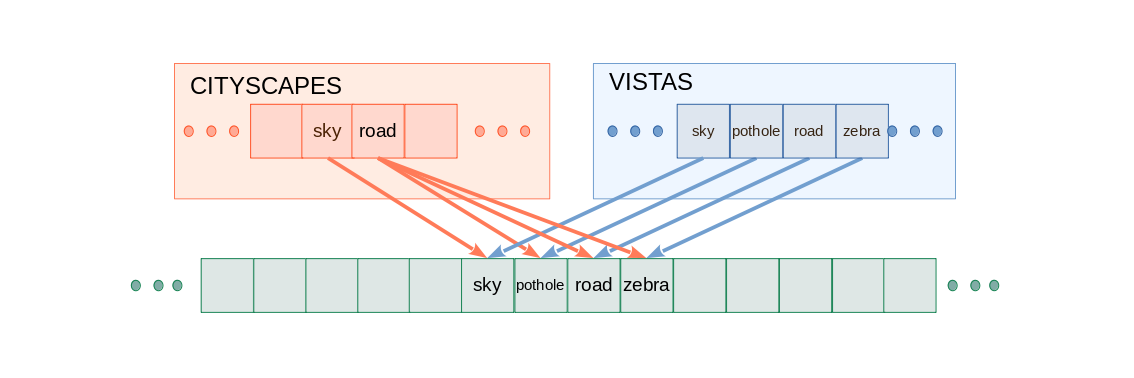}
    \caption{\\}
    \label{fig:mapping_graphs_train_u}
    \end{subfigure}
    \caption{Visualizations of training mappings
    in City-Vistas experiments.
    Naive concatenation (a) maps 
    each dataset-specific class  
    to the corresponding training logit.
    Partial merge (b) maps dataset-specific classes 
    to the common logit only if they match exactly.
    If this is not the case, 
    the classes remain separate.
    Our universal taxonomy (c) 
    maps each dataset-specific class 
    to one or more universal classes.
    }
    \label{fig:mapping_graphs_train}
\end{figure*}

\begin{figure*}
    \centering
    \begin{subfigure}{0.95\textwidth}
    \centering
    \includegraphics[width=0.95\textwidth]{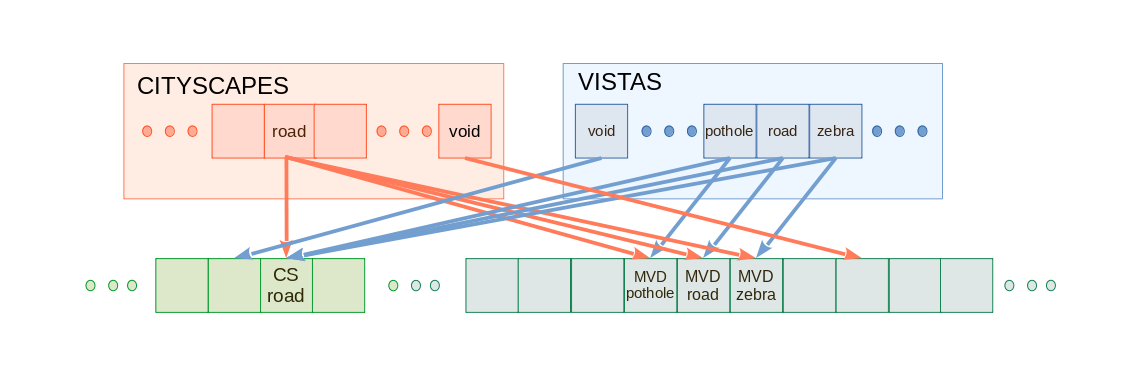}
    \caption{\\}
    \label{fig:mapping_graphs_eval_nc}
    \end{subfigure}
    \\
    \begin{subfigure}{0.95\textwidth}
    \centering
    \includegraphics[width=0.95\textwidth]{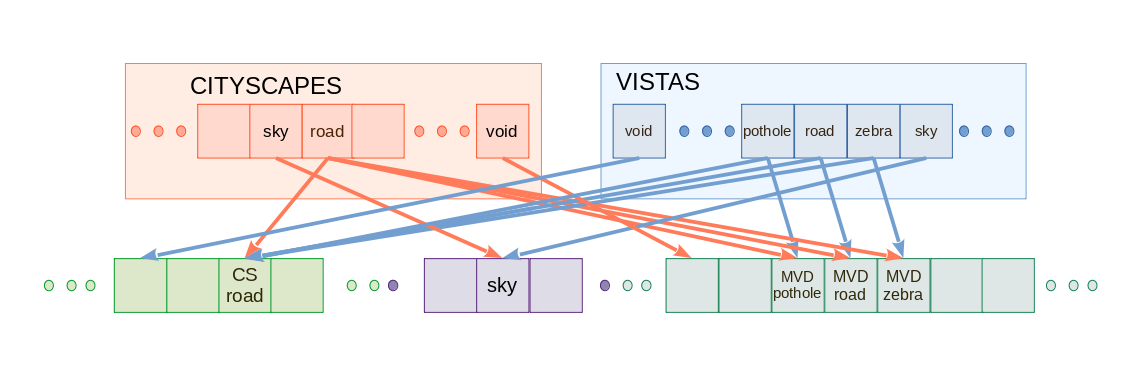}
    \caption{\\}
     \label{fig:mapping_graphs_eval_pm}
    \end{subfigure}
    \\
    \begin{subfigure}{0.95\textwidth}
    \centering
    \includegraphics[width=0.95\textwidth]{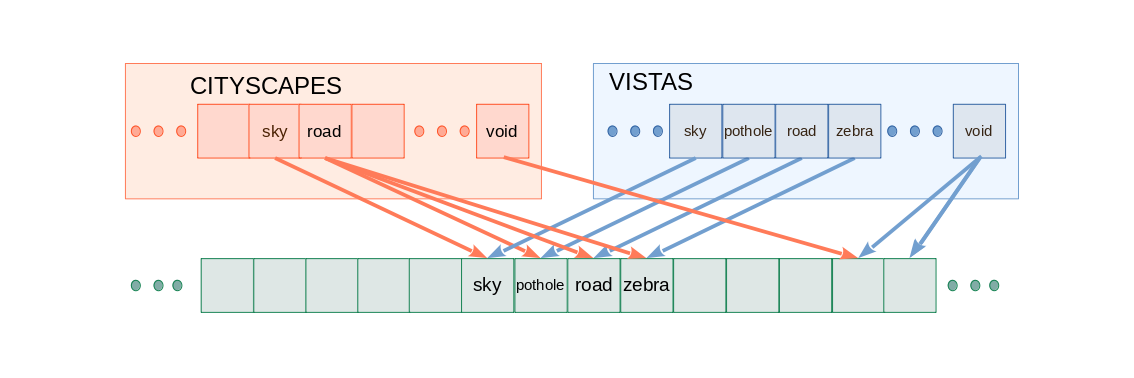}
    \caption{\\}
     \label{fig:mapping_graphs_eval_u}
    \end{subfigure}
    \caption{Visualizations of evaluation mappings
    in City-Vistas experiments
    for naive concatenation (a),
    partial merge (b) and universal taxonomy (c).
    In each of these cases,
    dataset-specific classes are mapped
    to training classes with which they overlap. 
    We extend dataset-specific taxonomies 
    with a void class 
    that maps to all training classes 
    which do not overlap 
    with any of dataset-specific classes.
    }
    \label{fig:mapping_graphs_eval}
\end{figure*}

Figures \ref{fig:mapping_graphs_train} and
\ref{fig:mapping_graphs_eval} visualize
training and evaluation mappings used 
in our City-Vistas experiments.
We show these mappings for the two baselines
and our universal taxonomy.
Please refer to Section 3 of the main paper
for a detailed description of
the procedure for recovering these mappings.

\end{appendices}

\end{document}